\documentclass[12pt]{article}
\usepackage{arxiv_new}
\usepackage{times}
\usepackage{longtable}
\usepackage{algorithm}
\usepackage{algorithmic}
\usepackage{nicefrac}
\usepackage{booktabs}
\usepackage{footmisc}
\usepackage[shortlabels]{enumitem}
\usepackage{multirow}
 \usepackage{soul}
 
\usepackage{array}
\usepackage{tabularx}
\usepackage{url}

\usepackage{makecell}
\usepackage{titlesec}
\usepackage{setspace}
\usepackage{bm}
\usepackage{mdframed}
\usepackage{bbm}
\usepackage{threeparttable}
\usepackage{subcaption}

\usepackage[T1]{fontenc}

\usepackage{titletoc}
\usepackage{longtable}
\usepackage{pifont} 
\usepackage{wrapfig}
\usepackage{forest}

\usepackage[table]{xcolor}
\usepackage{makecell}
\usepackage{graphicx}
\usepackage{amsmath}
\usepackage{caption}
\usepackage{tikz}
\usetikzlibrary{arrows.meta,calc,fit,positioning,shapes.geometric,shapes.misc}
\usepackage{adjustbox}
\usepackage[normalem]{ulem}
\newcommand{\largedot}{\raisebox{-0.2ex}{\scalebox{1.85}{$\bullet$}}}

\usepackage{enumitem}
\usepackage[most]{tcolorbox}
\usepackage{tabularx}
\usepackage{booktabs}
\usepackage{multirow}
\usepackage{adjustbox}
\usepackage{array}

\definecolor{archboxbg}{HTML}{FAFCFF}      
\definecolor{archboxframe}{HTML}{5E748C}   
\definecolor{archboxrule}{HTML}{3F566E}    
\definecolor{archboxhead}{HTML}{2F4358}    
\definecolor{archboxaccentA}{HTML}{8FA1B3}
\definecolor{archboxaccentB}{HTML}{C58C5B}
\definecolor{archboxaccentC}{HTML}{7D8F69}
\definecolor{archboxaccentD}{HTML}{9A7AA0}
\definecolor{archboxaccentE}{HTML}{C96A64}

\tcbset{
archquestionbox/.style={
enhanced,
breakable,
colback=archboxbg,
colframe=archboxframe,
boxrule=1.4pt,          
arc=1.5pt,
left=8pt,right=8pt,top=8pt,bottom=8pt,
before skip=8pt, after skip=8pt,
attach boxed title to top left={xshift=4pt,yshift=-2mm},
boxed title style={
colback=white,
colframe=archboxrule,
boxrule=1.0pt,         
arc=1pt,
left=5pt,right=5pt,top=3pt,bottom=3pt
},
fonttitle=\bfseries\small,
coltitle=archboxhead
}
}

\definecolor{surveyblue}{HTML}{4C78A8}
\definecolor{surveyorange}{HTML}{F58518}
\definecolor{surveygreen}{HTML}{54A24B}
\definecolor{surveypurple}{HTML}{B279A2}

\definecolor{scopegrayteal}{HTML}{7A9E9F}
\definecolor{scopebrown}{HTML}{C08457}
\definecolor{scopeviolet}{HTML}{7D6B91}

\definecolor{archblue}{HTML}{8FA1B3}
\definecolor{archbrown}{HTML}{C58C5B}
\definecolor{archgreen}{HTML}{7D8F69}
\definecolor{archpurple}{HTML}{9A7AA0}
\definecolor{archframe}{HTML}{D9D6CF}
\definecolor{archfill}{HTML}{F7F6F3}

\usepackage{tikz}

\usepackage{xcolor}
\definecolor{scoreDark}{HTML}{444444}
\definecolor{scoreMid}{HTML}{8A8A8A}
\definecolor{scoreLight}{HTML}{C9C9C9}

\definecolor{BerkeleyBlue}{RGB}{0,50,98}
\definecolor{RefOrange}{RGB}{230,120,40}

\usepackage[
    colorlinks=true,
    citecolor=RefOrange,
    linkcolor=BerkeleyBlue,
    urlcolor=RefOrange
]{hyperref}

\DeclareRobustCommand{\fullsup}{\textcolor{scoreDark}{\Large$\bullet$}}

\DeclareRobustCommand{\lowsup}{\textcolor{scoreDark}{\Large$\circ$}}

\definecolor{tabgray}{HTML}{F4F4F4}

\usepackage{fancyhdr} 

\usepackage[most]{tcolorbox} 

\definecolor{darkpastelgreen}{rgb}{0.01, 0.75, 0.24}

\titleformat*{\subparagraph}{\itshape}

\contentsmargin{2.2em}
\titlecontents{section}
  [0em]
  {\vspace{0.52em}\large\bfseries}
  {\contentslabel{3.0em}}
  {}
  {\titlerule*[0.7pc]{.}\contentspage}
\titlecontents{subsection}
  [3.0em]
  {\vspace{0.26em}\large}
  {\contentslabel{3.6em}}
  {}
  {\titlerule*[0.6pc]{.}\contentspage}

\newsavebox\actorsfigure

\usepackage{circledsteps}
\usepackage{fontawesome5}

\definecolor{LightGray}{RGB}{250,250,250}
\definecolor{Gray}{RGB}{240, 240, 240}

\definecolor{quoteLine}{RGB}{0,0,0} 
\definecolor{quoteText}{RGB}{0,0,0} 
\definecolor{authorText}{RGB}{0,0,0} 
\definecolor{hidden-draw}{RGB}{106,142,189} 
\definecolor{hidden-blue}{RGB}{194,230,247} 
\definecolor{hidden-orange}{RGB}{217, 230, 252}
\definecolor{hidden-red}{RGB}{189,106,118}    
\definecolor{hidden-yellow}{RGB}{189,175,106} 
\definecolor{hidden-green}{RGB}{106,189,142}  
\definecolor{hidden-orange}{RGB}{189,142,106} 
\definecolor{hidden-purple}{RGB}{142,106,189} 

\newcolumntype{L}[1]{>{\raggedright\arraybackslash}p{#1}}
\newcolumntype{C}[1]{>{\centering\arraybackslash}p{#1}}
\newcolumntype{M}[1]{>{\raggedright\arraybackslash}m{#1}}

\newcommand{\fullmark}{$\bullet$}
\newcommand{\halfmark}{$\circ$}
\newcommand{\surveytablecaption}[2]{\caption{\textbf{#1} #2}}

\NewDocumentCommand{\heng}
{ mO{} }{\textcolor{red}{\textsuperscript{\textit{Heng}}\textsf{\textbf{\small[#1]}}}}
\usepackage[numbers]{natbib}
\usepackage{amsfonts}
\usepackage{cleveref}

\definecolor{HumanOrange}{HTML}{F58518}
\definecolor{AIAutoBlue}{HTML}{4C78A8}
\definecolor{CoResearchGreen}{HTML}{54A24B}
\definecolor{ScopePurple}{HTML}{B279A2}

\definecolor{surveyblue}{HTML}{4C78A8}
\definecolor{surveyorange}{HTML}{F58518}
\definecolor{surveygreen}{HTML}{54A24B}
\definecolor{surveypurple}{HTML}{B279A2}

\newcommand{\roleSurvey}{\textsc{Survey}}
\newcommand{\roleSystem}{\textsc{System}}
\newcommand{\roleProject}{\textsc{Project}}

\definecolor{evalblue}{HTML}{4C78A8}
\definecolor{evalorange}{HTML}{F58518}
\definecolor{evalgreen}{HTML}{54A24B}
\definecolor{evalpurple}{HTML}{B279A2}
\definecolor{evalgray}{HTML}{8F8F8F}
\definecolor{tabgray}{gray}{0.97}

\newcommand{\roleBenchmark}{\textsc{Bench}}
\newcommand{\roleReview}{\textsc{Review}}
\newcommand{\roleAudit}{\textsc{Audit}}

\title{AutoResearch AI: Towards AI-Powered Research Automation for Scientific Discovery}

\author{
{\bfseries Guiyao Tie$^{1}$}\quad
{\bfseries Jiawen Shi$^{1}$}\quad
{\bfseries Dingjie Song$^{2}$}\quad
{\bfseries Yixiao Huang$^{1}$}\quad
{\bfseries Ziji Sheng$^{1}$}\quad
{\bfseries Xueyang Zhou$^{1}$}\quad\\
{\bfseries Yongchao Chen$^{3}$}\quad
{\bfseries Daizong Liu$^{4}$}\quad
{\bfseries Pan Zhou$^{1}$}\quad
{\bfseries Ran Xu$^{5}$}\quad
{\bfseries Lifang He$^{2}$}\quad\\
{\bfseries Qingsong Wen$^{6}$}\quad
{\bfseries Manling Li$^{7}$}\quad
{\bfseries Cong Lu$^{8}$}\quad 
{\bfseries Shuai Li$^{9}$}\quad
{\bfseries Pengtao Xie$^{10}$}\quad
{\bfseries Yixuan Yuan$^{11}$}\quad\\
{\bfseries Rui Meng$^{14}$}\quad
{\bfseries Lei Xing$^{13}$}\quad
{\bfseries Lichao Sun$^{2}$}\quad
{\bfseries Caiming Xiong$^{15}$}\quad
{\bfseries Philip S. Yu$^{12}$}\quad
{\bfseries Jianfeng Gao$^{16}$}\quad
\\[1.2em]
{\normalfont $^{1}$Huazhong University of Science and Technology}\quad
{\normalfont $^{2}$Lehigh University}\quad
{\normalfont $^{3}$Tsinghua University}\quad
{\normalfont $^{4}$Wuhan University}\quad
{\normalfont $^{5}$Salesforce Research}\hspace{0.3em}
{\normalfont $^{6}$Squirrel AI Learning}\hspace{0.3em}
{\normalfont $^{7}$Northwestern University}\hspace{0.3em}
{\normalfont $^{8}$Independent}\hspace{0.3em}
{\normalfont $^{9}$Shanghai Jiao Tong University}\quad
{\normalfont $^{10}$University of California San Diego}\quad
{\normalfont $^{11}$Chinese University of Hong Kong}\quad
{\normalfont $^{12}$University of Illinois Chicago}\quad
{\normalfont $^{13}$Stanford University}\quad
{\normalfont $^{14}$Google Cloud AI Research}\quad
{\normalfont $^{15}$Recursive Superintelligence}\quad
{\normalfont $^{16}$Microsoft Research}
}

\begin{document}
\maketitle




\begin{figure}[ht]
\centering
\includegraphics[width=1\linewidth]{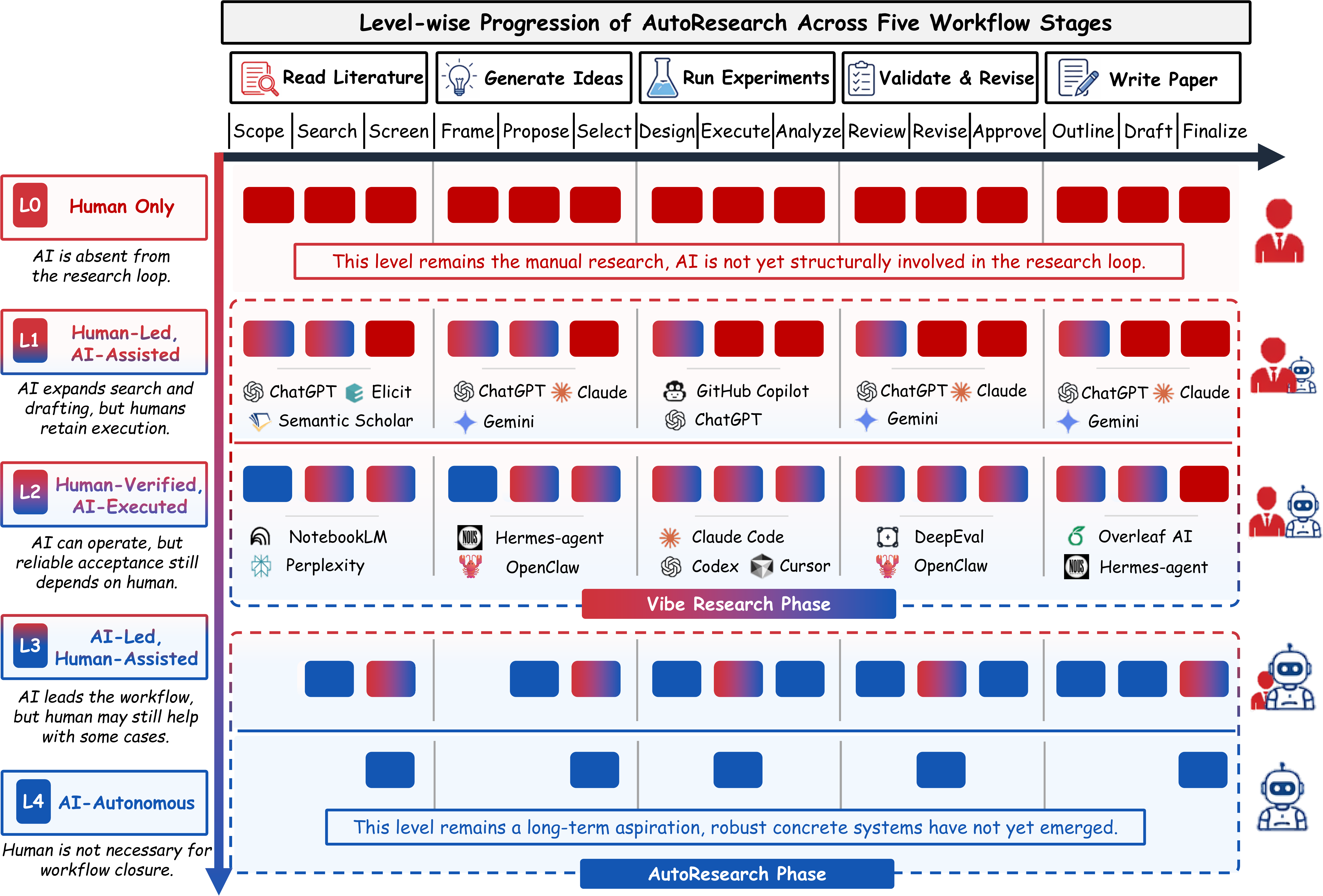}
\caption{
\textbf{Level-wise decomposition of AutoResearch.}
The figure shows human--AI responsibility shifts across the L0--L4 autonomy spectrum and five scientific workflow stages, distinguishing \emph{Vibe Research} (L1--L2) from broader \emph{AutoResearch} (L3--L4) at the workflow-step level.
}
\label{fig:cover_map}
\end{figure}

\newpage

\begin{abstract}
Scientific research is increasingly being reshaped by AI systems that move beyond isolated assistance and enter longer-horizon processes of literature grounding, hypothesis generation, experimentation, validation, reporting, and revision. This shift marks a transition from task-level AI for Science toward workflow-level research automation. However, the field remains fragmented: existing systems differ substantially in autonomy, domain scope, execution environment, validation mechanism, and reliance on human oversight. Although many systems can generate plausible ideas, operate tools, run bounded experiments, or produce polished artifacts, they still face persistent challenges in evidence preservation, reproducibility, rejection of weak directions, provenance tracking, cross-domain robustness, and accountable scientific closure.
This survey examines these developments through the lens of \emph{AutoResearch}, which we define as the developmental spectrum of AI-powered scientific workflow automation. Within this spectrum, \emph{Vibe Research} denotes the human-steered region where AI expands local research capability through prompt-based assistance and human-verified execution, while emerging AI-led systems begin to coordinate larger portions of the discovery loop without yet achieving robust autonomy. Rather than classifying prior work only by model family, agent architecture, or benchmark performance, we analyze how research systems redistribute control, evidence, execution, validation, and accountability across scientific workflows.
We organize the technical foundations of AutoResearch around five recurring workflow conditions: literature and research grounding, hypothesis formation and planning, experimentation and tool use, feedback, validation, and review, and reporting and knowledge communication. We further synthesize AI scientist systems, mixed-initiative co-research frameworks, benchmark ecosystems, domain-specific deployments, and open-source infrastructures within a unified analytical framework. To assess progress, we propose five evaluation dimensions—novelty, validity, impact, reliability, and provenance—that shift attention from task completion alone to the scientific credibility of workflow-level outputs.
Our analysis shows that the practical ceiling of AutoResearch is strongly domain-conditioned: higher autonomy is currently more credible in settings where research artifacts are structured, executable, and rapidly verifiable, and more limited where scientific claims depend on embodied experimentation, delayed validation, heterogeneous evidence, ethical constraints, or institutional accountability. By connecting conceptual boundaries, technical foundations, evaluation logic, and domain-conditioned autonomy ceilings, this survey clarifies the current landscape of AutoResearch and identifies the requirements for trustworthy AI participation in scientific inquiry.
\end{abstract}

\newpage
{\setstretch{1.05}\tableofcontents}

\newpage
\section{Introduction}\label{sec:introduction}

Artificial intelligence has influenced scientific research for many years, but the form of that influence has changed substantially. Earlier waves of AI for Science were dominated by specialized models and task-specific systems that targeted well-defined scientific subproblems, such as molecular property prediction, scientific imaging, automated data analysis, literature retrieval, and domain-specific simulation or optimization~\citep{luo2025llm4sr}. A canonical example is AlphaFold, whose success in protein structure prediction demonstrated how a highly capable AI system could transform an important scientific task while still operating within a relatively narrow and well-specified problem setting~\citep{Jumper2021AlphaFoldNature}. More recently, however, the capability frontier has shifted from narrow prediction and retrieval toward stronger language understanding, reasoning, retrieval-augmented synthesis, tool use, code generation, and iterative multi-step execution~\citep{Gridach2025Agentic, wei2025ai, Zhang2025TheEvolvingRoleofLar}. This change matters because it expands not only how well AI can perform isolated scientific tasks, but also how broadly it can participate across the research process itself: systems are increasingly able to assist with literature grounding, support idea generation, help formulate plans, execute code and tools, analyze intermediate outputs, and contribute to reporting and revision~\citep{ZHENG2025Automation, Muskaan_Goyal_2025, Hasib_2025}. The resulting transition is therefore not simply from weaker models to stronger models, but from local task enhancement to the growing possibility of workflow-level research automation. Recent systems such as \textit{The AI Scientist}~\citep{Lu2024AIScientist} make this shift especially visible, because they no longer target only one scientific subtask, but instead attempt to connect idea generation, code writing, experimentation, analysis, and manuscript production within an integrated research pipeline whose outputs still require scientific verification~\citep{Lu2024AIScientist, Yamada2025AIScientistV2, Kon2025Curie, PiFlow2025}. It is this broader transition-from task-specific AI for Science to increasingly workflow-oriented research automation-that motivates the present survey~\citep{Undermind2025Largelanguagemodelsforautoma, Liu2025AVisionforAutoResear}.

A recent wave of systems has begun to translate this broader possibility into concrete research practice. At the lighter end, literature-grounded and deep-research-style systems expand what AI can do in search, synthesis, and structured knowledge support, as illustrated by \textit{LitLLM}~\citep{Agarwal2024LitLLM}, \textit{OpenScholar}~\citep{OpenScholarGitHub}, and \textit{PaperQA2}~\citep{PaperQA2_2024, PaperQA2GitHub}. At a more execution-oriented level, controllable workspaces and coding substrates such as \textit{OpenHands}~\citep{Undermind2024OpenHandsAnOpenPlatformforAI}, \textit{Aider}~\citep{AiderGitHub}, and \textit{SWE-agent}~\citep{SWEAgentGitHub} have made it increasingly practical for AI to operate on files, tools, and experimental artifacts under human guidance. More recently, integrated AutoResearch systems and operational stacks have begun to connect broader spans of the research loop, from ideation and experiment design to execution, analysis, and drafting, as seen in \textit{The AI Scientist}~\citep{Lu2024AIScientist}, \textit{AI Scientist-v2}~\citep{Yamada2025AIScientistV2}, \textit{Agent Laboratory}~\citep{AgentLaboratoryGitHub}, \textit{AI-Researcher}~\citep{HKUDSAIResearcherGitHub}, \textit{ARIS}~\citep{ARISGitHub}, and \textit{NanoResearch}~\citep{nanoresearch2026}. Taken together, these developments suggest that research automation is no longer only a speculative ambition or a collection of isolated model demonstrations, but an emerging systems-level direction of AI for Science. At the same time, pipeline integration should not be equated with achieved scientific autonomy. Existing systems are already strong in search, drafting, coding, and some forms of bounded execution, but they remain much weaker at validation, rejection, exception handling, reproducibility, and accountable scientific closure~\citep{Chen2025AIRSBench, SPOT2025ScientificPaperErrorDetection, Gueroudji_2025, Xie2025How}. Existing surveys have recognized important parts of this landscape, but they still differ substantially in scope, unit of analysis, and implicit assumptions about autonomy~\citep{ZHENG2025Automation, Gridach2025Agentic, wei2025ai, Tie2025Survey, Chen2025AI4Research, Liu2025AVisionforAutoResear}. A workflow-centered account is therefore needed to compare these systems, their autonomy claims, and their scientific limits within a single analytical frame.

To compare this emerging but still fragmented landscape within a common analytical frame, this survey adopts a workflow-centered conception of research automation. We use the term \textbf{AutoResearch} to describe the broader reorganization of scientific practice in which AI is no longer confined to isolated analytical assistance, but increasingly participates in extended scientific processes involving literature grounding, ideation, experimentation, validation, reporting, and iterative continuation of research programs. More precisely, AutoResearch denotes a workflow-level paradigm of scientific inquiry in which human and AI contributions are distributed across the discovery loop under different allocations of control, execution, validation, and scientific accountability. As previewed in Figure~\ref{fig:cover_map}, this redistribution occurs across the major stages of scientific work rather than within a single isolated task. We formalize this transformation as a five-level spectrum of scientific workflow autonomy, denoted from \textcolor{RefOrange}{L0} to \textcolor{RefOrange}{L4}. These levels characterize how far AI participates in organizing, executing, validating, and closing the research workflow, rather than how frequently AI tools appear in the process.

\begin{figure*}[t]
\centering
\includegraphics[width=1\textwidth]{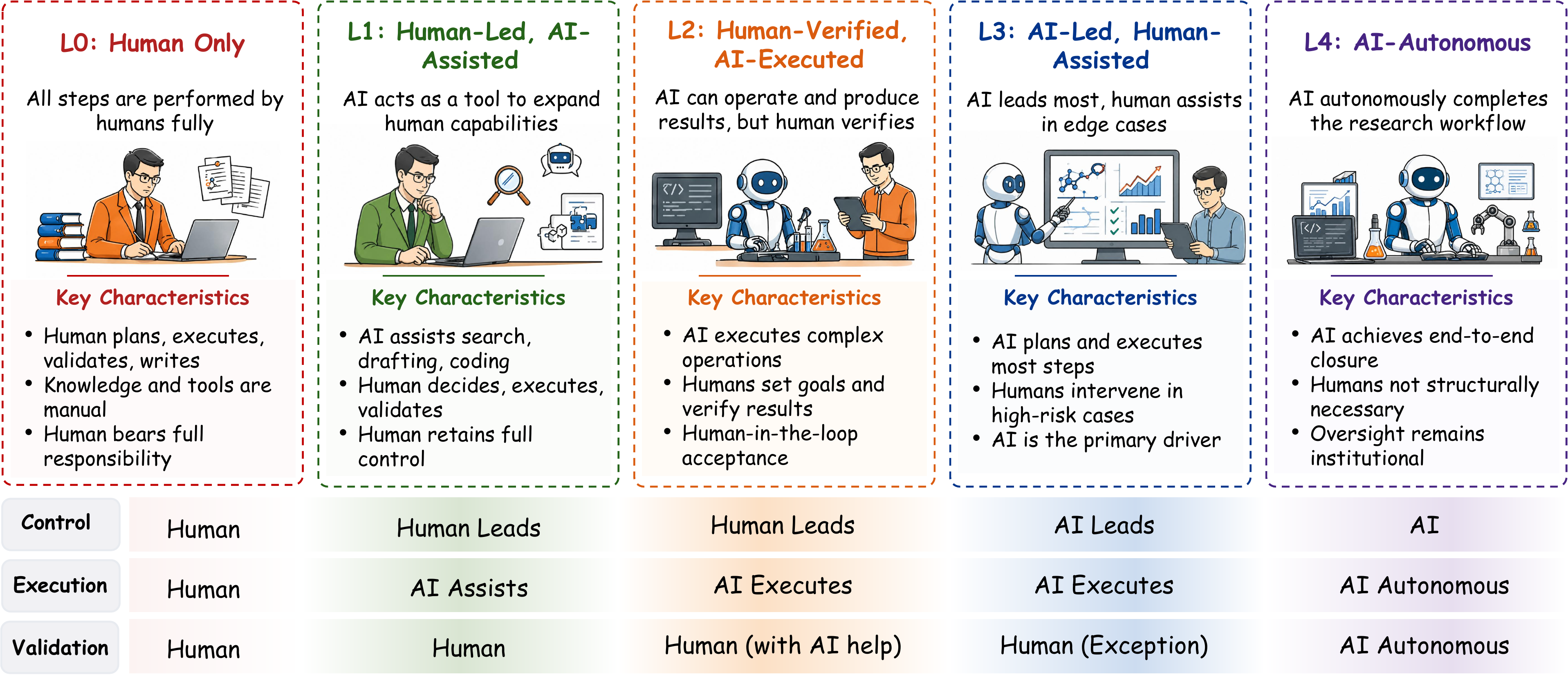}
\caption{
\textbf{Five-level autonomy spectrum of AutoResearch.}
The figure summarizes the \textcolor{RefOrange}{L0--L4} levels by comparing how workflow control, task execution, and validation authority shift from human research to AI-autonomous research. The higher levels define stricter autonomy targets rather than implying that current systems densely populate them.
}
\vspace{-20pt}
\label{fig:autonomy_levels}
\end{figure*}

Within this spectrum, \textcolor{RefOrange}{L1--L2} captures the human-steered region of AutoResearch, where bounded AI assistance and human-verified AI execution currently dominate. We refer to this region as \textbf{Vibe Research}, a practitioner-facing shorthand for workflows in which AI expands local research capability while humans retain scientific direction, verification, and accountability. \textcolor{RefOrange}{L3} marks the onset of AI-led AutoResearch, but we reserve this level for systems that can coordinate larger portions of the workflow and produce scientifically credible outputs without routine stepwise human verification. Current integrated pipelines therefore provide pressure toward \textcolor{RefOrange}{L3} rather than mature instances of it. \textcolor{RefOrange}{L4} denotes the aspirational regime in which AI can achieve routine workflow closure without humans being structurally necessary for ordinary execution, while still remaining subject to institutional oversight and scientific accountability. Figure~\ref{fig:autonomy_levels} summarizes this autonomy spectrum along four axes: workflow control, task execution, validation authority, and scientific responsibility. The levels are therefore descriptive allocations of control and responsibility, not a universal ranking of scientific desirability. The five levels can be defined as follows.

\begin{itemize}[nolistsep, leftmargin=*]

\item[] \textcolor[HTML]{F58518}{\largedot} \textit{\ul{L0: Human Only}.}
At \textbf{L0}, scientific inquiry remains human-led, human-executed, and human-verified throughout the workflow. Researchers identify problems, interpret prior work, formulate hypotheses, design and run experiments, evaluate evidence, and decide when a claim is sufficiently mature to enter the scientific record. The defining property of this level is therefore not simply that humans are present, but that scientific judgment, workflow closure, and accountability remain fully human-retained at every consequential transition. Digital tools may support local operations, but they do not redistribute scientific agency beyond the ordinary human research process. In this sense, L0 corresponds to the traditional organization of science in which criticism, validation, and acceptance remain embedded in human reasoning, disciplinary norms, and communal review~\citep{Popper1959LogicScientificDiscovery, Kuhn1962StructureScientificRevolutions}. It is this fully human-retained baseline that makes the later levels analytically meaningful~\citep{Merton1973SociologyScience}.

\item[] \textcolor[HTML]{4C78A8}{\largedot} \textit{\ul{L1: Human-Led, AI-Assisted}.}
At \textbf{L1}, the workflow remains decisively human-led, but AI becomes a routine source of bounded assistance within it. The characteristic pattern of this level is that researchers still organize the inquiry, decide what matters, and retain responsibility for all consequential judgments, while AI is used to accelerate specific cognitive tasks such as literature search, summarization, explanation, brainstorming, drafting, and lightweight analysis. What distinguishes L1 from L0 is therefore not a transfer of execution or closure, but the repeated insertion of AI as a local cognitive aid inside an otherwise human-organized workflow~\citep{Zhang2025TheEvolvingRoleofLar, Muskaan_Goyal_2025}. In practical terms, L1 is the regime most closely associated with prompt-based research assistance, where systems can be highly useful but remain tightly scoped: they inform the workflow without materially controlling it~\citep{Chen2025AI4Research}. General-purpose LLM interfaces such as GPT-4-class systems~\citep{OpenAI2024GPT4} and DeepSeek-style interfaces~\citep{DeepSeek2025DeepSeekR1} are representative of this operating mode.

\item[] \textcolor[HTML]{54A24B}{\largedot} \textit{\ul{L2: Human-Verified, AI-Executed}.}
At \textbf{L2}, AI begins to execute substantive parts of the research workflow, but the scientific authority for verification, acceptance, and accountability remains human-held. The defining transition from \textcolor{RefOrange}{L1} to \textcolor{RefOrange}{L2} is therefore not simply that AI becomes more helpful, but that it starts to perform work that would otherwise require direct human execution: reading and modifying files, generating and revising code, invoking tools, running analyses, producing intermediate artifacts, or coordinating several bounded steps inside a controllable environment. In this regime, humans no longer need to manually carry out every local operation, yet they still set the research agenda, decide whether a branch should continue, inspect whether outputs are valid, and determine whether results are reliable enough to enter the scientific workflow. This is why \textcolor{RefOrange}{L2} should be understood as \emph{human-verified AI execution}: AI can perform meaningful research labor, sometimes across multi-step or even pipeline-like workflows, but scientific closure remains dependent on human judgment. Representative examples include coding and execution substrates such as \textit{OpenHands}~\citep{OpenHandsGitHub}, \textit{Aider}~\citep{AiderGitHub}, and \textit{SWE-agent}~\citep{SWEAgentGitHub}; mixed-initiative co-research systems such as \textit{AI co-scientist}~\citep{gottweis2025towards} and \textit{FreePhD}~\citep{Li2025Build}; and integrated research pipelines such as \textit{The AI Scientist}~\citep{Lu2024AIScientist}, \textit{AI Scientist-v2}~\citep{Yamada2025AIScientistV2}, and \textit{Agent Laboratory}~\citep{AgentLaboratoryGitHub}. These systems differ in workflow span and execution capability, but they remain within \textcolor{RefOrange}{L2} when their hypotheses, methods, results, manuscripts, or deployment decisions still require human researchers to assess validity, novelty, reproducibility, usability, and final acceptance.

\item[] \textcolor[HTML]{B279A2}{\largedot} \textit{\ul{L3: AI-Led, Human-Assisted}.}
At \textbf{L3}, the research workflow begins to move from human-verified execution toward AI-led coordination. The defining property of this level is that AI does not merely perform bounded tasks or connect several modules, but starts to organize larger portions of the workflow, including grounding, planning, execution, validation, revision, and reporting. Humans remain involved, but their role shifts from routine stepwise verification toward higher-level supervision, assistance, exception handling, and intervention when the workflow becomes uncertain or scientifically insufficient. A system at this level should be able to maintain scientifically credible progress across multiple stages without requiring humans to inspect every consequential transition. Thus, the boundary between \textcolor{RefOrange}{L2} and \textcolor{RefOrange}{L3} is not determined by pipeline length alone, but by whether ordinary workflow control, branch selection, rejection, and continuation still depend on routine human verification. In this survey, \textcolor{RefOrange}{L3} is treated as the forward direction of AutoResearch and a stricter frontier for AI-led scientific workflow coordination, rather than a label assigned merely because a system implements an end-to-end research pipeline.

\item[] \textcolor[HTML]{7F7F7F}{\largedot} \textit{\ul{L4: AI-Autonomous}.}
At \textbf{L4}, AI would carry out scientific research end to end without humans being structurally necessary for routine workflow closure. This level requires more than broad automation: the system would need to formulate and continue research problems, ground hypotheses in prior work, plan and execute studies, validate results, reject weak directions, preserve provenance, and communicate findings under domain-appropriate standards of reliability and accountability. Relative to \textcolor{RefOrange}{L3}, the key difference is that human involvement is no longer required for ordinary workflow progress, although institutional oversight, governance, and post hoc audit may still remain necessary. In this survey, \textcolor{RefOrange}{L4} is therefore used as an analytical upper bound rather than as an achieved regime. Current systems remain far from this standard once rerun stability, domain validity, provenance, accountable rejection, and real scientific usefulness are taken seriously~\citep{Beel2025Evaluating, Agrawal2026Can, Luo2025More}.

\end{itemize}

Viewed through the \textcolor{RefOrange}{L0--L4} framework, the contemporary development of AutoResearch is best understood not as a uniform rise in the presence of AI, but as a selective redistribution of scientific labor across the research workflow. The pressure toward automation does not act evenly on all stages of inquiry. Literature search, drafting, coding, and certain forms of bounded tool use have proved comparatively easy to accelerate or partially externalize, whereas validation, rejection, interpretive judgment, exception handling, and accountable scientific sign-off remain markedly more resistant. Nor does this redistribution proceed in the same way across domains. Computational and formal sciences, where artifacts are machine-readable, replayable, and relatively cheap to verify, have advanced more quickly toward higher levels of workflow automation, whereas wet-lab biology, medicine, chemistry, and the social sciences remain more constrained by embodiment, experimental latency, heterogeneous evidence, and normative accountability~\citep{Tobias2025Autonomous, Gao2024Empowering, Tang2025AIResearcher, Hatakeyama_Sato_2025, Cao2024QuantumAgentSDL}. Consequently, the main empirical variation among present systems lies less in whether they have reached mature \textcolor{RefOrange}{L3}, and more in how far human-verified \textcolor{RefOrange}{L2} execution expands from local assistance to broader pipeline automation. AutoResearch therefore appears less as a single frontier and more as a layered, domain-conditioned reorganization of scientific work.

\begin{figure}[t]
\centering
\includegraphics[width=1\linewidth]{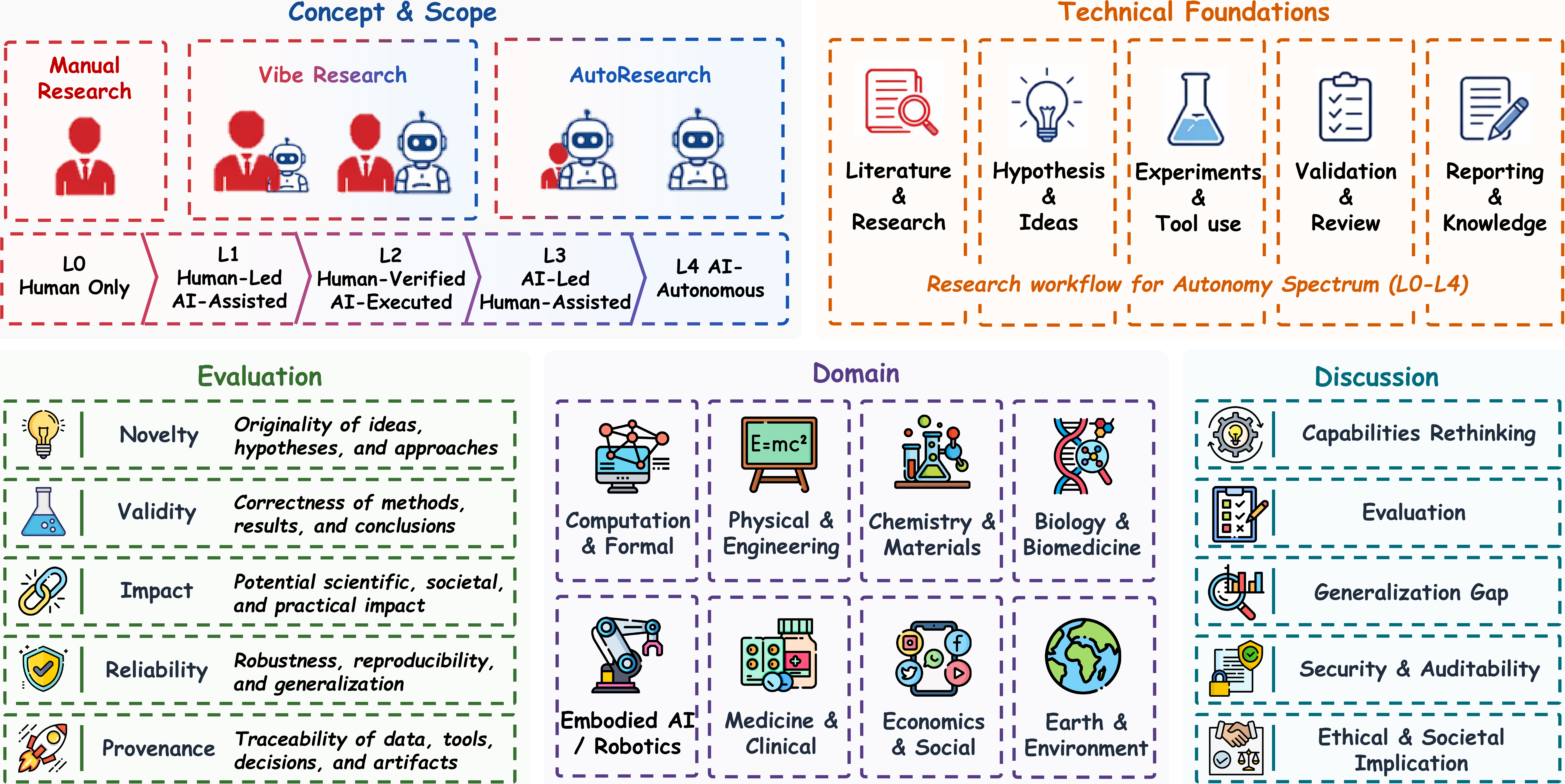}
\caption{\textbf{Overview of the AutoResearch survey framework.} The figure organizes the survey around a workflow-centered account of AutoResearch by linking five connected components: concept and scope, technical foundations, evaluation, domain-specific realizations, and broader discussion. Rather than treating research automation as a single model class or benchmark trend, it situates the field as a layered landscape spanning bounded assistance, human-verified AI execution, integrated pipeline automation, and stricter future-facing AI-led autonomy targets.}
\label{fig:survey_framework}
\vspace{-15pt}
\end{figure}

The literature has developed along the same structure. One part of the field remains centered on bounded assistance, including literature grounding, question answering, protocol planning, and related forms of prompt-based research support, and aligns most naturally with \textcolor{RefOrange}{L1}~\citep{Agarwal2024LitLLM, Vasu2025HypER, BioPlanner2023, Undermind2024ResearchAgentIterativeResear}. A second part moves into controllable environments in which AI can carry out substantial bounded work while humans retain acceptance authority, corresponding most naturally to \textcolor{RefOrange}{L2}~\citep{gottweis2025towards, Li2025Build, Shao2025OmniScientist}. A third part develops more integrated AutoResearch systems that attempt to coordinate broader spans of the discovery loop through planning, tool use, execution, analysis, reporting, and preliminary self-correction. In our taxonomy, however, these systems are best understood as advanced human-verified pipeline automation unless they can produce scientifically credible outputs without routine human verification. They therefore indicate pressure toward \textcolor{RefOrange}{L3} rather than mature occupation of it~\citep{Lu2024AIScientist, Jansen2025CodeScientist, Undermind2025AutonomousAgentsforScientifi}. Around these system lines, a growing layer of benchmarks, evaluation frameworks, and open-source infrastructures increasingly shapes how research automation is implemented, compared, and audited in practice~\citep{Chen2025Auto, Wang2025BioDSA, Liu2025ResearchBench, SPOT2025ScientificPaperErrorDetection, Gueroudji_2025, Undermind2025ResearcherBenchEvaluatingDee, Chen2025AIRSBench, KarpathyAutoresearchGitHub, ByteDanceDeerFlowGitHub, LangChainOpenDeepResearchGitHub, OpenHandsGitHub}. Existing surveys have captured important parts of this landscape, but they continue to differ in scope, unit of analysis, and underlying assumptions about autonomy~\citep{ZHENG2025Automation, Gridach2025Agentic, wei2025ai, Tie2025Survey, Chen2025AI4Research, Liu2025AVisionforAutoResear}. Figure~\ref{fig:survey_framework} organizes the remainder of this survey within that landscape by linking conceptual framing, technical foundations, evaluation, domain-specific realizations, and broader discussion into a single workflow-centered account of AutoResearch.

{\noindent\textbf{Contributions.}} Against this background, the goal of this survey is not simply to catalogue recent systems, but to provide a common framework for understanding how AI is reorganizing scientific work at the level of the research workflow. To that end, the paper makes three main contributions:

\begin{itemize}[nolistsep, leftmargin=*]

\item[] \textcolor[HTML]{4C78A8}{\largedot} \textbf{We provide a conceptual framework for AutoResearch as workflow-level scientific automation.}
We define AutoResearch as a workflow-level paradigm in which AI participates in the organization, execution, validation, and communication of scientific inquiry, rather than as a set of isolated AI-for-Science tools or standalone research agents. We introduce a five-level autonomy spectrum from \textcolor{RefOrange}{L0} to \textcolor{RefOrange}{L4}, and distinguish the human-steered \textit{Vibe Research} region of \textcolor{RefOrange}{L1--L2} from the stricter AutoResearch frontier at \textcolor{RefOrange}{L3--L4}. This framework provides a conservative vocabulary for comparing current systems by separating bounded assistance, human-verified AI execution, and pipeline automation from mature AI-led scientific autonomy. It also helps avoid equating broader workflow coverage with reliable autonomous research closure.

\item[] \textcolor[HTML]{F58518}{\largedot} \textbf{We develop a technical taxonomy of AutoResearch around five workflow conditions.}
We organize the technical foundations of AutoResearch around five recurring workflow conditions: literature and research grounding, hypothesis formation and planning, experimentation and tool use, feedback, validation, and review, and reporting and knowledge communication. This taxonomy clarifies how current systems redistribute scientific work across the research workflow, from evidence construction and idea selection to execution, rejection, revision, and artifact generation. It further shows that stronger automation requires not only capable modules, but also durable coupling among evidence, plans, environments, validation mechanisms, and communicable research artifacts. Through this view, different research agents, AI scientist systems, and workflow infrastructures can be compared within a common technical frame.

\item[] \textcolor[HTML]{54A24B}{\largedot} \textbf{We synthesize evaluation principles and domain-conditioned autonomy limits for AutoResearch.}
We organize AutoResearch evaluation around five dimensions of workflow-level scientific credibility: novelty, validity, impact, reliability, and provenance. These dimensions shift attention from whether a system can complete a task to whether its research outputs are original, correct, useful, reproducible, and traceable across the workflow. We further analyze how autonomy ceilings differ across domains, showing that stronger automation is currently more credible in executable and auditable settings, such as computational and formal sciences, while embodied, delayed, heterogeneous, or high-stakes domains remain more constrained by validation, safety, uncertainty, and accountability requirements. This domain-conditioned perspective explains why progress toward autonomous research is uneven rather than uniform across science.

\end{itemize}

\textbf{Paper Organization.}
The remainder of this survey is organized as follows. Section~\ref{sec:history_scope} introduces AutoResearch from a historical and conceptual perspective, clarifying its scope, boundaries, and relationship to adjacent strands of AI-for-Science and research automation. Section~\ref{sec:technical_foundations} examines its technical foundations through the major components of the scientific discovery loop, including literature grounding, hypothesis generation and planning, experimentation and tool use, validation, and reporting. Section~\ref{sec:evaluation_frameworks} develops a unified evaluative perspective centered on novelty, validity, impact, reliability, and provenance, and situates current benchmarks, audit instruments, and evaluation practices within that framework. Section~\ref{sec:domain} analyzes how the practical ceiling of AutoResearch differs across major scientific domains, highlighting why workflow portability remains limited in practice. Finally, Section~\ref{sec:challenges_governance} discusses capability boundaries, evaluation gaps, domain generalization limits, reliability, auditability, and the ethical and societal implications of AutoResearch.

\section{Overview of AutoResearch}
\label{sec:history_scope}

\begin{figure*}[t]
\centering
\includegraphics[width=0.94\linewidth]{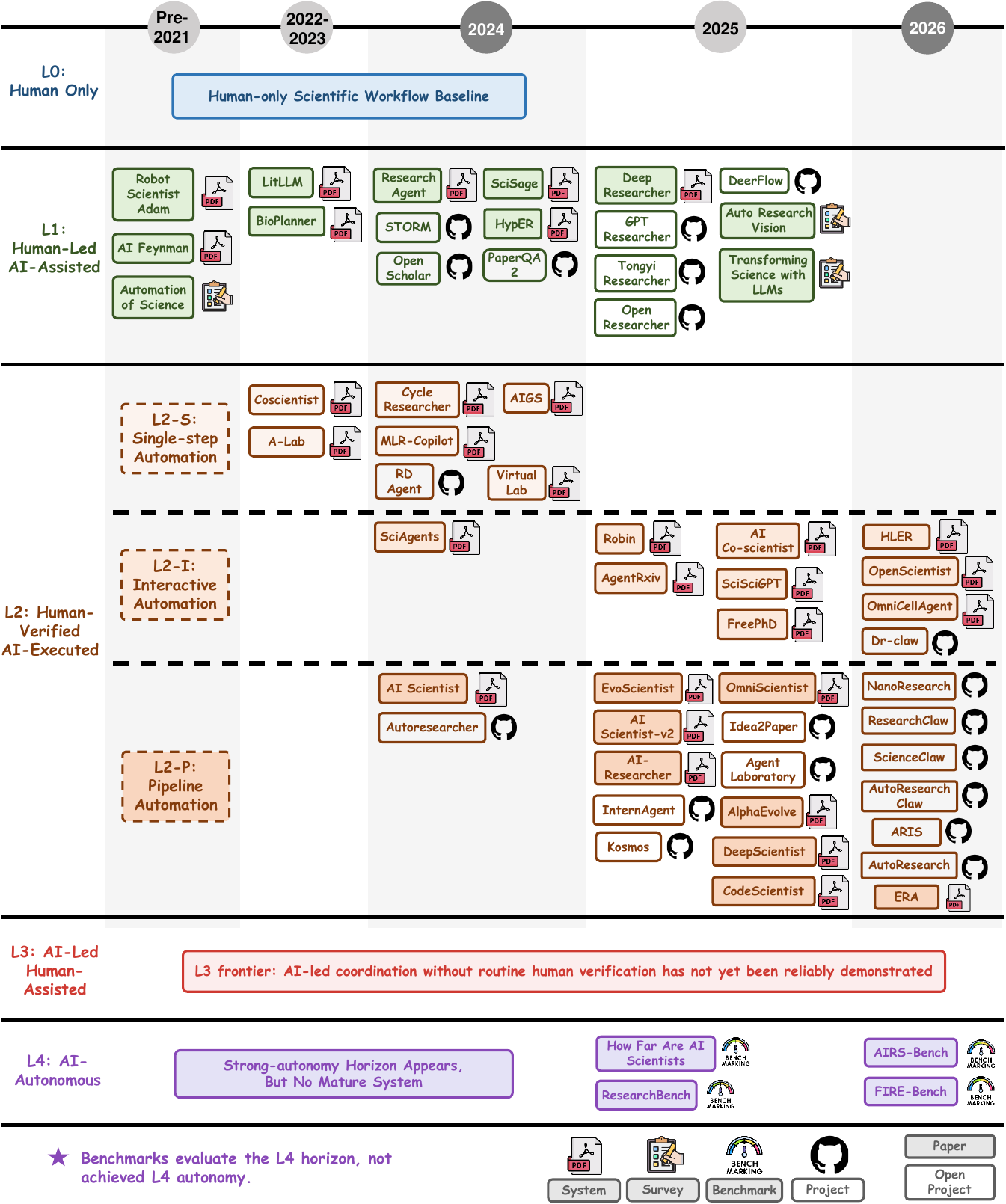}
\caption{
\textbf{Historical overview of AutoResearch.} The figure maps representative works, systems, benchmarks, and open-source infrastructures onto the L0–L4 autonomy spectrum, with L2 further divided into single-step execution, interactive workflow automation, and pipeline automation under human verification.
}
\label{fig:historical_lineage}
\vspace{-20pt}
\end{figure*}

Scientific work has become progressively more digital, instrumented, and software-mediated over the past decades, making larger portions of the research process searchable, executable, and open to partial automation~\citep{Kramer2023Automated}. Recent surveys and positioning papers increasingly characterize this shift not as the growth of isolated task tools, but as a broader reorganization of scientific workflows, research lifecycles, and agentic systems~\citep{ZHENG2025Automation, Zheng2025Agent4S, Chen2025AI4Research}. The contemporary AutoResearch landscape emerged within this transition, as advances in language models, scientific agents, and software-native research environments made bounded assistance, controllable execution, and longer-horizon workflow coordination increasingly operational~\citep{Tie2025Survey, Liu2025AVisionforAutoResear, Gridach2025Agentic}.
A central difficulty in mapping this landscape is that pipeline breadth can be mistaken for scientific autonomy. Many recent systems connect literature grounding, ideation, coding, experimentation, analysis, and writing, but their outputs still require human researchers to judge validity, novelty, usability, and acceptance. We therefore adopt a conservative placement rule: systems are assigned to the lowest autonomy regime consistent with their demonstrated workflow role, and integrated pipelines remain within \textcolor{RefOrange}{L2} when routine human verification is still structurally necessary. To make this distinction visible, this section further refines \textcolor{RefOrange}{L2} into single-step automated execution, interactive workflow automation, and pipeline automation under human verification. Figure~\ref{fig:historical_lineage} provides the historical scaffold for this account.

\subsection{History of AutoResearch}
\label{sec:history_symbolic}

The historical development of AutoResearch is most clearly visible in the gradual restructuring of the scientific workflow itself. Different parts of research became formalized, executable, and connectable at different times, allowing assistance, execution, coordination, and partial closure to accumulate unevenly across the discovery process~\citep{Kramer2023Automated, ZHENG2025Automation}. This trajectory is reflected in the maturation of workflow-centered views of research automation~\citep{Chen2025AI4Research}, the rise of agentic scientific systems~\citep{Gridach2025Agentic, Tie2025Survey}, and the appearance of longer-horizon research pipelines that couple literature work, planning, execution, and reporting inside shared operational loops~\citep{Liu2025AVisionforAutoResear, wei2025ai, Lu2024AIScientist}. The history below therefore focuses on how research automation expanded from human-centered scientific practice to knowledge-work assistance, bounded execution, interactive workflows, integrated human-verified pipelines, and finally to stricter autonomy frontiers.

\begin{itemize}[nolistsep, leftmargin=*]

\item[] \textcolor{HumanOrange}{\largedot} \textbf{Human-centered scientific practice as the baseline.}
Before research automation became a technical agenda, scientific inquiry was organized around human problem formulation, literature interpretation, hypothesis construction, experimental design, evidential assessment, and community-facing reporting. Classical accounts of science framed this regime through conjecture and criticism, paradigm-guided inquiry and rupture, and communal norms that stabilize knowledge claims~\citep{Popper1959LogicScientificDiscovery, Kuhn1962StructureScientificRevolutions, Merton1973SociologyScience}. The postwar expansion of scientific communication enlarged the scale of publication, collaboration, and institutional review, but did not redistribute scientific closure away from human researchers and research communities~\citep{deSollaPrice1963LittleScienceBigScience}. In the timeline, this appears as the human-only scientific workflow baseline: a reference point against which later automation changes the allocation of search, execution, validation, and reporting without immediately replacing scientific judgment.

\item[] \textcolor{CoResearchGreen}{\largedot} \textbf{Assistance, field framing, and knowledge-work automation.}
The first durable layer of AutoResearch emerged when scientific knowledge work became searchable, synthesizable, and partially formalizable. Early anchors such as \textit{Robot Scientist Adam}~\citep{King2004RobotScientistAdam} and \textit{AI Feynman}~\citep{Udrescu2020AIPhysRev} showed that selected components of discovery could be automated in structured settings, such as automated hypothesis testing, symbolic recovery, or reasoning over constrained scientific spaces, while broader discussions on the \textit{Automation of Science}~\citep{Kramer2023Automated} framed automation as a workflow-level question. With the rise of language models, systems such as \textit{BioPlanner}~\citep{BioPlanner2023} and \textit{LitLLM}~\citep{Agarwal2024LitLLM} moved this layer toward protocol reasoning and literature-centered research support. In 2024, retrieval- and synthesis-oriented systems including \textit{Research Agent}~\citep{Undermind2024ResearchAgentIterativeResear}, \textit{STORM}~\citep{StanfordStormGitHub}, \textit{OpenScholar}~\citep{OpenScholarGitHub}, \textit{SciSage}~\citep{Shi2025SciSage}, \textit{HypER}~\citep{Vasu2025HypER}, and \textit{PaperQA2}~\citep{PaperQA2_2024} made grounded search, multi-perspective synthesis, hypothesis support, and paper-grounded answering part of stable assistant workflows. \textit{STORM}, for example, is explicitly a retrieval- and multi-perspective-questioning system for grounded long-form writing rather than an executional research agent. By 2025--2026, \textit{Deep Research Arena}~\citep{Wan2025DeepResearch}, \textit{GPT Researcher}~\citep{GPTResearcherGitHub}, \textit{Tongyi Researcher}~\citep{AlibabaDeepResearchGitHub}, \textit{Open Researcher}~\citep{LangChainOpenDeepResearchGitHub}, and \textit{DeerFlow}~\citep{ByteDanceDeerFlowGitHub}, together with field-framing works such as \textit{Auto Research Vision}~\citep{Liu2025AVisionforAutoResear} and \textit{Transforming Science with LLMs}~\citep{wei2025ai}, consolidated AI as a recurrent cognitive and organizational layer in scientific work. Historically, this \textcolor{RefOrange}{L1} layer improves research throughput and organization, but it does not transfer executional authority or scientific closure away from human researchers.

\item[] \textcolor{HumanOrange}{\largedot} \textbf{L2-S: Single-step automated execution.}
The next historical transition occurred when AI systems began to execute bounded scientific operations rather than only support knowledge work. We describe this regime as \textcolor{RefOrange}{L2-S}, or single-step automated execution. Systems in this layer can carry out well-specified operations such as tool invocation, code execution, protocol enactment, laboratory control, model training, data-driven analysis, or bounded experimental support. \textit{Coscientist}~\citep{Boiko2023Autonomous} connected language models to search, code execution, laboratory documentation, and experimental automation in chemistry, while \textit{A-Lab}~\citep{Szymanski2023AutonomousLab} demonstrated autonomous materials synthesis through computation, historical or literature-derived knowledge, active learning, and robotic execution. Both works expanded executional capacity, but in controlled scientific domains rather than in general research workflows. In 2024, \textit{CycleResearcher}~\citep{Weng2024CycleResearcher}, \textit{MLR-Copilot}~\citep{Li2024MLR}, \textit{RD Agent}~\citep{RDAgent}, \textit{AIGS}~\citep{Liu2024AIGS}, and \textit{Virtual Lab}~\citep{virtualLab} extended bounded execution into planning, implementation, revision, and virtual experimental environments. The defining property of this layer is not full workflow autonomy, but the transfer of selected executable tasks from humans to AI under constrained goals, controlled settings, and external verification.

\item[] \textcolor{HumanOrange}{\largedot} \textbf{L2-I: Interactive workflow automation.}
A second \textcolor{RefOrange}{L2} pattern emerged as systems began to support multi-step workflows through interaction, feedback, and mixed-initiative control. We call this regime \textcolor{RefOrange}{L2-I}, or interactive workflow automation. Unlike \textcolor{RefOrange}{L2-S} systems, which execute bounded operations, \textcolor{RefOrange}{L2-I} systems help maintain progress across several research actions while relying on human feedback, steering, or acceptance. \textit{SciAgents}~\citep{Ghafarollahi2024SciAgents} extended execution from simple tool use toward multi-agent reasoning and scientific ideation over structured scientific representations. The 2025--2026 wave broadened this layer further: \textit{AI co-scientist}~\citep{gottweis2025towards}, \textit{SciSciGPT}~\citep{Shao2025SciSciGPT}, \textit{FreePhD}~\citep{Li2025Build}, \textit{Robin}~\citep{ghareeb2026robin}, and \textit{AgentRxiv}~\citep{Schmidgall2025AgentRxiv} pressed toward stronger mixed-initiative co-research through collaborative ideation, feedback, code or data-driven experimentation, paper generation, and research production. \textit{HLER}~\citep{Undermind2026HLERHumanintheLoopEconomicRe}, \textit{AI co-scientists for statistical genetics}~\citep{gottweis2025towards}, and \textit{Dr-claw}~\citep{song2026drclaw} further extended this pattern into economics, genetics, biomedical analysis, cellular research, and project-level assistance. Recent Nature publications further strengthened the visibility of mixed-initiative scientific discovery workflows, particularly through \textit{Co-scientist}~\citep{gottweis2026coscientist}, which demonstrated literature-grounded multi-agent hypothesis generation and collaborative scientific reasoning. 

\item[] \textcolor{AIAutoBlue}{\largedot} \textbf{L2-P: Pipeline automation under human verification.}
The strongest currently populated layer is \textcolor{RefOrange}{L2-P}: pipeline automation under human verification. Systems in this regime connect multiple research stages---such as literature grounding, ideation, implementation, experimentation, analysis, review, and writing---inside a longer operational loop. \textit{The AI Scientist}~\citep{Lu2024AIScientist} made this direction especially visible by coupling idea generation, code writing, experiment execution, figure production, paper drafting, and simulated review into a single end-to-end research framework, alongside early work on \textit{Autoresearcher}~\citep{KarpathyAutoresearchGitHub}. In 2025, \textit{Idea2Paper}~\citep{Idea2PaperGitHub}, \textit{Agent Laboratory}~\citep{AgentLaboratoryGitHub}, \textit{AlphaEvolve}~\citep{Novikov2025AlphaEvolve}, \textit{DeepScientist}~\citep{Weng2025DeepScientist}, \textit{CodeScientist}~\citep{Jansen2025CodeScientist}, and \textit{OmniScientist}~\citep{Shao2025OmniScientist} further expanded this pipeline view through paper generation, coding, experiment management, multi-agent ecosystems, or longer-horizon research production. \textit{AI Scientist-v2}~\citep{Yamada2025AIScientistV2}, \textit{AI-Researcher}~\citep{Tang2025AIResearcher}, \textit{InternAgent}~\citep{feng2026internagent}, and \textit{Kosmos}~\citep{mitchener2025kosmos} strengthened this frontier through agentic search, experiment management, persistent research state, and tighter coupling between literature, hypothesis generation, data analysis, and scientific reporting. By 2026, open infrastructures such as \textit{NanoResearch}~\citep{nanoresearch2026}, \textit{ResearchClaw}~\citep{ResearchClawGitHub}, \textit{ScienceClaw}~\citep{ScienceClawGitHub}, \textit{AutoResearchClaw}~\citep{liu2026autoresearchclaw}, \textit{ARIS}~\citep{ARISGitHub}, and \textit{EvoScientist}~\citep{Lyu2026EvoScientist} further shifted the field from isolated research-agent demonstrations toward reusable workspaces, tool-rich orchestration, persistent project state, and research-pipeline infrastructure. \textit{NeuroClaw}~\citep{wang2026neuroclaw} further extended this direction toward neuroscience-oriented research orchestration and agentic experimental workflows built on persistent scientific workspaces. Recent work such as \textit{Robin}~\citep{ghareeb2026robin} further expanded AutoResearch toward iterative scientific discovery pipelines that couple hypothesis generation, experimental analysis, and literature-guided workflow refinement inside semi-autonomous research loops. \textit{Empirical Research Assistance (ERA)}~\citep{aygun2026expertsoftware} further highlighted implementation-centric AutoResearch by using LLM-guided tree-search optimization to generate expert-level empirical scientific software across multiple computational domains.

The conservative placement of these systems is analytically important. Although they may coordinate broad spans of the research loop, they still depend on human researchers to assess whether generated hypotheses are meaningful, whether experiments are valid, whether results are reproducible, and whether manuscripts are scientifically usable. They therefore create pressure toward \textcolor{RefOrange}{L3}, but they are best classified as advanced \textcolor{RefOrange}{L2-P} unless routine human verification is no longer structurally necessary. Historically, this layer forms the bridge between AI as an acting component of scientific work and the stricter frontier of AI-led research coordination.

\item[] \textcolor{ScopePurple}{\largedot} \textbf{Autonomous closure as a benchmarked horizon.}
The final layer is not a densely populated system category, but a frontier of evaluation. \textcolor{RefOrange}{L3} remains the point at which AI-led research would require more than pipeline breadth: the system would need to coordinate larger portions of the workflow and produce scientifically credible intermediate and final outputs without routine stepwise human verification. Current systems show partial pressure toward this condition, but robust evidence for mature \textcolor{RefOrange}{L3} remains limited. \textcolor{RefOrange}{L4} is still further away, requiring autonomous scientific closure with reliable rejection, validation, provenance, reproducibility, and accountability. The timeline deliberately separates mature systems from the aspirational horizon: no current system is treated as a robust instance of fully autonomous scientific closure. Instead, the recent benchmark layer measures how far existing agents remain from that horizon. \textit{How Far Are AI Scientists from Changing the World?}~\citep{Xie2025How} sharpened the field's bottleneck analysis by foregrounding the gap between system ambition and scientific impact. \textit{ResearchBench}~\citep{Liu2025ResearchBench} reframed scientific discovery as a decomposable benchmark problem, while \textit{AIRS-Bench}~\citep{Chen2025AIRSBench} and \textit{FIRE-Bench}~\citep{Undermind2026FIREBenchEvaluatingAgentsont} pushed evaluation toward frontier research agents and full-cycle rediscovery tasks. This phase is historically important because the field is no longer defined only by increasingly capable systems; it is also defined by increasingly explicit tests of workflow closure, implementation reliability, evidence quality, and scientific reasoning. The resulting picture is asymmetric: autonomous discovery can be demonstrated in bounded settings and measured with sharper benchmarks, but stable internalization of domain-grounded validation, accountable acceptance, rejection, and trustworthy workflow closure remains unresolved.

\end{itemize}

\begin{table*}[ht]
\centering
\surveytablecaption{Representative works in the contemporary AutoResearch landscape.}{Selected surveys, workflow-level systems, and open-source projects are grouped by their structural role in the field. The level column indicates the primary autonomy regime suggested by each work's workflow scope, execution capability, and degree of human oversight, rather than a universal performance ranking.}
\label{tab:autoresearch_landscape_overview}
\scriptsize
\renewcommand{\arraystretch}{1.0}
\setlength{\tabcolsep}{3.8pt}
\rowcolors{2}{white}{black!2}
\begin{adjustbox}{width=1\textwidth}
\begin{tabular}{@{}p{4cm} p{1.5cm} p{1cm} p{4.4cm} p{1cm}@{}}
\toprule[1pt]
\textbf{Work} & \textbf{Type} & \textbf{Level} & \textbf{Workflow Emphasis} & \textbf{Date} \\
\midrule

\rowcolor{surveyblue!10}
\multicolumn{5}{@{}c}{\textit{Surveys and Positioning Papers}} \\
Automated Scientific Discovery~\cite{Kramer2023Automated} & \roleSurvey & -- & Historical and conceptual framing & 2023 \\
Workflow (R)evolution~\cite{ZHENG2025Automation} & \roleSurvey & -- & Workflow-level research automation & 2025 \\
Agent4S~\cite{Zheng2025Agent4S} & \roleSurvey & -- & Agentic science taxonomy & 2025 \\
Agentic AI Scientist~\cite{Gridach2025Agentic} & \roleSurvey & -- & AI scientist system taxonomy & 2025 \\
AI4Research~\cite{Chen2025AI4Research} & \roleSurvey & -- & AI-for-research ecosystem & 2025 \\
Survey of AI Scientists~\cite{Tie2025Survey} & \roleSurvey & -- & AI scientist landscape synthesis & 2025 \\
Vision for Auto Research~\cite{Liu2025AVisionforAutoResear} & \roleSurvey & -- & AutoResearch agenda and vision & 2025 \\
LLM4SR~\cite{luo2025llm4sr} & \roleSurvey & -- & LLM-assisted scientific research & 2025 \\

\cmidrule(l){1-5}
\rowcolor{surveyorange!10}
\multicolumn{5}{@{}c}{\textit{Workflow-Level Systems}} \\
BioPlanner~\cite{BioPlanner2023} & \roleSystem & \textcolor{RefOrange}{L1} & Protocol planning support & 2023 \\
Coscientist~\cite{Boiko2023Autonomous} & \roleSystem & \textcolor{RefOrange}{L2-S} & Chemistry tool-use execution & 2023 \\
LLM-driven Research~\cite{Ifargan2024Autonomous} & \roleSystem & \textcolor{RefOrange}{L2-P} & Data-to-paper pipeline & 2024 \\
SciAgents~\cite{Ghafarollahi2024SciAgents} & \roleSystem & \textcolor{RefOrange}{L2-I} & Multi-agent scientific reasoning & 2024 \\
CycleResearcher~\cite{Weng2024CycleResearcher} & \roleSystem & \textcolor{RefOrange}{L2-S} & Planning--revision execution loop & 2024 \\
The AI Scientist~\cite{Lu2024AIScientist} & \roleSystem & \textcolor{RefOrange}{L2-P} & Ideation-to-paper pipeline & 2024 \\
AgentRxiv~\cite{Schmidgall2025AgentRxiv} & \roleSystem & \textcolor{RefOrange}{L2-I} & Collaborative research workflow & 2025 \\
FreePhD~\cite{Li2025Build} & \roleSystem & \textcolor{RefOrange}{L2-I} & Human-verified research execution & 2025 \\
AI-Researcher~\cite{Tang2025AIResearcher} & \roleSystem & \textcolor{RefOrange}{L2-P} & Literature-to-report pipeline & 2025 \\
OmniScientist~\cite{Shao2025OmniScientist} & \roleSystem & \textcolor{RefOrange}{L2-P} & Multi-agent research ecosystem & 2025 \\
AI Scientist-v2~\cite{Yamada2025AIScientistV2} & \roleSystem & \textcolor{RefOrange}{L2-P} & Long-horizon research pipeline & 2025 \\
CodeScientist~\cite{Jansen2025CodeScientist} & \roleSystem & \textcolor{RefOrange}{L2-P} & Code-centric experiment pipeline & 2025 \\
DeepScientist~\cite{Weng2025DeepScientist} & \roleSystem & \textcolor{RefOrange}{L2-P} & Iterative discovery pipeline & 2025 \\
Rethinking the AI Scientist~\cite{weidener2026rethinking} & \roleSystem & \textcolor{RefOrange}{L2-I} & Interactive research-agent loop & 2026 \\
EvoScientist~\cite{Lyu2026EvoScientist} & \roleSystem & \textcolor{RefOrange}{L2-P} & Evolutionary research pipeline & 2026 \\
Co-Scientist~\cite{gottweis2026coscientist} & \roleSystem & \textcolor{RefOrange}{L2-I} & Multi-agent co-scientific discovery workflow & 2026 \\
Robin~\cite{ghareeb2026robin} & \roleSystem & \textcolor{RefOrange}{L2-P} & Multi-agent scientific discovery workflow & 2026 \\
ERA~\cite{aygun2026expertsoftware} & \roleSystem & \textcolor{RefOrange}{L2-P} & Empirical scientific software generation & 2026 \\

\cmidrule(l){1-5}
\rowcolor{surveygreen!10}
\multicolumn{5}{@{}c}{\textit{Open-Source Projects and Infrastructures}} \\
STORM~\cite{StanfordStormGitHub} & \roleProject & \textcolor{RefOrange}{L1} & Retrieval-grounded synthesis & 2024 \\
GPT Researcher~\cite{GPTResearcherGitHub} & \roleProject & \textcolor{RefOrange}{L1} & Deep-research report generation & 2025 \\
OpenScholar~\cite{OpenScholarGitHub} & \roleProject & \textcolor{RefOrange}{L1} & Literature-grounded QA & 2025 \\
PaperQA2~\cite{PaperQA2GitHub} & \roleProject & \textcolor{RefOrange}{L1} & Paper-grounded QA & 2025 \\
Open Deep Research~\cite{LangChainOpenDeepResearchGitHub} & \roleProject & \textcolor{RefOrange}{L1} & Recursive literature research & 2025 \\
DeerFlow~\cite{ByteDanceDeerFlowGitHub} & \roleProject & \textcolor{RefOrange}{L1} & Deep-research workflow & 2025 \\
OpenHands~\cite{OpenHandsGitHub} & \roleProject & \textcolor{RefOrange}{L2-S} & Software execution substrate & 2025 \\
Aider~\cite{AiderGitHub} & \roleProject & \textcolor{RefOrange}{L2-S} & Code-editing execution & 2025 \\
SWE-agent~\cite{SWEAgentGitHub} & \roleProject & \textcolor{RefOrange}{L2-S} & Repository-level execution & 2025 \\
autoresearch~\cite{KarpathyAutoresearchGitHub} & \roleProject & \textcolor{RefOrange}{L2-P} & Steerable research pipeline & 2025 \\
Agent Laboratory~\cite{AgentLaboratoryGitHub} & \roleProject & \textcolor{RefOrange}{L2-P} & Open research-agent pipeline & 2025 \\
ResearchClaw~\cite{ResearchClawGitHub} & \roleProject & \textcolor{RefOrange}{L2-P} & Research workflow stack & 2026 \\
ScienceClaw~\cite{ScienceClawGitHub} & \roleProject & \textcolor{RefOrange}{L2-P} & Scientific-agent workflow stack & 2026 \\
NanoResearch~\cite{nanoresearch2026} & \roleProject & \textcolor{RefOrange}{L2-P} & Integrated research pipeline & 2026 \\
AutoResearchClaw~\cite{liu2026autoresearchclaw} & \roleProject & \textcolor{RefOrange}{L2-P} & Multi-stage research orchestration & 2026 \\
ARIS~\cite{ARISGitHub} & \roleProject & \textcolor{RefOrange}{L2-P} & Research orchestration infrastructure & 2026 \\

\bottomrule[1pt]
\end{tabular}
\end{adjustbox}
\vspace{-8pt}
\end{table*}

\subsection{Contemporary Landscape of AutoResearch}
\label{sec:contemporary_landscape}

The contemporary AutoResearch landscape is organized less by a single canonical architecture than by a functional division of labor. One layer stabilizes knowledge support through literature grounding, source-grounded synthesis, question answering, planning, and report construction; systems such as \textit{STORM} and \textit{OpenScholar} exemplify this layer by making retrieval-augmented knowledge work more structured and persistent~\citep{StanfordStormGitHub, OpenScholarGitHub}. A second layer provides execution substrates, including code agents, tool use, laboratory interfaces, controlled environments, and software-agent execution~\citep{OpenHandsGitHub, AiderGitHub, SWEAgentGitHub}. A third layer connects these capabilities into longer research pipelines that span ideation, implementation, experimentation, analysis, paper generation, and review-style feedback; \textit{The AI Scientist} and \textit{AI Scientist-v2} are representative because they make the code-native end-to-end research loop explicit~\citep{Lu2024AIScientist, Yamada2025AIScientistV2}. Open-source projects and infrastructures provide the operational substrate around these systems, including software-agent execution, tool orchestration, persistent workspaces, and reusable research environments~\citep{StanfordStormGitHub, OpenHandsGitHub, AgentLaboratoryGitHub, liu2026autoresearchclaw}. The field therefore advances through the joint maturation of assistance, execution, interaction, and pipeline orchestration.

Table~\ref{tab:autoresearch_landscape_overview} selects works that are structurally consequential for this landscape rather than attempting an exhaustive catalog. The inclusion criterion is workflow relevance: each entry either clarifies the conceptual boundary of AutoResearch, introduces a reusable research-support or execution substrate, implements a multi-stage research process, provides open infrastructure for grounding, execution, orchestration, or reporting, or evaluates progress toward autonomous scientific closure. This selection makes the table a structural map of the field rather than a chronological bibliography. The primary-regime column follows the same conservative placement rule as Figure~\ref{fig:historical_lineage}. \textcolor{RefOrange}{L1} denotes systems that expand human research capability without transferring executional or validation authority. \textcolor{RefOrange}{L2-S} denotes bounded execution of well-specified scientific tasks. \textcolor{RefOrange}{L2-I} denotes interactive or mixed-initiative workflows in which AI participates across multiple steps but remains dependent on human steering or feedback. \textcolor{RefOrange}{L2-P} denotes integrated research pipelines that connect multiple stages but still require human verification of scientific validity, novelty, reproducibility, or usability. \textcolor{RefOrange}{L3} is reserved for AI-led research workflows that no longer require routine stepwise human verification, and \textcolor{RefOrange}{L4} remains an autonomy horizon rather than a populated practical category.

Under this rule, the current landscape is concentrated in \textcolor{RefOrange}{L1} and \textcolor{RefOrange}{L2}, with the most rapid expansion occurring inside \textcolor{RefOrange}{L2}. Literature-grounded assistants and deep-research systems populate \textcolor{RefOrange}{L1}; tool-using agents and bounded experimental systems populate \textcolor{RefOrange}{L2-S}; mixed-initiative co-research systems populate \textcolor{RefOrange}{L2-I}; and recent AI scientist systems and research infrastructures largely populate \textcolor{RefOrange}{L2-P}. This distribution does not weaken their significance. Rather, it clarifies the central empirical story of current AutoResearch: the field is moving from local assistance toward integrated human-verified pipeline automation, while mature AI-led autonomy remains a stricter frontier.

This distribution reflects a structural filtering of autonomy rather than a simple gap in model scale. AutoResearch advances fastest where workflow segments are modular, outputs are judgeable, environments are programmable, and feedback is sufficiently rapid to support iteration. It slows where scientific progress depends on long-horizon coordination, domain-specific interpretation, costly validation, heterogeneous evidence, rejection, provenance, or institutional accountability~\citep{Chen2025AI4Research, Zheng2025Agent4S, Chen2025AIRSBench, Xie2025How}. The contemporary field is therefore best read as a layered ecosystem: conceptual frameworks define the space, assistance systems stabilize knowledge work, execution systems expand what AI can do inside research workflows, and pipeline systems test how far these components can be connected into longer human-verified research loops. 
Section~\ref{sec:technical_foundations} examines the technical foundations that make higher levels operational, Section~\ref{sec:evaluation} analyzes evaluative frameworks for distinguishing stronger systems from stronger claims, and Section~\ref{sec:domain} studies the domain conditions that determine where scientific autonomy can be credibly sustained.
\section{Technical Foundations of AutoResearch}
\label{sec:technical_foundations}

The technical foundations of AutoResearch are best understood as workflow conditions that constrain how scientific activity becomes grounded, executable, revisable, and communicable. Contemporary systems increasingly combine retrieval, planning, tool use, experimentation, validation, and reporting, but their autonomy depends less on the presence of these modules than on how reliably they are coupled across stages of inquiry. This section therefore examines the five technical conditions through which bounded assistance, human-verified execution, and increasingly integrated pipeline automation become operational: literature grounding, hypothesis formation and planning, experimentation and tool use, feedback, validation, and review, and reporting and knowledge communication.


\begin{figure*}[h]
\centering
\includegraphics[width=1\linewidth]{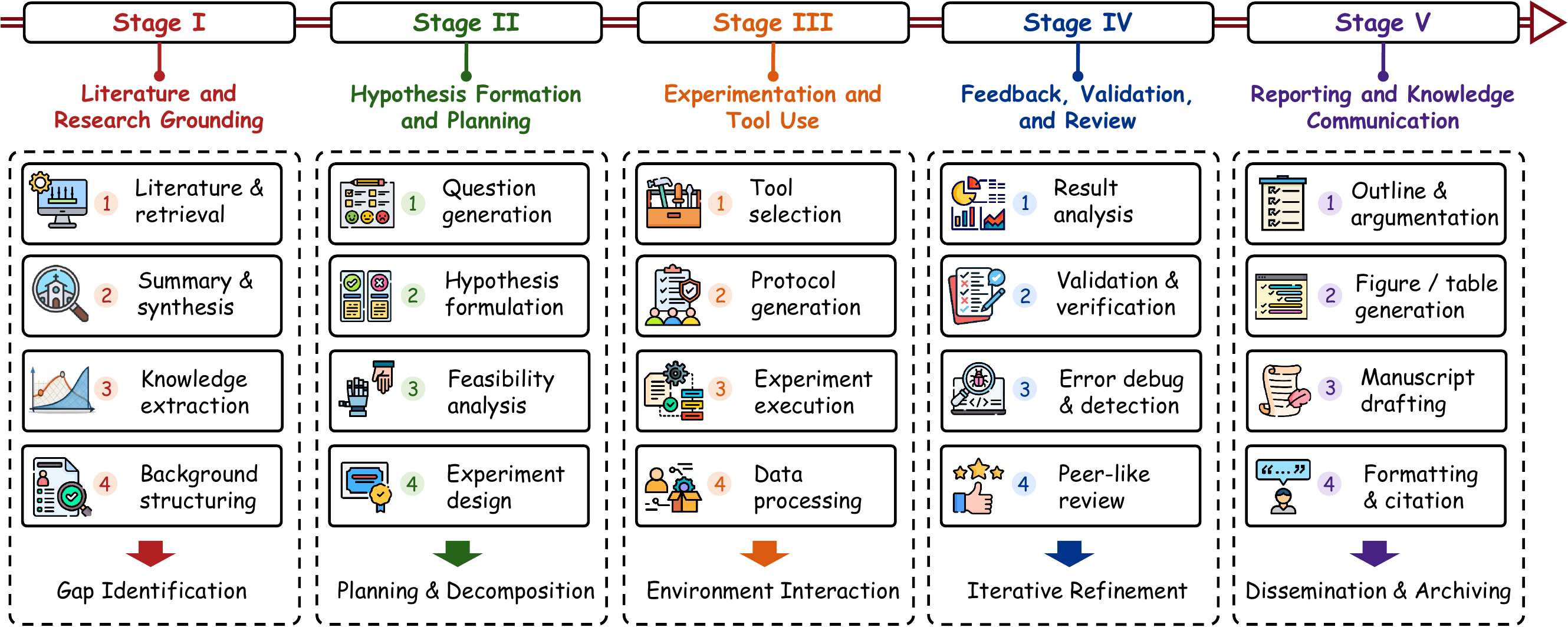}
\caption{\textbf{Technical workflow stages of AutoResearch.} The figure decomposes AutoResearch into five recurring stages, from literature grounding and hypothesis planning to experimentation, validation, and reporting. Each stage lists representative technical operations and the corresponding workflow function.}
\label{fig:technical_paradigms_map}
\vspace{-15pt}
\end{figure*}

\subsection{Workflow Conditions for Scientific Autonomy}
\label{sec:workflow_conditions}

Scientific autonomy depends on how research capabilities are organized across the workflow, not on the presence of individual components alone. Retrieval, planning, tool use, experimentation, validation, and reporting now appear in many research-oriented systems, but their contribution to autonomy depends on whether they support grounded reasoning, executable action, revision, rejection, and accountable communication across stages. Figure~\ref{fig:technical_paradigms_map} provides the workflow view adopted in this section. We therefore analyze AutoResearch through five technical conditions of scientific work—literature grounding; hypothesis formation and planning; experimentation and tool use; feedback, validation, and review; and reporting and knowledge communication—rather than through system-level labels such as single-agent, multi-agent, retrieval-augmented, or tool-using architectures.

Each condition supplies a different form of scientific constraint. Grounding constrains reasoning by evidence; planning constrains exploration by feasibility and comparison; execution constrains hypotheses through environments; validation constrains outputs through rejection pressure; and reporting constrains communication through provenance and artifact alignment. Together, these conditions determine whether a system remains a local assistant, becomes a human-verified executor, or supports broader pipeline-level coordination under human verification.

\begin{itemize}[nolistsep, leftmargin=*]

\item[] \textcolor[HTML]{4C78A8}{\largedot} \textit{\ul{Literature and Research Grounding.}}
Research automation depends first on how prior work is retrieved, filtered, interpreted, organized, and reused within the workflow. This stage includes search, reranking, summarization, citation handling, evidence extraction, claim tracking, and the construction of a usable scientific context for downstream reasoning. Its technical role is to anchor later stages in traceable evidence rather than generic model priors.

\item[] \textcolor[HTML]{F58518}{\largedot} \textit{\ul{Hypothesis Formation and Planning.}}
Grounded context must then be converted into candidate scientific directions. This stage includes proposal generation, task decomposition, feasibility assessment, prioritization, branching, and the organization of alternative research paths before execution begins. Its technical role is to make candidate hypotheses sufficiently grounded, operationalizable, and comparable for downstream testing.

\item[] \textcolor[HTML]{54A24B}{\largedot} \textit{\ul{Experimentation and Tool Use.}}
Scientific claims acquire resistance only when they are exposed to executable, computational, or empirical environments. This stage includes implementation, environment invocation, tool routing, protocol enactment, execution-time revision, and the handling of intermediate outputs generated by code, tools, simulators, instruments, or laboratory settings. Its technical role is to move the workflow beyond speculative reasoning by coupling claims to environments that can constrain, refine, or invalidate them.

\item[] \textcolor[HTML]{B279A2}{\largedot} \textit{\ul{Feedback, Validation, and Review.}}
Execution alone does not produce scientific progress unless outputs are checked, challenged, revised, or rejected. This stage includes reruns, baseline comparison, error detection, uncertainty surfacing, reviewer-style critique, verification, and other mechanisms that determine whether results should persist. Its technical role is to introduce rejection pressure and distinguish stronger results from weaker ones.

\item[] \textcolor[HTML]{E45756}{\largedot} \textit{\ul{Reporting and Knowledge Communication.}}
The final stage translates workflow state into communicable scientific artifacts. It includes drafting, revision, figure and table generation, review responses, artifact packaging, and explicit links between claims, evidence, code, and provenance. Its technical role is to keep outputs interpretable, inspectable, and reusable as scientific objects rather than merely polished text.

\end{itemize}

Taken together, these stages define the technical pathway through which AutoResearch systems move from assistance to execution and integrated pipeline automation. Stronger workflow automation requires more than the presence of retrieval tools or writing modules; it requires a workflow in which grounding informs planning, planning directs execution, execution produces evidence, validation applies rejection pressure, and reporting preserves provenance and accountability. This coupling also explains why broad pipeline automation remains insufficient for mature AI-led autonomy when routine human verification is still necessary. The following subsections analyze these stages in turn.

\subsection{Stage I: Literature and Research Grounding}
\label{sec:workflow_literature}

Literature and research grounding is the first major technical stage of AutoResearch because every later part of the workflow depends on how prior work is accessed, filtered, represented, revised, and reused. In scientific research, this stage is not reducible to a generic retrieval front end. It establishes the evidential basis on which hypotheses are proposed, experiments are designed, results are interpreted, and reports are written. Recent survey work increasingly treats literature handling as a foundational part of the research lifecycle rather than as a peripheral information-access tool, precisely because weaknesses at this stage propagate into every downstream component of the workflow~\citep{Chen2025AI4Research, Zheng2025Agent4S}. The central technical concern is therefore not only whether a system can find relevant papers, but whether it can construct a scientific context that remains usable, inspectable, and updateable as the workflow evolves.

This stage includes a broad family of operations that extend well beyond search alone. It covers query formulation, document retrieval, reranking, summarization, citation handling, claim extraction, evidence comparison, contradiction surfacing, relation construction, and the organization of prior work into a form that can support later reasoning. In some systems, this grounding layer remains relatively light, serving mainly as local search and synthesis support; in others, it becomes a stronger evidential substrate that links claims to sources, preserves intermediate evidence state, or structures relations among methods, datasets, results, and limitations. The technical goal of this stage is therefore to transform raw literature access into workflow-relevant scientific grounding: later stages should not operate on free-floating model priors or compressed summaries alone, but on evidence that remains traceable, revisable, and sufficiently structured to constrain planning, execution, validation, and reporting. 

\begin{figure*}[ht]
\centering
\includegraphics[width=1\linewidth]{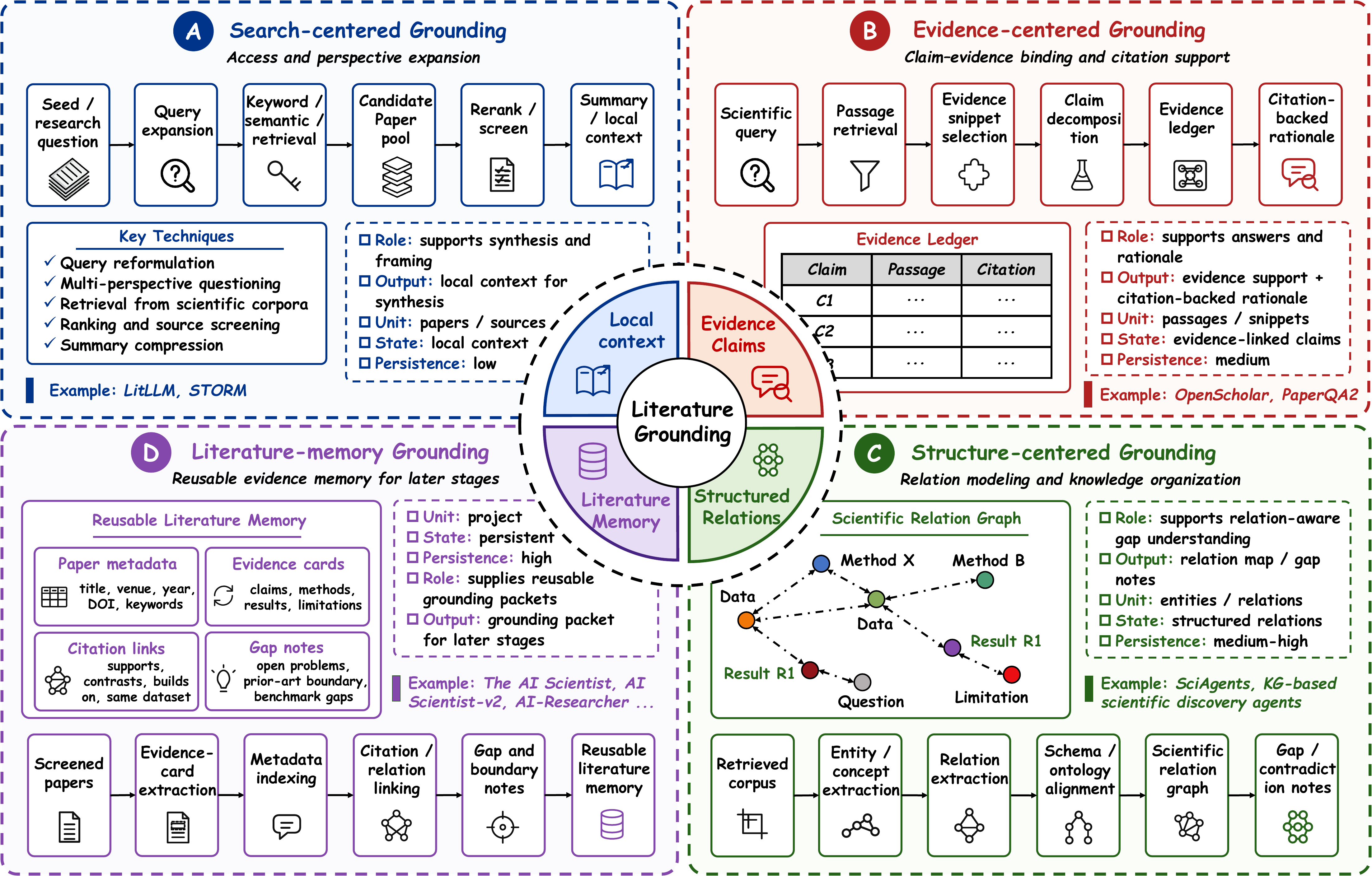}
\caption{\textbf{Literature grounding regimes as evidence-state construction.} The figure compares four recurring regimes that transform retrieved literature into reusable evidence states, including local context, evidence-linked claims, structured scientific relations, and literature memory. Downstream planning, execution, validation, and reporting are treated as consumers of these states rather than components of grounding itself.}
\label{fig:literature_grounding_patterns}
\end{figure*}

Figure~\ref{fig:literature_grounding_patterns} summarizes the main grounding regimes that recur across current systems, and the discussion below examines how these regimes differ in evidential strength, persistence, and workflow integration. Current grounding techniques can be organized into four recurring regimes: search-centered, evidence-centered, structure-centered, and literature-memory grounding. This distinction emphasizes not retrieval strength alone, but how retrieved material is represented, preserved, and propagated as usable scientific evidence across the workflow.

\begin{itemize}[nolistsep, leftmargin=*]

\item[] \textcolor[HTML]{4C78A8}{\largedot} \textit{\ul{1) Search-centered grounding.}} This is the lightest and most widespread regime. Its workflow is typically organized as \textit{query formulation $\rightarrow$ retrieval / reranking $\rightarrow$ summary compression $\rightarrow$ local scientific context construction}. \textit{LitLLM}~\citep{Agarwal2024LitLLM} is the clearest anchor here: it retrieves papers from a query abstract, reranks them, and generates a related-work section grounded in the retrieved set. \textit{STORM}~\citep{StanfordStormGitHub} can be read as an adjacent case because it combines retrieval with perspective-guided question asking to support long-form article generation, although its emphasis is broader article writing rather than scientific evidential control. The strength of this regime lies in breadth and speed: it lowers the cost of corpus exploration and provides workable local context for early-stage synthesis. Its limit is that grounding remains shallow and weakly persistent. Retrieved papers may support a single answer or summary while leaving contradiction handling, citation discipline, and downstream reuse largely external to the workflow.

\item[] \textcolor[HTML]{F58518}{\largedot} \textit{\ul{2) Evidence-centered grounding.}} This regime strengthens grounding by inserting an explicit evidence layer between retrieval and downstream reasoning. Its workflow is closer to \textit{retrieve $\rightarrow$ identify supporting passages / claims / citations $\rightarrow$ construct evidence-linked answer or rationale $\rightarrow$ pass grounded state downstream}. \textit{OpenScholar}~\citep{OpenScholarGitHub} is central here because it is explicitly designed to answer scientific queries by identifying relevant passages from a large literature corpus and producing citation-backed synthesis. \textit{PaperQA2}~\citep{PaperQA2_2024} also belongs here because its contribution is not just document retrieval but source-linked scientific synthesis, and \textit{HypER}~\citep{Vasu2025HypER} extends this regime into literature-grounded hypothesis generation with provenance. The gain is stronger evidential discipline: claims are no longer supported by compressed context alone, but by explicit links to supporting materials. The limitation is that evidence may still remain stage-local. A system can preserve evidence during answering or ideation without yet maintaining that evidence as a persistent control state across later planning, execution, and revision.

\item[] \textcolor[HTML]{54A24B}{\largedot} \textit{\ul{3) Structure-centered grounding.}} This regime makes scientific relations themselves explicit. Its workflow is closer to \textit{retrieve papers $\rightarrow$ extract entities / concepts / relations $\rightarrow$ organize them into structured representations $\rightarrow$ use that structure for synthesis, gap detection, or ideation support}. \textit{SciAgents}~\citep{Ghafarollahi2024SciAgents} is the strongest anchor because it combines retrieval with ontological knowledge graphs and graph-based reasoning to uncover interdisciplinary relations and guide scientific hypothesis generation. The advantage of this regime is that grounding moves beyond document handling toward relation modeling: methods, concepts, datasets, limitations, and open gaps can be represented as structured objects rather than only as text snippets. This is especially valuable for scientific synthesis and novelty assessment. Its weakness is that structure quality becomes the upper bound of grounding quality. Weak extraction, incomplete graphs, or oversimplified schemas can create the appearance of organized scientific context while discarding ambiguity and caveat.

\item[] \textcolor[HTML]{B279A2}{\largedot} \textit{\ul{4) Literature-memory grounding.}} This regime treats literature grounding as a reusable evidence state rather than a one-shot retrieval or summarization step. Its workflow is closer to \textit{screened papers $\rightarrow$ evidence-card extraction $\rightarrow$ metadata indexing $\rightarrow$ citation / relation linking $\rightarrow$ gap and boundary notes $\rightarrow$ reusable literature memory}. In this setting, the system preserves paper metadata, claim--method--result--limitation cards, citation links, and prior-art boundaries so that later stages can query a grounded literature state instead of relying on compressed summaries or model memory alone. End-to-end research systems such as \textit{AI Scientist}~\citep{Lu2024AIScientist}, \textit{AI Scientist-v2}~\citep{Yamada2025AIScientistV2}, \textit{AI-Researcher}~\citep{Tang2025AIResearcher}, and \textit{Agent Laboratory}~\citep{Schmidgall2025AgentLab} contain such literature-review, literature-search, or novelty-check components, but in this stage they should be read only as examples of literature grounding rather than as evidence that the full research loop belongs inside grounding. The strength of this regime is carry-forward: literature evidence can be preserved as reusable grounding packets for planning, execution, validation, and reporting. Its limitation is provenance and scope control. If the system does not record which papers, passages, claims, or prior-art comparisons support later decisions, persistent literature memory may create continuity without auditable evidential grounding.

\end{itemize}

\begin{tcolorbox}[archquestionbox,
title={Analytical implications of current grounding regimes}]
Current grounding systems differ less in whether they retrieve documents than in how retrieved material is made scientifically usable. Search-centered regimes primarily expand access and construct local context; evidence-centered regimes strengthen citation-linked support and claim-level discipline; structure-centered regimes make cross-paper relations explicit; and literature-memory regimes preserve prior work as reusable evidence state that can be queried by later stages. Read together, these regimes show that the central technical frontier of grounding is not document retrieval in isolation, but whether literature can survive compression, remain source-faithful, acquire reusable structure, and be preserved with enough provenance to support later reasoning. In this sense, the unresolved bottleneck is evidential state construction rather than literature access alone.
\end{tcolorbox}

Literature grounding should therefore be understood not merely as retrieval, but as evidential state management. Its technical frontier is whether evidence remains source-faithful, structurally usable, and operationally persistent across synthesis, planning, execution, revision, and reporting. Thus, grounding is bounded less by retrieval coverage than by the durability, inspectability, and constraining power of evidence.

\subsection{Stage II: Hypothesis Formation and Planning}
\label{sec:workflow_hypothesis_planning}

Hypothesis formation and planning constitute the second major technical stage of AutoResearch because this is where grounded scientific context is converted into a candidate direction. If literature grounding determines what a system knows about prior work, hypothesis formation and planning determine what the system is prepared to do with that knowledge. In scientific workflows, this stage is not exhausted by generic idea generation. It includes the formulation of candidate hypotheses, the decomposition of research goals into intermediate steps, the comparison of alternative directions, the anticipation of feasibility constraints, and the selection of directions that are worth exposing to downstream execution. Recent survey work increasingly treats planning as a distinct stage of the research lifecycle precisely because later experimentation and validation are only as meaningful as the proposals that survive this stage~\citep{Chen2025AI4Research, Zheng2025Agent4S}. The central technical concern is therefore not merely whether a system can produce more ideas, but whether it can structure exploration so that candidate directions become evidence-aware, operationalizable, and worth testing.

\begin{figure*}[ht]
\centering
\includegraphics[width=1\linewidth]{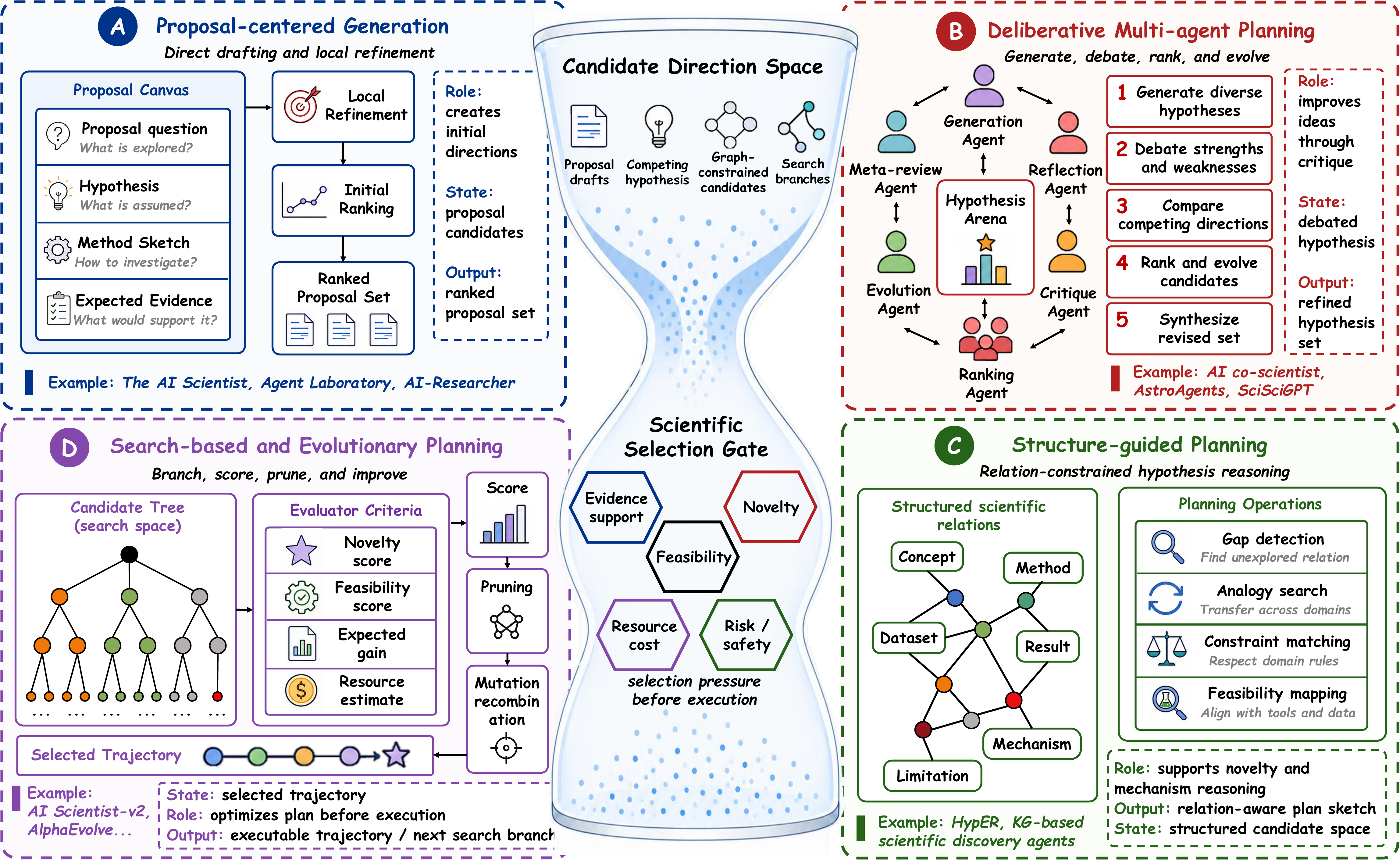}
\caption{\textbf{Hypothesis formation and planning regimes in AutoResearch.} The figure compares four recurring regimes for generating and selecting candidate research directions: proposal-centered generation, deliberative multi-agent planning, structure-guided planning, and search-based evolutionary planning. The central selection gate summarizes the main constraints used to filter directions before execution, including evidence support, novelty, feasibility, resource cost, and risk or safety.}
\label{fig:hypothesis_generation_patterns}
\end{figure*}

This stage includes a wide range of operations that sit between grounding and execution. It covers proposal generation, analogy and recombination, decomposition, feasibility assessment, branch expansion, prioritization, candidate comparison, and the organization of research directions before costly intervention begins. In some systems, this layer remains relatively light, serving mainly as local ideation support; in others, it becomes a stronger planning substrate that explicitly introduces critique, scientific structure, or search over alternatives. The technical goal is to transform grounded context into disciplined scientific direction: later stages should not inherit arbitrary or weakly filtered ideas, but candidate hypotheses and plans that are sufficiently grounded, comparable, and rejectable to justify execution. 

Figure~\ref{fig:hypothesis_generation_patterns} summarizes the main ideation-and-planning regimes that recur across current systems, and the discussion below examines how these regimes differ in search breadth, structural control, and downstream executability. Current hypothesis-formation techniques can be organized into four recurring regimes: proposal-centered ideation, deliberative multi-agent ideation, structure-guided ideation, and search-based ideation and planning. This distinction emphasizes not idea volume but how candidate directions are generated, constrained, compared, and selected before execution.


\begin{itemize}[nolistsep, leftmargin=*]

\item[] \textcolor[HTML]{4C78A8}{\largedot} \textit{\ul{1) Proposal-centered ideation.}} This is the most compact planning regime. Its workflow is typically organized as \textit{grounded context construction $\rightarrow$ candidate proposal $\rightarrow$ local refinement $\rightarrow$ ranked hypothesis or plan output}. \textit{ResearchAgent}~\citep{Undermind2024ResearchAgentIterativeResear} is a strong anchor here because it automatically proposes research problems, methods, and experiment designs while iteratively refining them over scientific literature. \textit{HypER}~\citep{Vasu2025HypER} also belongs here because it focuses on literature-grounded hypothesis generation with explicit reasoning and provenance. The strength of this regime lies in coherence: one controller can maintain a stable task representation and map grounded context directly into a candidate hypothesis or plan. Its limitation is that critique and selection pressure remain relatively concentrated. Candidate ideas may be locally plausible and well phrased while still lacking strong feasibility filtering or sufficiently independent rejection.

\item[] \textcolor[HTML]{F58518}{\largedot} \textit{\ul{2) Deliberative multi-agent ideation.}} This regime distributes ideation and planning across multiple roles rather than concentrating them in a single controller. Its workflow is closer to \textit{grounded context $\rightarrow$ multi-agent proposal generation $\rightarrow$ cross-agent critique / comparison $\rightarrow$ refined candidate set}. \textit{PiFlow}~\citep{PiFlow2025} is a central anchor because it frames scientific discovery as principle-aware multi-agent exploration rather than unconstrained hypothesis generation. \textit{Robin}~\citep{ghareeb2026robin} also belongs here because its literature-search and data-analysis agents iteratively generate and update hypotheses in a lab-in-the-loop workflow, and \textit{OmniScientist}~\citep{Shao2025OmniScientist} extends this logic by embedding collaborative protocols and human participation inside a broader scientific ecosystem. The gain is heterogeneity: different agents can propose, compare, criticize, and refine candidate directions before execution begins. The limitation is that apparent diversity can be overstated. If arbitration is weak or role separation is mostly rhetorical, multi-agent discussion can generate more text without generating meaningfully stronger scientific distinction.

\item[] \textcolor[HTML]{54A24B}{\largedot} \textit{\ul{3) Structure-guided ideation.}} This regime strengthens planning by imposing explicit scientific structure on the idea-formation process. Its workflow is closer to \textit{structured grounding $\rightarrow$ principle / graph / facet-guided proposal formation $\rightarrow$ constrained candidate synthesis}. \textit{Scideator}~\citep{Radensky2024Scideator} is a representative anchor because it grounds scientific ideation in research-paper facet recombination, allowing users to explore candidate directions by recombining purposes, mechanisms, and evaluations extracted from prior work. \textit{SciAgents}~\citep{Ghafarollahi2024SciAgents} also belongs here because it uses graph-grounded scientific reasoning to represent and exploit cross-paper relations for hypothesis generation. The gain of this regime is stronger scientific anchoring: candidate directions are not generated only from stylistic plausibility, but from explicit structural relations in prior work. The limitation is that structure quality becomes the upper bound of ideation quality. Weak facet extraction, incomplete graphs, or oversimplified schemas can narrow the search space prematurely and suppress unconventional but valuable directions.

\item[] \textcolor[HTML]{B279A2}{\largedot} \textit{\ul{4) Search-based ideation and planning.}} In this regime, ideation is no longer treated as a one-shot proposal step. Instead, multiple candidate branches are generated, preserved, scored, and pruned under explicit search logic. \textit{AI Scientist-v2}~\citep{Yamada2025AIScientistV2} is the clearest anchor because it uses agentic tree search to expand and refine research directions before committing them to later stages. \textit{AlphaEvolve}~\citep{Novikov2025AlphaEvolve} belongs here in a narrower algorithmic-discovery sense because it frames candidate improvement as iterative search under evaluator feedback, and \textit{NovelSeek}~\citep{Zhang2025NovelSeek} extends this pattern toward multi-agent hypothesis refinement and verification-oriented planning. The gain is that uncertainty is represented structurally rather than rhetorically: instead of pretending that one proposal is obviously best, the workflow preserves alternatives long enough for comparison and pruning. The limitation is that branching helps only if selection criteria are scientifically meaningful rather than merely stylistic or benchmark-convenient. Otherwise, search enlarges the space of proposals without strengthening the quality of commitment.

\end{itemize}

\begin{tcolorbox}[archquestionbox,
title={Analytical implications of current ideation and planning regimes}]
Current planning systems differ less in whether they generate candidate directions than in how those directions are constrained before execution. Proposal-centered regimes primarily improve local coherence; deliberative multi-agent regimes expand comparison and critique; structure-guided regimes make the sources of candidate generation more explicit; and search-based regimes preserve alternatives long enough for nontrivial pruning and refinement. Read together, these regimes show that the technical frontier of planning is not idea generation in isolation, but whether candidate directions remain grounded, comparable, feasible, and rejectable before downstream resources are committed. In this sense, the unresolved bottleneck is disciplined scientific search rather than ideation volume alone.
\end{tcolorbox}

Hypothesis formation and planning should therefore be understood less as a creativity problem than as a search-and-selection problem. The technical frontier lies in whether grounded context can be converted into candidate directions that remain evidence-aware, operationalizable, and meaningfully filtered as the workflow moves toward execution and validation. This is why the ceiling of planning is ultimately set not by the fluency of generated ideas, but by whether the workflow can preserve enough structure, critique, and selection pressure to distinguish viable scientific directions from merely plausible ones.

\subsection{Stage III: Experimentation and Tool Use}
\label{sec:workflow_execution}

Experimentation and tool use constitute the third major technical stage of AutoResearch because they determine how candidate scientific directions are translated into concrete actions. If grounding establishes the evidential basis of the workflow and planning determines which directions are worth pursuing, execution determines whether those directions can be realized through code, tools, instruments, protocols, or human-gated interventions. In scientific workflows, this stage is not exhausted by generic tool calling. It includes repository editing, code execution, simulator use, scientific software invocation, tool routing, parameterized API calls, protocol enactment, robotic action, instrument scheduling, and expert-mediated execution procedures. Recent work increasingly treats this stage as a decisive separator between systems that only assist ideation and those that begin to sustain more consequential workflow automation, precisely because execution is the point at which plans must be bound to actionable substrates rather than remain as speculative reasoning alone~\citep{Chen2025AI4Research, Zheng2025Agent4S, Boiko2023Autonomous}.

\begin{figure*}[ht]
\centering
\includegraphics[width=1\linewidth]{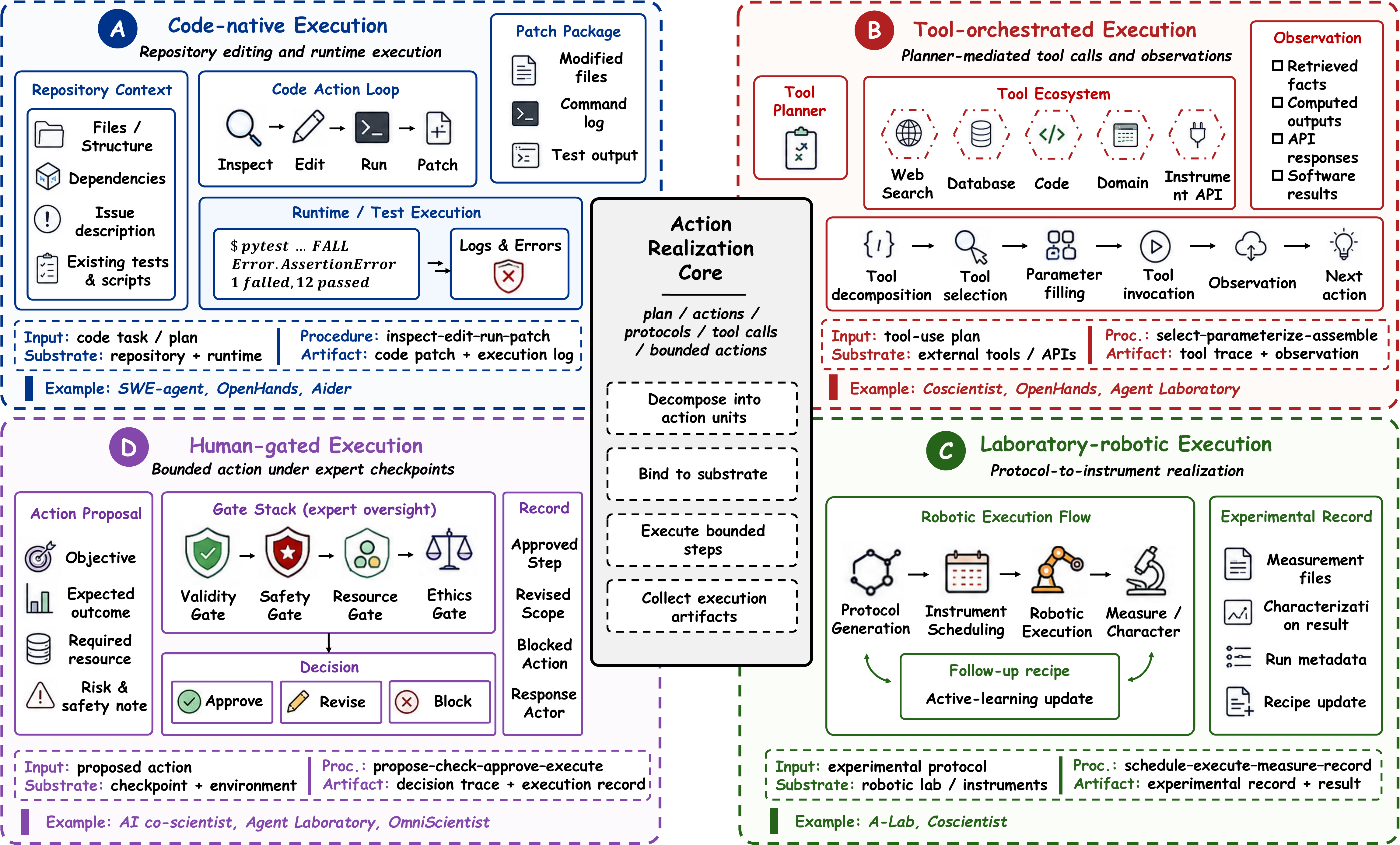}
\caption{\textbf{Experimentation and tool use regimes as action realization.} The figure compares four recurring regimes that translate executable research plans into concrete actions: code-native execution, tool-orchestrated execution, laboratory-robotic execution, and human-gated execution. Together, these regimes show how plans are realized through repositories, tools, instruments, or expert checkpoints before producing execution artifacts and observations.}
\label{fig:execution_pipeline_patterns}
\end{figure*}

This stage covers a broad family of operations that sit between planning and validation. It includes implementation, environment selection, tool routing, parameterization, experiment running, protocol translation, intermediate-result handling, and bounded revision during execution. In some systems, execution remains relatively light, serving mainly as code running or software invocation in programmable settings; in others, it becomes a stronger substrate that binds plans to specialized tools, simulators, laboratory robots, instruments, or human-controlled checkpoints. The technical goal of this stage is therefore to transform candidate scientific directions into actionable execution procedures: later validation should not inherit only plausible plans, but concrete artifacts such as patches, tool traces, observation bundles, run records, measurement files, and decision traces produced by identifiable execution substrates.

Figure~\ref{fig:execution_pipeline_patterns} summarizes the main execution regimes that recur across current systems, and the discussion below examines how these regimes differ in execution substrate, action procedure, and workflow control. Current execution techniques can be organized into four recurring regimes: code-native, tool-orchestrated, laboratory-robotic, and human-gated execution. The first three mainly differ by the substrate through which actions are realized, while human-gated execution captures a control pattern in which consequential steps remain mediated by explicit human checkpoints.


\begin{itemize}[nolistsep, leftmargin=*]

\item[] \textcolor[HTML]{4C78A8}{\largedot} \textit{\ul{1) Code-native execution.}} This is the most widespread execution regime in machine-learning-centered AutoResearch systems. Its workflow is typically organized as \textit{code task / plan $\rightarrow$ repository inspection $\rightarrow$ file localization $\rightarrow$ edit / patch $\rightarrow$ command execution $\rightarrow$ runtime or test execution $\rightarrow$ patch package}. \textit{The AI Scientist}~\citep{Lu2024AIScientist} is the clearest anchor here because it connects idea generation to executable ML experiments, result analysis, and paper drafting inside a software-native loop. Execution substrates such as \textit{OpenHands}~\citep{OpenHandsGitHub}, \textit{Aider}~\citep{AiderGitHub}, and \textit{SWE-agent}~\citep{SWEAgentGitHub} are adjacent anchors because they demonstrate how autonomous coding environments can inspect repositories, edit files, run commands, and package patches under tool access, even when they are not themselves full scientific discovery systems. The strength of this regime lies in programmability: implementation, command execution, local testing, and patch generation can all be orchestrated in a software environment with relatively low friction. Its limitation is that executability can easily be mistaken for scientific adequacy. A runnable script, a passing pipeline, or a clean patch package does not by itself guarantee meaningful task framing, fair comparison, or strong methodological support.

\item[] \textcolor[HTML]{F58518}{\largedot} \textit{\ul{2) Tool-orchestrated execution.}} This regime strengthens execution by routing plans through external scientific tools rather than through code generation alone. Its workflow is closer to \textit{tool-use plan $\rightarrow$ task decomposition $\rightarrow$ tool selection $\rightarrow$ parameter filling $\rightarrow$ tool invocation $\rightarrow$ observation assembly $\rightarrow$ next-action proposal}. \textit{ChemCrow}~\citep{Bran2024ChemCrow} is a central anchor because it integrates an LLM with chemistry tools for synthesis planning, molecular analysis, and domain-specific task completion. \textit{Biomni}~\citep{Biomni2025} belongs here because it couples biomedical reasoning to external tools and scientific resources, and \textit{MM-Agent}~\citep{Liu2025MMAgent} is relevant because it emphasizes scientific tool integration across multimodal settings. \textit{Graph of Skills}~\citep{li2026graph} introduced a dependency-aware structural retrieval framework for scalable multi-step agent skill orchestration. The gain is stronger domain coupling: execution is no longer limited to generic scripts, but engages directly with specialized software, calculators, simulators, databases, APIs, and domain-specific scientific utilities. The limitation is that execution quality becomes tightly dependent on tool fidelity, routing correctness, parameter specification, and hidden environment assumptions. A workflow may appear more capable primarily because its tools are strong, while its own decomposition, selection, or control logic remains comparatively fragile.

\item[] \textcolor[HTML]{54A24B}{\largedot} \textit{\ul{3) Laboratory-robotic execution.}} This regime is strongest in terms of physical action realization and narrowest in terms of portability. Its workflow is closer to \textit{experimental protocol or recipe $\rightarrow$ protocol generation $\rightarrow$ instrument scheduling $\rightarrow$ robotic execution $\rightarrow$ measurement / characterization $\rightarrow$ experimental record}. \textit{Mobile Robot Chemist}~\citep{Burger2020MobileRobotChemist} and \textit{A-Lab}~\citep{Szymanski2023AutonomousLab} are foundational anchors because they show how automated experimentation can be linked to closed-loop scientific search in physical environments. More recent systems such as \textit{AutoLabs}~\citep{Panapitiya2025AutoLabs}, \textit{ORGANA}~\citep{Darvish2025ORGANA}, and \textit{QuantumAgent SDL}~\citep{Cao2024QuantumAgentSDL} extend this pattern by coupling protocol generation, robotic control, measurement, and experimental records inside increasingly agentic workflows. The gain is that candidate plans are realized as instrumented experimental actions: recipes meet scheduling constraints, protocols meet hardware interfaces, and outcomes are preserved as run records or measurement files rather than only textual claims. The cost is severe dependence on local infrastructure, hardware interfaces, safety protocols, and domain-specific procedures. This regime is therefore strong in physical action realization but weak in portability and generality.

\item[] \textcolor[HTML]{B279A2}{\largedot} \textit{\ul{4) Human-gated execution.}} In this regime, execution remains inside the workflow but is not treated as fully autonomous. Instead, consequential steps are gated by expert checkpoints, collaborative protocol refinement, or explicit handoff between system-generated plans and human-controlled action. Its workflow is closer to \textit{proposed action $\rightarrow$ objective / resource / risk specification $\rightarrow$ expert checkpoint $\rightarrow$ validity, safety, resource, and ethics gates $\rightarrow$ approve / revise / block $\rightarrow$ bounded execution record}. \textit{AI co-scientist}~\citep{gottweis2025towards} is the clearest anchor because it is explicitly framed as a collaborative system that integrates scientific reasoning, tool use, and human feedback rather than removing scientists from the loop. \textit{Free PhD}~\citep{Li2025Build}, \textit{Agent Laboratory}~\citep{Schmidgall2025AgentLab}, and \textit{OmniScientist}~\citep{Shao2025OmniScientist} also belong here because they embed execution in workflows where AI can propose and run substantial subwork, while humans retain visible intervention rights at consequential stages. The gain is stronger accountability: risky, costly, or high-stakes actions can be approved, revised, or blocked before they harden into downstream consequences. The limitation is that autonomy ceilings remain lower and system performance depends partly on human availability, expertise, and willingness to intervene.

\end{itemize}

\begin{tcolorbox}[archquestionbox,
title={Analytical implications of current execution regimes}]
Current execution systems differ less in whether they can act than in how plans are bound to actionable substrates. Code-native regimes realize plans through repositories, runtimes, tests, and patch artifacts; tool-orchestrated regimes route plans through external tools, APIs, and domain software; laboratory-robotic regimes translate protocols into instrument actions, measurements, and experimental records; and human-gated regimes preserve accountability through explicit checkpoints before bounded execution. Read together, these regimes show that the central technical frontier of execution is not action in isolation, but whether executable plans can be converted into inspectable artifacts through substrates that are scientifically meaningful, controllable, and safe. In this sense, the unresolved bottleneck is not the ability to run code or call tools, but the ability to sustain action realization under conditions that remain traceable, bounded, and suitable for later validation.
\end{tcolorbox}

Experimentation and tool use should therefore be understood less as a generic action problem than as an action-realization problem. The technical frontier lies in whether candidate scientific directions can be decomposed, bound to suitable execution substrates, executed through bounded procedures, and preserved as inspectable artifacts for later validation. This is why the ceiling of execution is ultimately set not by whether a system can do something, but by whether its actions are operationally traceable, scientifically meaningful, and sufficiently governed before their outputs are carried forward. This is also why many execution-capable systems remain human-verified: they can act, but consequential interpretation, validation, and acceptance remain human-held.

\subsection{Stage IV: Feedback, Validation, and Review}
\label{sec:workflow_validation}

Feedback, validation, and review constitute the fourth major technical stage of AutoResearch because they determine whether intermediate and final outputs are revised, rejected, or allowed to persist as candidate scientific claims. If grounding establishes the evidential basis of the workflow, planning shapes candidate direction, and execution exposes those directions to operational or empirical resistance, validation determines whether the resulting outputs are subjected to sufficiently strong filtering to support credible scientific progress. In scientific workflows, this stage is not reducible to metric reporting or simple reruns. It includes baseline comparison, rerun validation, error detection, critical review, uncertainty surfacing, verification, revision guidance, and the broader mechanisms by which weak results are challenged before they harden into scientific artifacts. Recent evaluation work increasingly treats this stage as a central bottleneck for autonomous research systems, precisely because the presence of execution does not by itself imply the presence of meaningful rejection pressure~\citep{Chen2025AIRSBench, Xie2025How, Undermind2026FIREBenchEvaluatingAgentsont}.

\begin{figure*}[ht]
\centering
\includegraphics[width=1\linewidth]{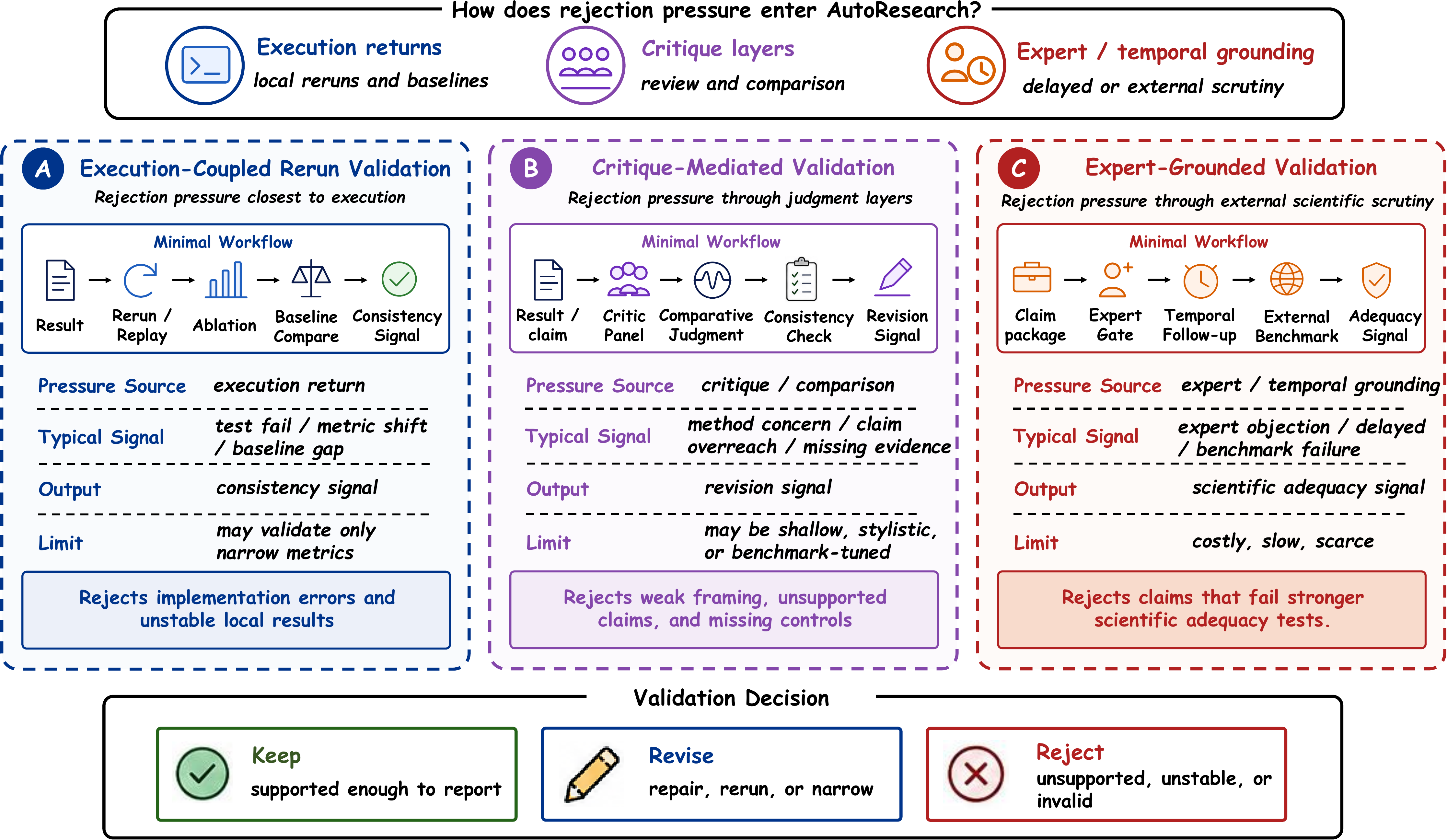}
\caption{\textbf{Validation and rejection regimes in AutoResearch workflows.} The figure compares three recurring regimes through which rejection pressure enters AutoResearch: execution-coupled reruns, critique-mediated review, and expert-grounded evaluation. The resulting signals support keep, revise, or reject decisions.}
\vspace{-5pt}
\label{fig:validation_pipeline_patterns}
\end{figure*}

This stage covers a wide range of operations that sit between execution and reporting. It includes consistency checks, ablation-oriented reruns, benchmark comparison, critic modules, reviewer-style assessment, rubric-based scoring, expert feedback, and temporally extended evaluation of whether claimed findings remain meaningful under stronger scrutiny. In some systems, validation remains relatively light, functioning mainly as local rerun checking or benchmark comparison inside a bounded experimental loop; in others, it becomes a stronger substrate that introduces reviewer-style critique, structured revision signals, or delayed external evaluation. The technical goal of this stage is therefore to transform raw outputs into scientifically discriminated outputs: later reporting should not inherit only results that ran successfully or look persuasive but results that have survived enough rejection pressure to justify further interpretation, communication, and reuse. 

Figure~\ref{fig:validation_pipeline_patterns} summarizes the main validation regimes that recur across current systems, and the discussion below examines how these regimes differ in rejection strength, evaluative fidelity, and relation to scientific accountability. Current validation techniques can be organized into three recurring regimes: execution-coupled rerun validation, critique-mediated validation, and expert- or temporally-grounded validation. This distinction emphasizes not scoring alone, but where and how rejection pressure enters the workflow before weak outputs stabilize as scientific claims.


\begin{itemize}[nolistsep, leftmargin=*]

\item[] \textcolor[HTML]{4C78A8}{\largedot} \textit{\ul{1) Execution-coupled rerun validation.}} This is the most direct validation regime because it keeps rejection pressure close to the execution loop itself. Its workflow is typically organized as \textit{candidate result $\rightarrow$ rerun / ablation / baseline comparison $\rightarrow$ consistency check $\rightarrow$ keep, revise, or reject}. \textit{PaperBench}~\citep{Starace2025PaperBench} is a useful anchor here because it evaluates whether agents can reproduce and implement papers under execution-grounded conditions rather than only narrating them. Neighboring systems and evaluations that couple results tightly to reruns or ablations play a similar role by forcing candidate claims to survive immediate methodological resistance before they are retained. The strength of this regime is that validation remains close to the environment that produced the result, which makes inconsistency and implementation error relatively easy to surface. Its limitation is that reruns and local checks may still validate the wrong thing: weak baselines, narrow tasks, or poorly framed metrics can make an output appear stable without making it scientifically meaningful.

\item[] \textcolor[HTML]{F58518}{\largedot} \textit{\ul{2) Critique-mediated validation.}} This regime shifts more of the rejection burden into judgment layers that compare, criticize, and revise outputs rather than merely rerunning them. Its workflow is closer to \textit{candidate result $\rightarrow$ critic / reviewer / comparison module $\rightarrow$ structured feedback $\rightarrow$ revision or rejection}. \textit{LLM-REVal}~\citep{Undermind2025LLMREValCanWeTrustLLMReviewe} is a central anchor because it explicitly studies multi-round review and revision dynamics in an academic-style setting, rather than treating evaluation as a one-shot score. \textit{Can Large Language Models Provide Useful Feedback on Research Papers?}~\citep{liang2024can} also belongs here because it directly examines the usefulness and limitations of LLM-generated paper feedback at scale. The gain of this regime is that scientific weakness is not reduced to execution failure alone; systems can be challenged on framing, novelty, clarity, missing controls, or reviewer-style concerns. The limitation is that critique layers can themselves be stylistically shallow or benchmark-tuned, thereby reproducing the appearance of scientific judgment without fully capturing it.

\item[] \textcolor[HTML]{54A24B}{\largedot} \textit{\ul{3) Expert- or temporally-grounded validation.}} This regime strengthens validation by grounding it in expert scrutiny, delayed follow-up, or rediscovery-oriented evaluation rather than in one-shot automated checking. Its workflow is closer to \textit{candidate result $\rightarrow$ expert or delayed evaluation $\rightarrow$ stronger evidential pressure $\rightarrow$ retain or reject}. \textit{FIRE-Bench}~\citep{Undermind2026FIREBenchEvaluatingAgentsont} is the clearest anchor because it evaluates agent systems on the rediscovery of scientific insights and thereby imposes a stronger notion of scientific adequacy than local execution success alone. \textit{AIRS-Bench}~\citep{Chen2025AIRSBench} is also relevant here because it evaluates agents on full research-lifecycle tasks and makes visible where apparently capable systems still fail under stronger end-to-end standards. The gain of this regime is evaluative fidelity: candidate claims face criteria that are closer to the forms of pressure exerted by real scientific communities or stronger benchmark surrogates for them. The limitation is cost, latency, and scarcity. This regime is much harder to automate deeply, which is precisely why it remains one of the strongest bottlenecks on movement toward credible scientific autonomy.

\end{itemize}

\begin{tcolorbox}[archquestionbox,
title={Analytical implications of current validation regimes}]
Current validation systems differ less in whether they assign scores than in how much rejection pressure they can introduce into the workflow. Execution-coupled regimes primarily test local repeatability; critique-mediated regimes expand the space of objections and revision signals; expert- or temporally-grounded regimes impose stronger standards of scientific adequacy beyond one-shot execution traces. Read together, these regimes show that the central technical frontier of validation is not evaluation in the abstract, but whether the workflow can reject attractive but weak outputs before they stabilize as scientific claims. In this sense, the unresolved bottleneck is robust internal and external filtering rather than checking alone.
\end{tcolorbox}

Feedback, validation, and review should therefore be understood less as a scoring problem than as a rejection problem. The technical frontier lies in whether outputs can be challenged under pressures strong enough to preserve validity, provenance, and accountable acceptance as the workflow moves toward reporting and reuse. This is why the ceiling of validation is ultimately set not by the availability of critics or benchmarks in isolation, but by whether the workflow can sustain enough discriminative pressure to distinguish scientifically credible results from merely executable or persuasive ones. This validation gap is the central reason why pipeline breadth alone does not establish mature AI-led autonomy: current systems can often generate, execute, and critique, but they still struggle to internalize rejection standards strong enough to replace routine human verification.

\subsection{Stage V: Reporting and Knowledge Communication}
\label{sec:workflow_reporting}

Reporting and knowledge communication constitute the fifth major technical stage of AutoResearch because they determine how workflow state is translated into scientific artifacts that can be inspected, critiqued, and reused. If grounding provides the evidential basis of the workflow, planning shapes candidate direction, execution exposes those directions to external environments, and validation determines which outputs survive stronger scrutiny, reporting determines how those filtered outputs are presented to scientific audiences and preserved as communicable objects. In scientific workflows, this stage is not reducible to fluent text generation. It includes drafting, revision, figure and table production, review response writing, artifact packaging, and the linking of claims to evidence, code, and provenance. Recent survey work increasingly treats scientific writing, multimodal content generation, and AI-based review support as interconnected parts of the research lifecycle, precisely because weak communication can distort, overstate, or conceal the epistemic status of results even when earlier stages have performed well.

This stage covers a broad family of operations that sit between validation and public-facing scientific output. It includes long-form drafting, related-work synthesis, section-level revision, figure and table construction, review response generation, paper-level artifact assembly, and the alignment of text with supporting evidence, code, and metadata. In some systems, reporting remains relatively light, functioning mainly as writing assistance or post hoc summarization; in others, it becomes a stronger communication substrate that links scientific content to executable artifacts, reviewer interaction, or multimodal output. The technical goal of this stage is therefore to transform workflow state into communicable scientific artifacts without severing the connection between claims and the evidence on which they rest. 

\begin{figure*}[t]
\centering
\includegraphics[width=1\linewidth]{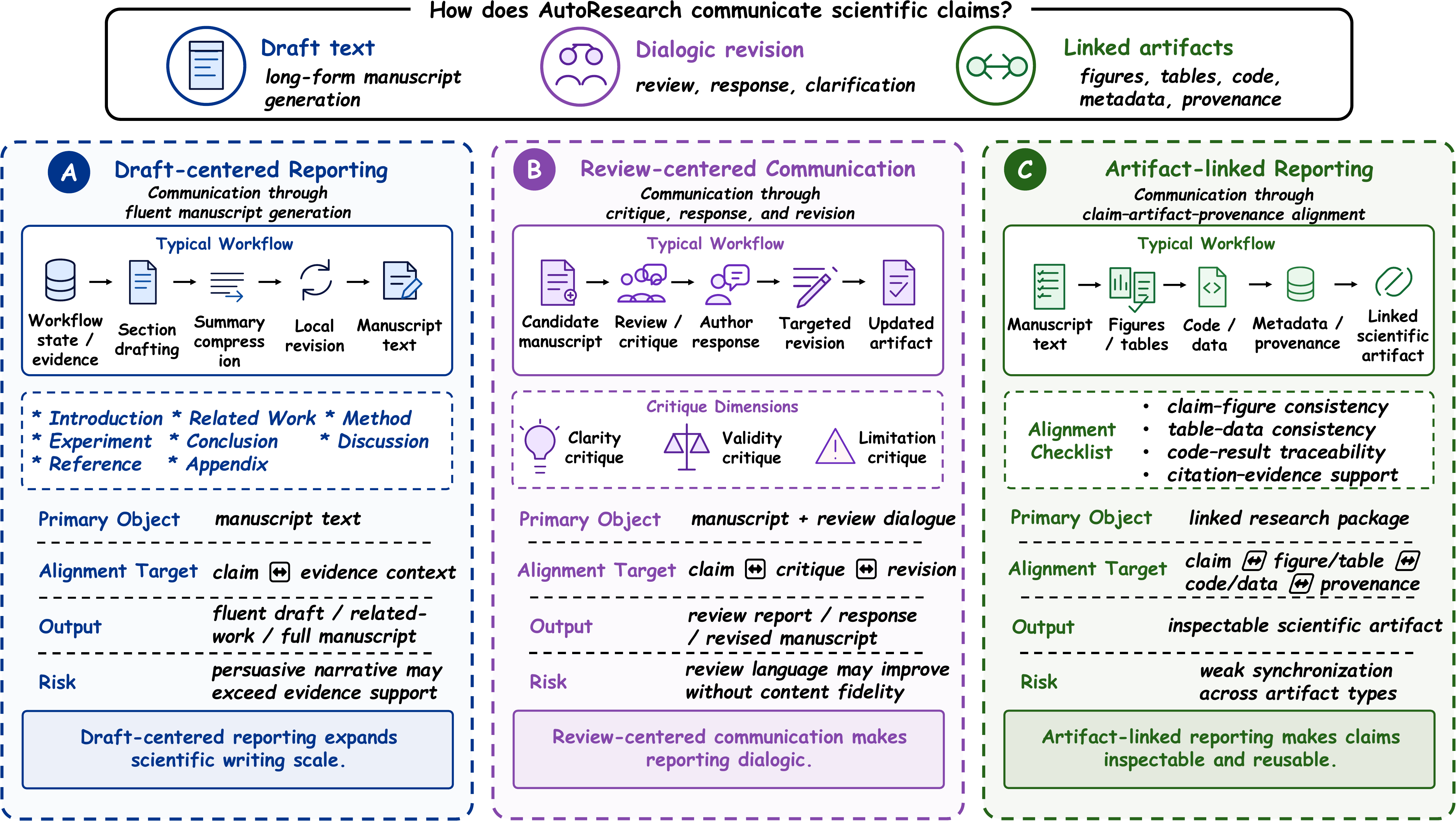}
\caption{\textbf{Reporting and knowledge communication regimes in AutoResearch.} The figure compares three recurring regimes for communicating scientific claims: draft-centered reporting, review-centered communication, and artifact-linked reporting. These regimes differ in how claims remain aligned with manuscript text, critique, figures, tables, code, metadata, and provenance.}
\label{fig:reporting_pipeline_patterns}
\end{figure*}

Figure~\ref{fig:reporting_pipeline_patterns} summarizes the main reporting regimes that recur across current systems, and the discussion below examines how these regimes differ in communicative scope, artifact linkage, and preservation of evidential discipline. Current reporting techniques can be organized into three recurring regimes: draft-centered reporting, review-centered communication, and artifact-linked reporting. This distinction emphasizes not text fluency alone, but how well communication preserves the alignment between claims, uncertainty, evidence, and reusable artifacts.

\begin{itemize}[nolistsep, leftmargin=*]

\item[] \textcolor[HTML]{4C78A8}{\largedot} \textit{\ul{1) Draft-centered reporting.}} This is the most widespread reporting regime. Its workflow is typically organized as \textit{workflow state or evidence context $\rightarrow$ section drafting / summarization $\rightarrow$ local revision $\rightarrow$ manuscript text}. \textit{LitLLM}~\citep{Agarwal2024LitLLM} is a relevant anchor because it produces literature-grounded related-work sections from retrieved papers rather than generic summaries. \textit{The AI Scientist}~\citep{Lu2024AIScientist} also belongs here because it drafts full papers as part of an end-to-end research loop, and \textit{AI Scientist-v2}~\citep{Yamada2025AIScientistV2} extends this pattern toward workshop-style manuscript production and review-oriented workflows. The strength of this regime lies in communicative scale: long-form scientific text can be generated quickly and iteratively from grounded workflow state. Its limitation is that fluency can outpace epistemic discipline. Draft-centered systems may produce coherent and persuasive narratives even when the underlying evidence, uncertainty, or methodological support remains thin.

\item[] \textcolor[HTML]{F58518}{\largedot} \textit{\ul{2) Review-centered communication.}} This regime strengthens reporting by embedding communication inside revision and feedback loops rather than treating writing as a one-way output channel. Its workflow is closer to \textit{candidate manuscript or section $\rightarrow$ review / critique $\rightarrow$ response or revision $\rightarrow$ updated artifact}. \textit{Can Large Language Models Provide Useful Feedback on Research Papers?}~\citep{liang2024can} is a central anchor because it directly studies the quality and usefulness of LLM-generated feedback on scientific manuscripts. \textit{LLM-REVal}~\citep{Undermind2025LLMREValCanWeTrustLLMReviewe} also belongs here because it explicitly models review-and-revision dynamics over multiple rounds rather than reducing evaluation to one-shot scoring. The gain of this regime is that reporting becomes dialogic: text is shaped not only by initial drafting but also by objection, response, clarification, and targeted revision. Its limitation is that review-like interaction can still remain stylistically shallow. A system may become better at producing review language and response language without necessarily improving the fidelity of the underlying scientific content.

\item[] \textcolor[HTML]{54A24B}{\largedot} \textit{\ul{3) Artifact-linked reporting.}} This regime is strongest when communication remains tied to scientific artifacts rather than to text alone. Its workflow is closer to \textit{validated workflow state $\rightarrow$ manuscript text + figures / tables / code / metadata $\rightarrow$ linked scientific artifact}. \textit{The AI Scientist}~\citep{Lu2024AIScientist} is again relevant here because it integrates result visualization and manuscript drafting inside the same loop rather than treating them as independent outputs. \textit{PaperBench}~\citep{Starace2025PaperBench} is an adjacent anchor because it makes reproducibility and implementation fidelity central to paper-level evaluation, thereby highlighting the importance of artifact-linked communication. The strength of this regime lies in inspectability. Claims, figures, tables, and executable artifacts can be cross-checked rather than read in isolation. Its limitation is that artifact linkage is technically demanding: once communication must preserve provenance across multiple artifact types, weak synchronization between text and evidence becomes a major source of failure.

\end{itemize}

\begin{tcolorbox}[archquestionbox,
title={Analytical implications of current reporting regimes}]
Current reporting systems differ less in whether they can generate polished text than in how tightly communication remains linked to evidence, critique, and artifacts. Draft-centered regimes primarily expand long-form composition; review-centered regimes introduce objection, response, and revision dynamics; and artifact-linked regimes strengthen the alignment between manuscript claims and the figures, tables, code, or metadata that support them. Read together, these regimes show that the central technical frontier of reporting is not writing quality in isolation but whether communication preserves evidential discipline while making results reusable and inspectable. In this sense, the unresolved bottleneck is communicative closure rather than fluency alone.
\end{tcolorbox}

Reporting and knowledge communication should, therefore, be understood less as a text-generation problem than as a scientific artifact-alignment problem. The technical frontier lies in whether workflow outputs can be communicated in ways that remain interpretable, inspectable, and reusable without severing their connection to evidence, uncertainty, and provenance. This is why the ceiling of reporting is ultimately set not by how polished the final manuscript appears, but by whether communication preserves enough epistemic discipline for results to withstand scrutiny and support downstream scientific reuse.

Across the five stages, the technical foundations of AutoResearch reveal a common pattern: progress is strongest when a workflow can preserve constraints across stages, and weakest when those constraints break before scientific closure. Literature grounding must preserve evidence rather than merely retrieve documents; hypothesis formation must support structured search rather than merely generate plausible ideas; experimentation must expose claims to meaningful environmental resistance rather than merely execute tools; validation must impose rejection pressure rather than merely assign scores; and reporting must maintain claim--evidence--provenance alignment rather than merely produce fluent manuscripts. This cross-stage coupling explains why current systems can already support powerful human-verified research pipelines, yet still fall short of mature AI-led scientific autonomy. The central bottleneck is not local generation, tool use, or writing quality in isolation, but whether the full workflow can reliably preserve evidence, reject weak directions, and maintain accountable scientific closure without routine human verification.
\section{Evaluation of AutoResearch}
\label{sec:evaluation}
\label{sec:evaluation_frameworks}

Evaluation has become a central challenge for AutoResearch because the field is no longer optimizing isolated scientific subtasks but increasingly organizing workflow-level inquiry in which literature grounding, hypothesis formation, execution, validation, and reporting must be judged together rather than in isolation. Under these conditions, strong local performance is not enough: a system may generate interesting ideas yet fail methodologically, complete experiments yet remain brittle across reruns, or produce polished reports whose claims cannot be traced back to reliable evidence. The object of evaluation in this chapter is therefore not an isolated model capability but the \emph{scientific quality} and \emph{autonomy profile} of research workflows operating under realistic discovery constraints. This also makes a further gap increasingly visible: workflow automation is advancing faster than workflow verifiability, so rigorous evaluation must ask not only whether a system can produce research-like outputs but also whether those outputs remain scientifically credible, traceable, and sufficiently verified to support stronger claims of autonomy.

\subsection{Evaluative Burdens Across the AutoResearch Spectrum}
\label{sec:evaluation_level_burdens}

Evaluation in AutoResearch cannot proceed as if the same evidence supported the same claim under all conditions of workflow autonomy. Systems may participate in broadly similar stages of scientific work—such as literature grounding, hypothesis formation, execution, validation, and reporting-yet still differ substantially in where control resides, how decisions are made, how evidence is propagated across stages, and who remains responsible when claims later fail or succeed. For this reason, the same benchmark success, rerun, or polished artifact does not carry invariant evaluative meaning. What changes across the AutoResearch spectrum is not the existence of evaluation itself, but the burden that evaluation is expected to discharge. In more human-centered settings, the central question is often whether AI support is locally useful without degrading validity, provenance, or interpretation. In more strongly automated settings, evaluation must bear a heavier burden: it must establish whether longer workflow spans remain auditable, failure-exposing, and sufficiently verified to justify stronger claims about scientific autonomy. The evaluation burden grows precisely when routine human verification is removed from intermediate decisions: the workflow itself must then supply the evidence, rejection pressure, provenance, and accountability that human researchers previously provided.

The practical importance of this distinction is greatest in the currently occupied region from \textcolor{RefOrange}{L1} through advanced \textcolor{RefOrange}{L2}, where polished outputs and strong local performance can easily be over-read as evidence of stronger closure than current systems actually sustain. \textcolor{RefOrange}{L3} is retained as a stricter frontier for AI-led workflows whose scientific quality no longer depends on routine human verification, and \textcolor{RefOrange}{L4} remains an analytical upper bound rather than a realized target~\citep{Popper1959LogicScientificDiscovery, Merton1973SociologyScience, gottweis2025towards, Gueroudji_2025, Biswas_2026, Zheng2025Agent4S}. This framing also explains why integrated research pipelines should not be classified by breadth alone. A system that connects many stages but still relies on human researchers to judge validity, novelty, reproducibility, and usability remains within the human-verified region of the spectrum.

\begin{tcolorbox}[archquestionbox,
title={How should evaluative burden be interpreted across the AutoResearch spectrum?}]
The key distinction is not that each part of the AutoResearch spectrum demands a wholly separate standard, but that the same apparent success must be interpreted differently as control, verification, and responsibility shift across the workflow. In human-centered research, evaluation remains institutionally distributed: scientific adequacy is judged through expert interpretation, peer scrutiny, replication, and community uptake, so contextual judgment is strong even when coordination is slow. Once AI enters as bounded assistance, the main burden is to show that local support improves scientific work without weakening validity or provenance; the principal evaluative risk is to mistake productivity gain for stronger autonomy. When execution becomes partially delegable under human verification, evaluation must expand further to account for oversight load, intervention points, and the collaborative usefulness of mixed-initiative workflows, because outcome quality alone no longer reveals how much of the process still depends on human rescue. For systems approaching AI-led workflow coordination, the burden shifts again: local competence is no longer enough unless longer workflow spans remain auditable, failure-exposing, and meaningfully constrained by evidence, validation, and accountability. The unrealized upper bound is full autonomous closure, where scientific and governance conditions would need to be jointly satisfied across the entire workflow, including end-to-end verification, robust provenance, responsibility retention, and the reliable rejection of weak claims. No current system meets that standard.
\end{tcolorbox}

\subsection{Scientific Quality and Autonomy Assessment}
\label{sec:evaluation_framework}

Although evaluative burdens shift across the AutoResearch spectrum, they still require a common vocabulary. In this survey, that vocabulary is organized around two complementary judgment targets: \textbf{scientific quality}, which asks whether a workflow produces outputs that are scientifically credible, and \textbf{autonomy assessment}, which asks how much of that workflow is actually exercised by the system. The first target concerns the epistemic status of the resulting claims and artifacts; the second concerns the strength of the workflow-level autonomy claim that those results can legitimately support. In this sense, scientific quality evaluates whether the output deserves scientific trust, whereas autonomy assessment evaluates whether the workflow deserves the autonomy claim attached to it. This distinction is necessary because AutoResearch is especially prone to overclaim: polished outputs, strong local execution, and apparent pipeline closure can easily be mistaken for stronger scientific autonomy than the underlying evidence warrants.

\begin{figure*}[ht]
\centering
\includegraphics[width=1\linewidth]{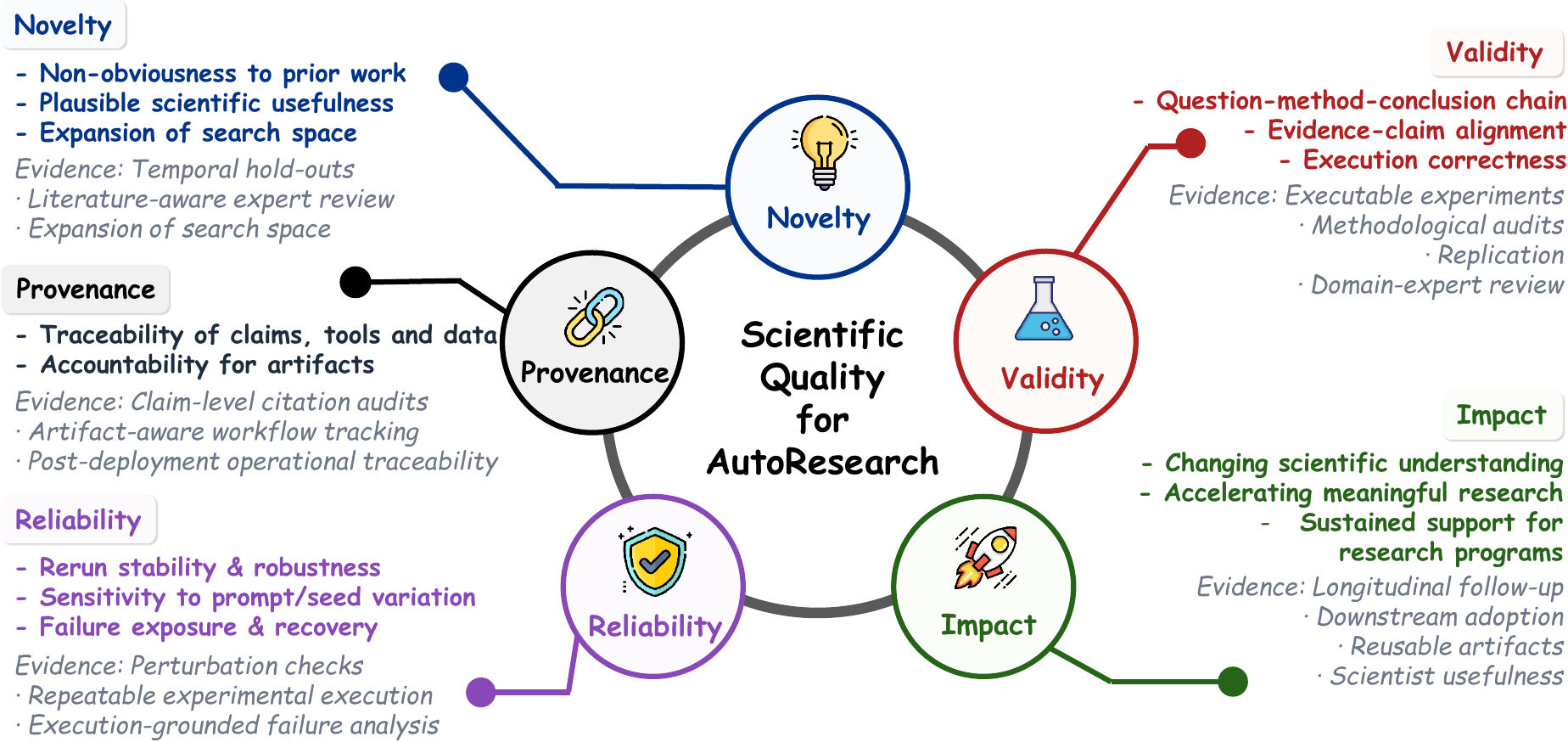}
\caption{\textbf{Scientific-quality dimensions and evidence instruments in AutoResearch.}
The figure separates five judgment targets-novelty, validity, impact, reliability, and provenance-from the evidence instruments used to support them, such as benchmarks, expert review, reruns, artifact tracing, and longitudinal follow-up.}
\label{fig:five_metric_framework}
\end{figure*}

Scientific quality is developed here through five jointly necessary dimensions: \textbf{novelty}, \textbf{validity}, \textbf{impact}, \textbf{reliability}, and \textbf{provenance}. These dimensions are the \emph{judgment targets}; benchmarks, expert review, reruns, artifact tracing, and longitudinal follow-up are the \emph{evidence instruments}. Figure~\ref{fig:five_metric_framework} summarizes this distinction. Its importance is methodological rather than merely descriptive: benchmark success is not itself a scientific judgment, but only one source of evidence that may or may not support one component of scientific quality. The central evaluation error in current AutoResearch practice is therefore not simply under-measurement, but \emph{authority borrowing}: a system is often presented as if strength on one dimension implied strength on all the others.
The five dimensions should therefore be read not as a loose checklist, but as a professional vocabulary for scientific judgment. Each dimension anchors a distinct evaluative question, demands different forms of evidence, and fails in a different way when overclaimed. Their role is not to multiply criteria for their own sake, but to prevent the reduction of scientific evaluation to surface fluency, one-shot benchmark success, or workflow completion.

\begin{itemize}[nolistsep, leftmargin=*]

    \item[] \textcolor[HTML]{4C78A8}{\largedot} \textit{\ul{Novelty.}} Novelty asks whether a workflow advances scientific understanding beyond literature-adjacent recombination. In rigorous terms, novelty is not exhausted by unfamiliarity. It requires some combination of non-obviousness relative to prior work, plausible scientific usefulness, and the capacity to open search directions that experts regard as worth pursuing. The corresponding evaluation problem is therefore literature-relative and temporally sensitive: a claim may appear novel only because the comparison class is weak, the retrieval base is incomplete, or the judgment ignores frontier recency. \textit{ResearchBench} is important here because it evaluates inspiration retrieval, hypothesis composition, and hypothesis ranking as structured components of discovery rather than reducing novelty to generic generation quality. \textit{ResearcherBench} tightens the same criterion under deeper literature and synthesis pressure, while \textit{BioDisco} adds temporal and expert-mediated assessment that is closer to how scientific novelty is actually judged in practice~\citep{Liu2025ResearchBench, Undermind2025ResearcherBenchEvaluatingDee, BioDisco2025}. Human evaluations of AI-generated research ideas further reinforce that plausible ideation does not automatically imply durable originality. In collaborative settings such as AI co-scientist, the stronger and more defensible novelty claim is often that AI expands the scientist's search space rather than that it independently secures a contribution~\citep{Undermind2024CanLLMsGenerateNovelResearch, gottweis2025towards}. Novelty is thus best understood as a frontier-sensitive and expert-mediated dimension, not as a rhetorical property of outputs.

    \item[] \textcolor[HTML]{F58518}{\largedot} \textit{\ul{Validity.}} Validity is the decisive scientific criterion because it asks whether the full question--method--execution--conclusion chain is warranted. This includes methodological adequacy, execution correctness, evidence--claim alignment, and the legitimacy of the resulting inference. A workflow can be original, coherent, and even reproducible in a narrow sense while still failing validity if its conclusions are unsupported, its controls are weak, or its analysis is mismatched to the question it claims to answer. \textit{BioDSA-1K} is especially valuable because it evaluates biomedical hypothesis validation through hypothesis decision accuracy, evidence--conclusion alignment, reasoning correctness, executability, and explicitly non-verifiable cases in which the available data do not warrant a conclusion~\citep{Wang2025BioDSA}. \textit{EXP-Bench} extends the same concern to end-to-end research experiments, while \textit{PaperBench} and \textit{SciReplicate-Bench} reinforce that repeatable execution and reproducible scientific warrant are inseparable in stronger validation settings~\citep{Kon2025EXPBench, Starace2025PaperBench, SciRepBench2025}. \textit{BLADE} and \textit{ControlA} show that validity can also erode through workflow-control failure, weak statistical discipline, and execution drift even when a system appears operationally competent~\citep{Gu2024BLADE, Gueroudji_2025}. Validity is therefore the dimension most directly tied to the limits of automatic verification: current systems still lack robust mechanisms for ensuring that workflow outputs can be checked reliably against the evidential standards they purport to satisfy.

    \item[] \textcolor[HTML]{54A24B}{\largedot} \textit{\ul{Impact.}} Impact asks whether a workflow matters beyond local task completion. A workflow can be novel and valid without changing scientific understanding, accelerating meaningful research work, or producing artifacts that domain experts regard as useful over time. Impact is therefore the dimension least well captured by standard benchmarks and the one most easily replaced by weak proxies such as leaderboard movement, paper acceptance, or polished presentation. More defensible evidence comes from scientist-facing usefulness, reusable artifacts, downstream adoption, and longitudinal follow-up. \textit{BioDisco} and \textit{AI co-scientist} are useful here because they move evaluation toward expert-facing utility and domain-relevant follow-up rather than pipeline completion alone~\citep{BioDisco2025, gottweis2025towards}. \textit{LiveResearchBench} and related collaboration-oriented settings point in the same direction by treating value as sustained support for a research program rather than as one-shot assistance~\citep{Wang2025LiveResearchBench, Li2025Build}. \textit{AI Scientist-v2} moved the discussion forward by linking automated research systems to workshop-level outcomes and stronger internal search structure, but even there, impact claims remain substantially weaker than claims about local task success~\citep{Yamada2025AIScientistV2}. Realism-oriented critiques, including \emph{How Far Are AI Scientists from Changing the World?}, make the same point more directly: apparent capability and field-level effect can diverge sharply~\citep{Xie2025How}. Impact should therefore be read as a downstream-science dimension, not as a simple reward for workflow completion.

    \item[] \textcolor[HTML]{B279A2}{\largedot} \textit{\ul{Reliability.}} Reliability asks whether a workflow behaves consistently enough to be trusted as a research instrument. It includes rerun stability, sensitivity to prompt or seed variation, robustness under tool or environment noise, and the ability to detect, expose, and recover from failure. The relevant question is not whether a system can succeed once, but whether it behaves stably enough under repetition and perturbation for its outputs to support stronger scientific claims. \textit{MLAgentBench} and \textit{PaperBench} are useful because they expose systems to repeatable but demanding experimental and replication tasks rather than one-shot successes~\citep{huang2023mlagentbench, Starace2025PaperBench}. \textit{SPOT}, \textit{FIRE-Bench}, and \textit{AIRS-Bench} sharpen the same issue by foregrounding verification weakness, discovery-suite instability, and full-lifecycle failure exposure rather than merely counting successful runs~\citep{SPOT2025ScientificPaperErrorDetection, Undermind2026FIREBenchEvaluatingAgentsont, Chen2025AIRSBench}. \textit{ControlA} remains relevant because fragile workflow control can produce apparent progress without stable operational backing~\citep{Gueroudji_2025}. Reliability is particularly important for AI-led workflows because a single fluent end-to-end run can create a misleading narrative of capability that disappears under reruns, changed seeds, altered tools, or stricter auditing.

    \item[] \textcolor[HTML]{7F7F7F}{\largedot} \textit{\ul{Provenance.}} Provenance asks whether claims, data use, intermediate artifacts, reviews, revisions, and deployment-time interventions remain traceable after the workflow ends. It links evaluation to accountability. A workflow may produce outputs that look convincing while still remaining untrustworthy if later readers cannot reconstruct what evidence was used, what tools were called, what interventions occurred, or where a conclusion actually came from. \textit{CiteME} is central because it evaluates citation attribution directly and shows that claim support cannot be reduced to citation formatting alone~\citep{CiteME2025}. \textit{LitSearch} shows that even literature-facing retrieval quality materially affects downstream trustworthiness when support tracking is weak~\citep{LitSearch2024}. Review-oriented resources such as \textit{LLM-REVal} show that weak traceability distorts not only research production but also research assessment~\citep{Undermind2025LLMREValCanWeTrustLLMReviewe}. Deployment-oriented documentation efforts such as the \textit{AI Agent Index} extend provenance beyond generation by emphasizing transparency, technical documentation, and governance visibility in deployed agentic systems~\citep{Undermind2026The2025AIAgentIndexDocumenti}. Together with artifact-aware systems and paper-level decomposition benchmarks such as \textit{AI-Researcher} and \textit{PaperBench}, these resources show that output quality without traceability remains insufficient for scientific trust~\citep{Tang2025AIResearcher, Starace2025PaperBench}. Provenance is therefore not an auxiliary bookkeeping concern; it is the condition under which correction, responsibility, and post hoc verification remain possible.

\end{itemize}

Taken jointly, these five dimensions block the most common evaluation error in this literature: borrowing the authority of one scientific virtue to imply the presence of the others. Novelty without validity is speculation. Validity without provenance is difficult to trust. Reliability without impact is operational competence without scientific importance. Provenance without novelty or validity is accountability without discovery.
However, scientific quality is only one side of the evaluation problem. AutoResearch also requires \textbf{autonomy assessment}: not a second scientific-quality checklist, but a workflow-sensitive judgment about what kind of autonomy claim the observed evidence actually supports. The decisive variables here are \textbf{task substitution}, \textbf{decision authority}, \textbf{workflow closure}, and \textbf{responsibility retention}~\citep{Zheng2025Agent4S, Undermind2026TheAgenticResearcherAPractic, gottweis2025towards}. A system may score well on several scientific-quality dimensions while remaining strongly human-steered if humans still define the agenda, interpret the results, and absorb final responsibility. Conversely, a workflow may appear highly autonomous in operational terms while still failing the scientific-quality frame. These variables are especially important for distinguishing advanced human-verified pipelines from mature AI-led workflows. Pipeline breadth may increase task substitution, but it does not establish AI-led autonomy unless decision authority, rejection, stopping, and responsibility are also redistributed away from routine human verification.

\begin{tcolorbox}[archquestionbox,
title={What autonomy assessment adds beyond scientific quality}]
Autonomy assessment determines what kind of system claim is warranted by the observed workflow. \textbf{Task substitution} identifies which stages are genuinely carried by the system rather than merely accelerated; without this distinction, bounded assistance is easily redescribed as autonomy. \textbf{Decision authority} identifies who actually controls framing, branch selection, rejection, revision, and stopping; without this distinction, surface automation can conceal deeper human steering. \textbf{Workflow closure} identifies whether grounding, planning, execution, validation, and reporting form a coherent loop or remain a collection of strong local segments; without this distinction, local competence is too easily mistaken for end-to-end capability. \textbf{Responsibility retention} identifies who remains accountable for unsafe transitions, unsupported claims, and downstream correction; without this distinction, stronger autonomy rhetoric can outpace the governance conditions under which the workflow still operates. These variables do not replace scientific quality. They prevent a different but equally consequential error: conflating what a workflow achieves with how much of that workflow is actually exercised---and answerable for---by the system itself.
\end{tcolorbox}

\subsection{Evaluation Instruments and Benchmark Landscape}
\label{sec:evaluation_instruments_benchmarks}

Current evaluation resources for AutoResearch do not yet constitute a unified benchmark regime. They instead form a heterogeneous instrument stack in which different resources test different parts of the scientific workflow, impose different evidential burdens, and operationalize different notions of scientific quality. This fragmentation is not merely a symptom of an immature field. It reflects a deeper methodological property of AutoResearch evaluation: novelty, validity, impact, reliability, and provenance cannot be reduced to a single task format, model score, or judge protocol. A system that performs well on research ideation may still fail during execution; a system that reproduces a paper may not generate valuable new directions; and a system that writes a coherent report may still lack source-faithful attribution or deployment-time accountability.

Table~\ref{tab:benchmark_taxonomy_autoresearch} organizes the current landscape from this instrument-oriented perspective. We retain resources whose primary contribution is a directly usable evaluation instrument for research agents, deep-research systems, scientific workflow automation, or accountability-sensitive research assistance. Rather than grouping benchmarks only by topic or domain, the table groups them by evaluative role. Discovery benchmarks primarily test structured search, decomposition, and frontier-facing ideation; experimentation and verification benchmarks stress executable research behavior, replication, and empirical discipline; deep-research and synthesis benchmarks evaluate long-horizon retrieval, report construction, and open-world information integration; and review- and provenance-oriented instruments target claim support, citation traceability, reviewer reliability, and deployment documentation. This organization highlights a central feature of the field: current benchmarks are complementary constraints on AutoResearch behavior, not substitutes for one another.

\begin{table*}[t]
\centering
\surveytablecaption{Evaluation instruments and benchmark landscape for AutoResearch.}{The table summarizes representative instruments whose primary contribution is a directly usable evaluation resource for AutoResearch workflows. Novel. = novelty; Valid. = validity; Reliab. = reliability; Proven. = provenance. Filled circles denote primary coverage, while hollow circles denote secondary or indirect coverage. The table does not assign autonomy levels to benchmarks; it indicates which dimensions each instrument can directly or indirectly support.}
\label{tab:benchmark_taxonomy_autoresearch}
\renewcommand{\arraystretch}{1.15}
\setlength{\tabcolsep}{4pt}
\rowcolors{2}{white}{tabgray}
\resizebox{\textwidth}{!}{%
\begin{tabular}{@{}l c c c c c c c l c@{}}
\toprule[1pt]
\textbf{Instrument} & \textbf{Role} & \textbf{Novel.} & \textbf{Valid.} & \textbf{Impact} & \textbf{Reliab.} & \textbf{Proven.} & \textbf{Unit} & \textbf{Domain} & \textbf{Year} \\
\midrule

\rowcolor{evalblue!10}
\multicolumn{10}{@{}c}{\textit{Discovery benchmarks}} \\
Auto-Bench~\cite{Chen2025Auto} & \roleBenchmark & \fullsup & \lowsup & \lowsup & \lowsup & \lowsup & Agent run & General & 2025 \\
DiscoveryBench~\cite{Majumder2024DiscoveryBench} & \roleBenchmark & \fullsup & \lowsup & \lowsup & \lowsup & \lowsup & Discovery task & General & 2024 \\
ResearchBench~\cite{Liu2025ResearchBench} & \roleBenchmark & \fullsup & \lowsup & \lowsup & \lowsup & \lowsup & Research task & General & 2025 \\
ResearcherBench~\cite{Undermind2025ResearcherBenchEvaluatingDee} & \roleBenchmark & \fullsup & \lowsup & \lowsup & \lowsup & \lowsup & Frontier research & AI research & 2025 \\
AIRS-Bench~\cite{Chen2025AIRSBench} & \roleBenchmark & \fullsup & \lowsup & \lowsup & \fullsup & \lowsup & Research suite & General ML research & 2026 \\

\cmidrule(l){1-10}
\rowcolor{evalorange!10}
\multicolumn{10}{@{}c}{\textit{Experimentation and verification}} \\
MLAgentBench~\cite{huang2023mlagentbench} & \roleBenchmark & \lowsup & \fullsup & \lowsup & \fullsup & \lowsup & ML experiment & Machine learning & 2023 \\
EXP-Bench~\cite{Kon2025EXPBench} & \roleBenchmark & \lowsup & \fullsup & \lowsup & \fullsup & \lowsup & Experiment task & AI research & 2025 \\
PaperBench~\cite{Starace2025PaperBench} & \roleBenchmark & \lowsup & \fullsup & \lowsup & \fullsup & \fullsup & Paper replication & Machine learning & 2025 \\
CORE-Bench~\cite{Siegel2024CORE} & \roleBenchmark & \lowsup & \fullsup & \lowsup & \fullsup & \fullsup & Reproducibility task & Mixed science & 2024 \\
ScienceAgentBench~\cite{Chen2024ScienceAgentBench} & \roleBenchmark & \lowsup & \fullsup & \lowsup & \fullsup & \lowsup & Scientific task & General science & 2024 \\
SciReplicate-Bench~\cite{SciRepBench2025} & \roleBenchmark & \lowsup & \fullsup & \lowsup & \fullsup & \lowsup & Reproduction task & NLP research & 2025 \\
SPOT~\cite{SPOT2025ScientificPaperErrorDetection} & \roleBenchmark & \lowsup & \fullsup & \lowsup & \fullsup & \lowsup & Paper verification & General science & 2025 \\
FIRE-Bench~\cite{Undermind2026FIREBenchEvaluatingAgentsont} & \roleBenchmark & \lowsup & \fullsup & \lowsup & \fullsup & \lowsup & Rediscovery task & AI/ML research & 2026 \\

\cmidrule(l){1-10}
\rowcolor{evalgreen!10}
\multicolumn{10}{@{}c}{\textit{Deep research and synthesis}} \\
BioDSA-1K~\cite{Wang2025BioDSA} & \roleBenchmark & \lowsup & \fullsup & \lowsup & \lowsup & \lowsup & Hypothesis task & Biomedicine & 2025 \\
DeepScholar-Bench~\cite{Patel2025DeepScholar} & \roleBenchmark & \lowsup & \lowsup & \lowsup & \lowsup & \lowsup & Related-work synthesis & Scientific literature & 2025 \\
LiveResearchBench~\cite{Wang2025LiveResearchBench} & \roleBenchmark & \lowsup & \lowsup & \fullsup & \fullsup & \lowsup & Research task & General / web research & 2025 \\
Deep Research Bench~\cite{Bosse2025Deep} & \roleBenchmark & \lowsup & \lowsup & \lowsup & \fullsup & \lowsup & Deep research task & General & 2025 \\
DeepResearch Arena~\cite{Wan2025DeepResearch} & \roleBenchmark & \lowsup & \lowsup & \lowsup & \lowsup & \lowsup & Seminar-grounded task & General & 2025 \\
DRBench~\cite{Abaskohi2025DRBench} & \roleBenchmark & \lowsup & \lowsup & \lowsup & \lowsup & \lowsup & Enterprise research task & Enterprise / web + private KB & 2025 \\

\cmidrule(l){1-10}
\rowcolor{evalpurple!10}
\multicolumn{10}{@{}c}{\textit{Review and provenance}} \\
CiteME~\cite{CiteME2025} & \roleAudit & \lowsup & \lowsup & \lowsup & \lowsup & \fullsup & Citation attribution & Scientific literature & 2024 \\
LitSearch~\cite{LitSearch2024} & \roleBenchmark & \lowsup & \lowsup & \lowsup & \lowsup & \fullsup & Retrieval query & Scientific literature & 2024 \\
LLM-REVal~\cite{Undermind2025LLMREValCanWeTrustLLMReviewe} & \roleReview & \lowsup & \lowsup & \lowsup & \lowsup & \fullsup & Review trace & General science & 2025 \\
AI Agent Index~\cite{Undermind2026The2025AIAgentIndexDocumenti} & \roleAudit & \lowsup & \lowsup & \lowsup & \fullsup & \fullsup & Deployed agent profile & General agents & 2025 \\

\bottomrule[1pt]
\end{tabular}
}
\end{table*}

\begin{itemize}[nolistsep, leftmargin=*]

    \item[] \textcolor[HTML]{4C78A8}{\largedot} \textbf{Evaluation has shifted from task accuracy to workflow-specific scientific burden.} Early scientific-agent evaluation often borrowed the logic of conventional benchmark design: define a task, score the output, and compare systems by aggregate performance. The newer AutoResearch landscape moves beyond this pattern. Discovery-oriented instruments such as \textit{DiscoveryBench}~\cite{Majumder2024DiscoveryBench}, \textit{ResearchBench}~\cite{Liu2025ResearchBench}, \textit{ResearcherBench}~\cite{Undermind2025ResearcherBenchEvaluatingDee}, and \textit{AIRS-Bench}~\cite{Chen2025AIRSBench} evaluate whether agents can structure search, formulate research directions, or operate over frontier-facing research tasks. Their central burden is not answer correctness alone, but whether the system can organize plausible scientific exploration under constraints of novelty and relevance. This makes them useful for assessing the upper layers of research cognition, but insufficient for determining whether a workflow can produce empirically reliable science.

    \item[] \textcolor[HTML]{F58518}{\largedot} \textbf{Execution-centered benchmarks impose stronger validity pressure, but remain domain-bounded.} Experimentation and verification instruments place a different burden on AutoResearch systems. \textit{MLAgentBench}~\cite{huang2023mlagentbench}, \textit{EXP-Bench}~\cite{Kon2025EXPBench}, \textit{ScienceAgentBench}~\cite{Chen2024ScienceAgentBench}, \textit{PaperBench}~\cite{Starace2025PaperBench}, \textit{CORE-Bench}~\cite{Siegel2024CORE}, and \textit{SciReplicate-Bench}~\cite{SciRepBench2025} move evaluation closer to executable behavior, reproduced results, or paper-level verification. These benchmarks are especially important because they expose the gap between producing a convincing research plan and carrying out scientifically disciplined work. At the same time, their realism is uneven: machine-learning and code-centric settings are more easily packaged into reproducible environments, while wet-lab, clinical, field-based, or socially interpreted research remains much harder to benchmark with comparable fidelity. Thus, execution-centered evaluation strengthens validity and reliability assessment, but it does not by itself solve the domain-general evaluation problem.

    \item[] \textcolor[HTML]{54A24B}{\largedot} \textbf{Deep-research benchmarks evaluate synthesis capacity, not full scientific closure.} Deep-research resources such as \textit{DeepScholar-Bench}~\cite{Patel2025DeepScholar}, \textit{LiveResearchBench}~\cite{Wang2025LiveResearchBench}, \textit{Deep Research Bench}~\cite{Bosse2025Deep}, \textit{DeepResearch Arena}~\cite{Wan2025DeepResearch}, and \textit{DRBench}~\cite{Abaskohi2025DRBench} capture another important frontier: the ability to search across large, changing, or private information spaces and produce structured, citation-aware reports. This is central to \textcolor{RefOrange}{L1--L2} research assistance and increasingly important for pipeline-oriented systems approaching the \textcolor{RefOrange}{L3} frontier, where literature grounding and report construction must persist across multiple steps. However, strong performance on deep-research tasks should not be mistaken for autonomous scientific discovery. These benchmarks primarily evaluate synthesis, coverage, citation use, and report usefulness; they usually do not require the system to generate new evidence, reject weak hypotheses through experimentation, or close the loop between claims and empirical validation.

    \item[] \textcolor[HTML]{B279A2}{\largedot} \textbf{Review and provenance instruments remain necessary because benchmark scores do not establish accountability.} Even when a system completes a discovery, replication, or deep-research benchmark, key accountability questions remain unresolved: which sources supported which claims, whether citations were faithful, whether review judgments were reliable, and whether deployed agents were documented well enough for external inspection. Instruments such as \textit{CiteME}~\cite{CiteME2025}, \textit{LitSearch}~\cite{LitSearch2024}, \textit{LLM-REVal}~\cite{Undermind2025LLMREValCanWeTrustLLMReviewe}, and the \textit{AI Agent Index}~\cite{Undermind2026The2025AIAgentIndexDocumenti} therefore occupy a distinct methodological role. They do not merely supplement performance benchmarks; they evaluate the traceability and inspectability conditions under which research automation can be trusted. This distinction is crucial for AutoResearch because scientific autonomy requires not only task success, but also defensible provenance, reviewability, and downstream accountability.

\end{itemize}

The current benchmark landscape therefore supports a cautious conclusion. AutoResearch evaluation is advancing from isolated output scoring toward a richer set of workflow-specific instruments, but it has not yet converged on a unified regime of automatic scientific verification. Discovery benchmarks, execution benchmarks, deep-research benchmarks, and provenance instruments each constrain a different failure mode. Their combination makes present evaluation more informative than any single benchmark family, but also reveals the central unresolved challenge: scientifically credible AutoResearch will require evaluation protocols that connect novelty, execution, validation, reliability, and provenance within the same workflow rather than measuring them as separate fragments. Until such coupled evaluation becomes reliable, benchmark success should be treated as evidence for bounded capability rather than as sufficient support for mature AI-led autonomy.

\section{Domains of AutoResearch}
\label{sec:domain}

AutoResearch progresses unevenly because scientific autonomy is constrained by the structure of the domain in which it is deployed. Across disciplines, the same agent architecture may correspond to different levels of autonomy depending on the manipulability of the research object, the cost and speed of feedback, the observability of intermediate states, the reversibility of interventions, and the accountability burden attached to the final claim. Code-native and formal domains provide executable artifacts, replayable environments, and comparatively explicit correctness signals, enabling stronger workflow closure. Empirical domains introduce embodiment, instrumentation, sample variability, and delayed validation. Clinical and social domains further impose regulatory, ethical, causal, and interpretive constraints that cannot be absorbed by automation alone. Thus, domain is not a peripheral application context for AutoResearch, but a determining factor in its autonomy ceiling~\citep{Zheng2025Agent4S, ZHENG2025Automation, wei2025ai, Pauloski_2025, Gridach2025Agentic, Tie2025Survey}.

\begin{figure*}[ht]
\centering
\includegraphics[width=1\linewidth]{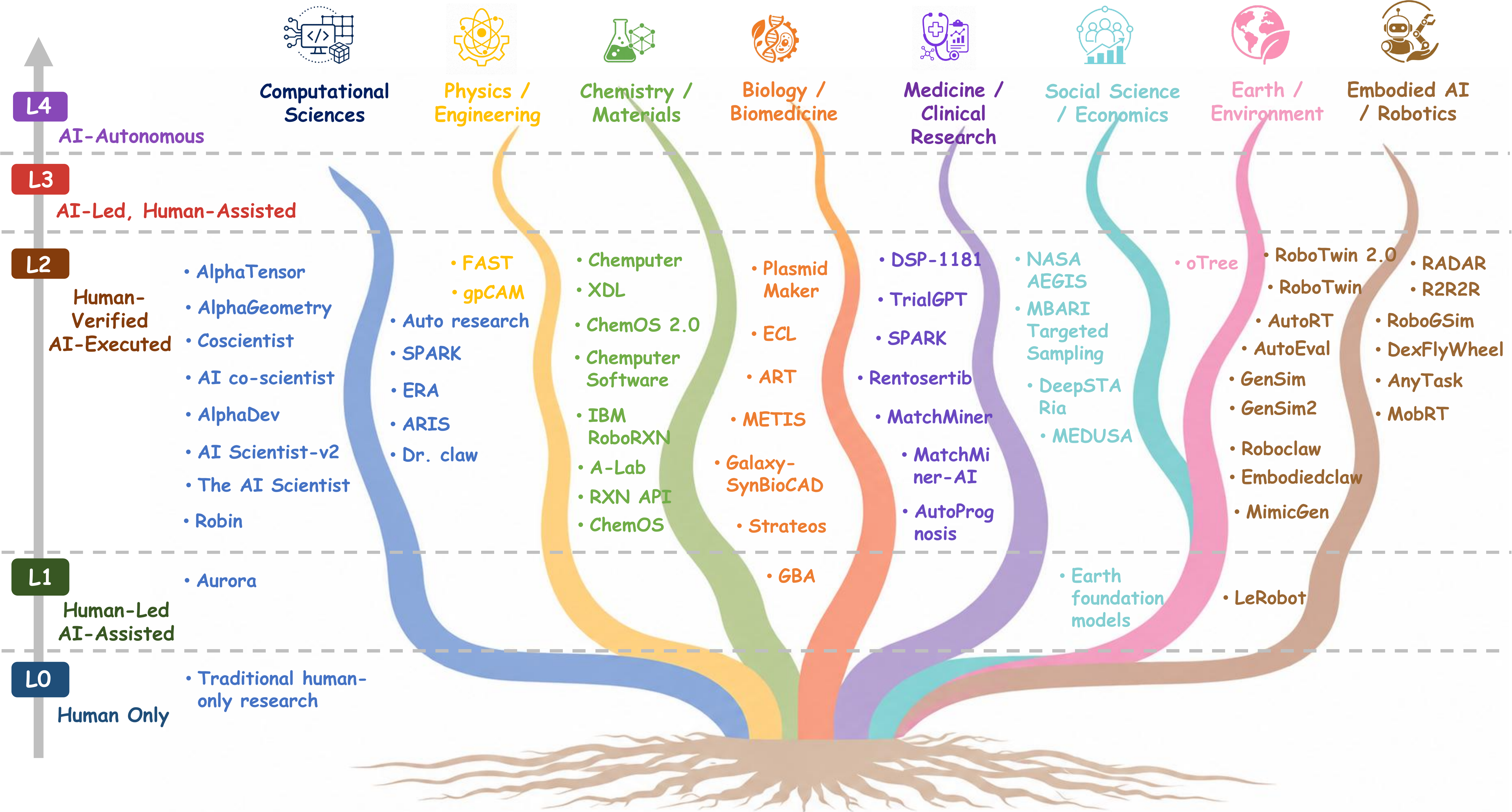}
\caption{\textbf{Domain-conditioned autonomy ceilings in AutoResearch.}
The figure summarizes how major scientific domains differ in their current autonomy center of gravity within the \textcolor{RefOrange}{L0--L4} framework. Code-native and formally checkable domains are closer to higher automation because execution and validation can be more readily closed, whereas empirical, clinical, social, Earth-scale, and embodied domains remain constrained by embodiment, accountability, causal validity, delayed verification, and non-manipulability.}
\label{fig:domain_level_profile}
\end{figure*}

Figure~\ref{fig:domain_level_profile} summarizes this domain-conditioned landscape by positioning major scientific areas along the \textcolor{RefOrange}{L0--L4} spectrum. The profile shows a clear gradient from computationally closed domains to empirically and socially constrained domains. Computational and formal sciences occupy the highest current position because their workflows can be represented, executed, verified, and repeated within digital environments. Physical sciences, chemistry and materials, and selected areas of biology form an intermediate band where simulators, robotic platforms, and closed-loop optimization support partial AI-led workflow closure, but only within bounded design spaces and instrument regimes. Medicine, economics and social sciences, and Earth/environmental sciences remain more constrained: their workflows may be data-rich and increasingly AI-assisted, yet their claims depend on patient safety, causal interpretation, governance, non-manipulable systems, or long-horizon validation. The resulting pattern indicates that current AutoResearch exhibits advanced L2 workflow automation in favorable substrates, while no domain yet provides convincing evidence for robust L3 or L4 scientific autonomy.

\subsection{Computational and Formal Sciences}
\label{sec:domain_computational}

Computational and formal sciences currently define the highest observed center of gravity of AutoResearch. Their advantage lies in the domain substrate: research artifacts are already digital, executable, replayable, and comparatively inspectable. Code repositories, datasets, benchmarks, simulators, logs, evaluation scripts, proof objects, and versioned outputs connect research actions to fast and explicit feedback. This substrate has made computational and formal sciences the primary testbed for general AutoResearch systems that couple literature grounding, idea generation, implementation, experiment execution, result analysis, validation, and paper writing within a unified workflow.

\begin{table*}[h]
\centering
\surveytablecaption{Representative computational AutoResearch systems.}{The table maps representative computational AutoResearch systems to the five workflow stages. The emphasis is on general workflow-level systems operating in computational, machine-learning, or code-native research environments. Stages: I = Literature \& Research Grounding, II = Hypothesis \& Planning, III = Experimentation \& Tool Use, IV = Feedback / Validation / Review, V = Reporting \& Knowledge Communication. \fullmark\ = stage substantively covered; \halfmark\ = partial or human-assisted; -- = not covered.}
\label{tab:computational_systems}
\renewcommand{\arraystretch}{1.10}
\setlength{\tabcolsep}{2.4pt}
\rowcolors{2}{white}{black!2}
\begin{adjustbox}{max width=\textwidth}
\small
\begin{tabularx}{\textwidth}{@{}p{3.65cm} p{0.5cm} p{0.5cm} p{0.5cm} p{0.5cm} p{0.5cm} p{3.65cm} X@{}}
\toprule[1pt]
\textbf{System / Project} & \textbf{I} & \textbf{II} & \textbf{III} & \textbf{IV} & \textbf{V} & \textbf{Setting} & \textbf{Workflow Emphasis} \\
\midrule

The AI Scientist~\citep{Lu2024AIScientist}
 & \fullmark & \fullmark & \fullmark & \fullmark & \fullmark
 & \makecell[l]{ML research}
 & End-to-end research loop \\

AI Scientist-v2~\citep{Yamada2025AIScientistV2}
 & \fullmark & \fullmark & \fullmark & \fullmark & \fullmark
 & \makecell[l]{ML research}
 & Long-horizon scientist workflow \\

AI-Researcher~\citep{Tang2025AIResearcher}
 & \fullmark & \fullmark & \fullmark & \halfmark & \fullmark
 & \makecell[l]{General AI research}
 & Literature-to-paper pipeline \\

CodeScientist~\citep{Jansen2025CodeScientist}
 & \halfmark & \fullmark & \fullmark & \fullmark & \halfmark
 & \makecell[l]{Code-centric research}
 & Code-native iteration \\

DeepScientist~\citep{Weng2025DeepScientist}
 & \fullmark & \fullmark & \fullmark & \fullmark & \halfmark
 & \makecell[l]{Frontier AI research}
 & Long-horizon discovery \\

Agent Laboratory~\citep{AgentLaboratoryGitHub}
 & \fullmark & \fullmark & \fullmark & \halfmark & \fullmark
 & \makecell[l]{Computational research}
 & Human-piloted workflow \\

NanoResearch~\citep{nanoresearch2026}
 & \fullmark & \fullmark & \fullmark & \halfmark & \fullmark
 & \makecell[l]{Autonomous workflow}
 & Compact research loop \\

ARIS~\citep{ARISGitHub}
 & \halfmark & \fullmark & \fullmark & \fullmark & \halfmark
 & \makecell[l]{Research infrastructure}
 & Research orchestration \\

\bottomrule
\end{tabularx}
\end{adjustbox}
\end{table*}

\begin{itemize}[nolistsep, leftmargin=*]

    \item[] \textcolor[HTML]{4C78A8}{\largedot} \textbf{Executable substrate.}
    This domain shortens the distance between action and verification. Candidate ideas can be instantiated as code changes, training runs, benchmark submissions, proof steps, or simulation calls, while feedback emerges from compilation, runtime traces, unit tests, formal checkers, benchmark scores, and reproducibility scripts. Such substrate conditions support rapid iteration, state preservation, alternative comparison, and failure inspection. The autonomy advantage is therefore substrate-level rather than task-level: computational and formal artifacts are more readily represented, executed, replayed, and audited than those in wet-lab, clinical, social, or Earth-scale settings.

    \item[] \textcolor[HTML]{F58518}{\largedot} \textbf{Workflow-level systems.}
    General AutoResearch systems are most densely instantiated in computational and code-native research because these settings allow agents to connect proposal, implementation, execution, evaluation, and reporting within the same digital loop. \textit{The AI Scientist} establishes the canonical end-to-end pattern from idea generation to code execution, paper drafting, and review-style feedback~\citep{Lu2024AIScientist}, while \textit{AI Scientist-v2} extends this line toward longer-horizon scientific search and stronger planning--validation coupling~\citep{Yamada2025AIScientistV2}. \textit{AI-Researcher} frames research as an integrated pipeline from literature review and hypothesis generation to experimentation and manuscript preparation~\citep{Tang2025AIResearcher}, and \textit{CodeScientist} sharpens the code-native side through iterative implementation, debugging, execution, and result refinement~\citep{Jansen2025CodeScientist}. \textit{DeepScientist} further emphasizes long-horizon discovery with memory over intermediate findings and progressive validation~\citep{Weng2025DeepScientist}. Infrastructure-oriented projects extend this operating model: \textit{Agent Laboratory} supports human-piloted computational research from user idea to report generation~\citep{AgentLaboratoryGitHub}, \textit{NanoResearch} packages a compact autonomous research loop~\citep{nanoresearch2026}, and \textit{ARIS} emphasizes orchestration, execution management, validation support, and workflow persistence~\citep{ARISGitHub}. Together, these systems position computational research as the main experimental substrate for general research-agent architectures.

    \item[] \textcolor[HTML]{54A24B}{\largedot} \textbf{Autonomy boundary.}
    Computational and formal sciences support selective \textcolor{RefOrange}{L2} behavior in favorable settings, but not robust \textcolor{RefOrange}{L3, L4} autonomy. Their strengths concentrate in executable production and local evaluation, while their weaknesses remain in problem selection, significance assessment, termination criteria, and scientific positioning. Current systems can generate hypotheses, edit code, execute experiments, compare metrics, and draft papers, yet they remain less reliable at selecting consequential problems, identifying scientifically meaningful gains, terminating unpromising directions, and situating findings within broader research programs. The central boundary is the gap between executable feedback and scientific significance: code-native environments can indicate whether an artifact runs or a metric improves, but not whether the resulting claim is important, generalizable, or scientifically well motivated.

\end{itemize}

\subsection{Physical Sciences and Engineering}
\label{sec:domain_physical}

Physical sciences and engineering occupy an intermediate position in the AutoResearch landscape: stronger than high-accountability clinical and social domains, but less closed than code-native computational research. Their autonomy profile is shaped by the divide between \emph{simulation-native} and \emph{instrument-native} workflows. Governing equations, simulator states, numerical solvers, structured design spaces, and explicit validation targets support agentic planning, execution, and revision in digital or semi-digital environments. Empirical closure, however, remains tied to apparatus, calibration, measurement, physical robustness, and environmental control. The domain therefore sits in advanced \textcolor{RefOrange}{L2}: simulation-heavy workflows can support narrow workflow automation, while instrument-coupled autonomy remains bounded by laboratory integration.

\begin{table*}[h]
\centering
\surveytablecaption{Representative physical-science and engineering AutoResearch systems.}{Representative systems spanning symbolic discovery, computational physics workflows, and bounded instrument-coupled autonomy. Stage notation follows Table \ref{tab:computational_systems}.}
\label{tab:physical_science_systems}
\renewcommand{\arraystretch}{1.10}
\setlength{\tabcolsep}{2.4pt}
\rowcolors{2}{white}{black!2}
\begin{adjustbox}{max width=\textwidth}
\small
\begin{tabularx}{\textwidth}{@{}p{3.65cm} p{0.5cm} p{0.5cm} p{0.5cm} p{0.5cm} p{0.5cm} p{3.65cm} X@{}}
\toprule[1pt]
\textbf{System} & \textbf{I} & \textbf{II} & \textbf{III} & \textbf{IV} & \textbf{V} & \textbf{Subdomain} & \textbf{Workflow Emphasis} \\
\midrule

AI Feynman~\citep{Udrescu2020AIPhysRev}
 & -- & \fullmark & \fullmark & \fullmark & --
 & \makecell[l]{Symbolic physics}
 & Equation discovery \\

PhysMaster~\citep{Miao2025PhysMaster}
 & \fullmark & \fullmark & \fullmark & \halfmark & \halfmark
 & \makecell[l]{Theoretical and\\ computational physics}
 & Physics research agent \\

QuantumAgent SDL~\citep{Cao2024QuantumAgentSDL}
 & -- & \halfmark & \fullmark & \fullmark & --
 & \makecell[l]{Quantum devices}
 & Closed-loop lab control \\

\bottomrule
\end{tabularx}
\end{adjustbox}
\end{table*}

\begin{itemize}[nolistsep, leftmargin=*]

    \item[] \textcolor[HTML]{4C78A8}{\largedot} \textbf{Simulation-native substrate.}
    The strongest autonomy in this domain appears where scientific actions are expressed as symbolic recovery, numerical simulation, parameter search, model fitting, or solver-mediated validation. Candidate hypotheses and designs can be evaluated through equations, simulators, computational experiments, and reproducible numerical pipelines rather than slow physical intervention. This makes physics and engineering more formalizable than medicine or social science, while still less frictionless than computational research because validation remains tied to modeling assumptions, solver fidelity, measurement protocols, and physical confirmation.

    \item[] \textcolor[HTML]{F58518}{\largedot} \textbf{Main systems.}
    Current systems concentrate around model-native discovery and bounded experimental control. \textit{AI Feynman} represents symbolic physical discovery through equation recovery from data under structured priors~\citep{Udrescu2020AIPhysRev}. \textit{PhysMaster} moves toward contemporary AutoResearch by coupling physics-oriented literature retrieval, abstract reasoning, numerical computation, and task execution in theoretical and computational physics workflows~\citep{Miao2025PhysMaster}. \textit{QuantumAgent SDL} represents instrument-coupled autonomy, where procedures, state transitions, experiment execution, and result analysis are organized into a closed-loop calibration workflow for quantum devices~\citep{Cao2024QuantumAgentSDL}. These systems position physical AutoResearch around simulation-native reasoning and tightly instrumented experimental loops.

    \item[] \textcolor[HTML]{54A24B}{\largedot} \textbf{Autonomy boundary.}
    The domain's ceiling is set by the gap between model-side closure and physical-side closure. Simulation-native systems can automate planning, execution, and validation inside bounded numerical environments, but this does not generalize to open-ended physical experimentation. Instrument-native workflows remain constrained by calibration drift, apparatus heterogeneity, data-reduction pipelines, measurement uncertainty, and limited protocol portability. Physical sciences and engineering therefore rise above bounded assistance where simulators or tightly controlled instruments provide reliable feedback, but they do not yet support broad autonomous workflow closure across open-ended experimental practice.

\end{itemize}

\subsection{Embodied Intelligence}
\label{sec:domain_embodied}

Embodied intelligence occupies a distinctive position in the AutoResearch landscape because its research loop sits between pure software research and fully physical experimentation. Unlike code-native domains, embodied research depends on simulation assets, task programs, robot embodiments, trajectory datasets, scene resets, and benchmarkable evaluation pipelines. Unlike wet-lab science, however, a large fraction of this loop can still be executed inside simulators or tightly instrumented robot platforms. The most mature form of autonomy in this domain is therefore not autonomous scientific discovery, but the automation of research infrastructure: task and scene generation, trajectory synthesis, real-to-sim data expansion, policy evaluation, and reproducible workflow execution~\citep{Zho26, Li26, Ahn24, Zho25, Mu25, Che25}.

\begin{table*}[h]
\centering
\surveytablecaption{Representative AutoResearch systems for embodied intelligence.}{The table maps representative embodied AutoResearch systems to the five workflow stages. The emphasis is on systems that automate embodied-AI development, robot-learning workflows, task and data generation, digital-twin construction, real-to-sim data factories, and reproducible experimentation infrastructure. Stage notation follows Table \ref{tab:computational_systems}.}
\label{tab:embodied_systems}
\renewcommand{\arraystretch}{1.12}
\setlength{\tabcolsep}{3.6pt}
\rowcolors{2}{white}{black!2}

\newcolumntype{Y}{>{\centering\arraybackslash}p{0.48cm}}

\begin{adjustbox}{max width=\textwidth}
\small
\begin{tabularx}{\textwidth}{@{}p{3.05cm} Y Y Y Y Y >{\raggedright\arraybackslash}p{3.35cm} >{\raggedright\arraybackslash}X@{}}
\toprule[1pt]
\textbf{System / Platform} 
& \textbf{I} & \textbf{II} & \textbf{III} & \textbf{IV} & \textbf{V} 
& \textbf{Setting} 
& \textbf{Workflow Emphasis} \\
\midrule

EmbodiedClaw~\citep{Zho26}
& \halfmark & \fullmark & \fullmark & \fullmark & \halfmark
& \makecell[tl]{Embodied-AI\\development workflow}
& Environment creation, trajectory synthesis, and evaluation \\

RoboClaw~\citep{Li26}
& -- & \halfmark & \fullmark & \fullmark & --
& \makecell[tl]{Long-horizon robot\\learning lifecycle}
& Data collection, policy learning, and self-resetting execution \\

AutoRT~\citep{Ahn24}
& -- & \fullmark & \fullmark & \halfmark & --
& \makecell[tl]{Real-world robot\\fleet orchestration}
& Instruction generation and real-robot data collection \\

AutoEval~\citep{Zho25}
& -- & \halfmark & \fullmark & \fullmark & \halfmark
& \makecell[tl]{Real-world policy\\evaluation}
& Automated policy evaluation and scene reset \\

RoboTwin~\citep{Mu25}
& -- & \fullmark & \fullmark & \fullmark & --
& \makecell[tl]{Generative digital twins\\for dual-arm manipulation}
& Digital-twin generation and benchmark construction \\

RoboTwin 2.0~\citep{Che25}
& -- & \fullmark & \fullmark & \fullmark & --
& \makecell[tl]{Scalable bimanual\\data generator and benchmark}
& Scalable bimanual data generation and evaluation \\

\bottomrule
\end{tabularx}
\end{adjustbox}
\end{table*}

\begin{itemize}[nolistsep, leftmargin=*]

    \item[] \textcolor[HTML]{4C78A8}{\largedot} \textbf{Workflow automation rather than autonomous discovery.}
    The clearest signal from the recent literature is that embodied AutoResearch is becoming a workflow-automation layer above existing simulators and robot-learning stacks, not a system for end-to-end autonomous science. EmbodiedClaw turns high-frequency embodied development operations into executable conversational skills, including environment construction, benchmark transformation, trajectory synthesis, and evaluation~\citep{Zho26}. RoboClaw closes the loop further by coupling data collection, policy learning, and execution through self-resetting action structure, which reduces the manual burden of long-horizon robot learning~\citep{Li26}. AutoRT and AutoEval extend this pattern into real-robot settings by automating data collection and evaluation infrastructure at scale~\citep{Ahn24, Zho25}.

    \item[] \textcolor[HTML]{F58518}{\largedot} \textbf{Automatic task, trajectory, and dataset generation.}
    A second major trend is the automation of the core research artifacts used to train embodied systems. GenSim and GenSim2 use language-model-based generation to expand task sets, create simulation programs, and synthesize demonstrations at scale~\citep{Wan23b, Hua24}. RoboGen goes further toward a fully generative loop in which the system proposes tasks, builds simulation environments, selects supervision mechanisms, and learns skills with minimal human specification~\citep{Wan23d}. MimicGen and SkillMimicGen address a complementary bottleneck: they amplify a small number of human demonstrations into much larger training corpora by adapting demonstrations to new contexts and recomposing skills into new task instances~\citep{Man23, Gar24}. In this part of the landscape, the key automated object is no longer only the benchmark itself, but the pipeline that manufactures task diversity and usable trajectories.

    \item[] \textcolor[HTML]{F58518}{\largedot} \textbf{Digital twins and real-to-sim data factories.}
    The strongest recent systems in embodied AutoResearch are often data factories that connect real scenes to scalable simulation. RoboTwin and RoboTwin 2.0 are the clearest examples in this search: they combine generative digital twins, task-code synthesis, and benchmarkable data production to support dual-arm and bimanual manipulation research~\citep{Mu25, Che25}. Around them is a broader family of real-to-sim systems that automate the construction of transfer-relevant training worlds. Digital Cousins automatically creates semantically and geometrically analogous simulated environments for robust policy learning~\citep{Dai24}. R2R2R, RoboGSim, Re3Sim, and ReBot all reduce dependence on manual teleoperation or expensive hardware collection by reconstructing or replaying real scenes into scalable simulation pipelines, then using those pipelines for data generation, training, or evaluation~\citep{Yu25b, Li24c, Han25, Fan25}. This makes embodied intelligence one of the clearest domains where AutoResearch already functions as a research-engineering flywheel: real data seeds simulation, simulation scales data, and scaled data feeds model improvement.

    \item[] \textcolor[HTML]{54A24B}{\largedot} \textbf{Autonomy boundary.}
    The present autonomy ceiling in embodied intelligence is best described as selective workflow autonomy rather than robust scientific autonomy. Within bounded simulators, robot fleets, or standardized evaluation setups, current systems can already automate task generation, scene construction, trajectory collection, benchmark execution, and policy evaluation~\citep{Zho26, Li26, Ahn24, Zho25, Che25}. However, these systems still depend on human oversight for questions that matter scientifically: whether generated scenes preserve the right affordances, whether synthetic trajectories capture meaningful task structure, whether evaluation protocols reflect real capability, and whether sim-to-real transfer is trustworthy across hardware, sensing, and environment variation~\citep{Mu25, Dai24, Han25, Fan25}. Older infrastructure systems such as SIERRA, EMS, Robotarium, and the Real Robot Challenge help with reproducibility and scale, but they do not remove this validation burden~\citep{Har23, Lin23, Wil21, Bau21}. For that reason, embodied intelligence currently fits advanced-L2 AutoResearch more convincingly than L3 or L4 autonomy.

\end{itemize}

\subsection{Chemistry and Materials}
\label{sec:domain_chemistry}

Chemistry and materials provide the clearest empirical path toward narrow higher-autonomy AutoResearch. The domain combines structured scientific representations with increasingly mature robotic and computational execution: molecular graphs, reaction recipes, retrosynthesis plans, synthesis conditions, materials compositions, robotic action spaces, assay outputs, and characterization data can be encoded, optimized, and linked to feedback. This places the domain in advanced \textcolor{RefOrange}{L2}, with the strongest systems approaching but not yet establishing reliable \textcolor{RefOrange}{L3} autonomy. Its strongest autonomy appears where bounded design spaces, instrumented laboratories, high-throughput characterization, or computational screening connect candidate generation to experimental or model-side validation. Broad closure remains limited by physical synthesis, protocol transfer, failed-experiment integration, laboratory-specific reproducibility, and heterogeneous evidence.

\begin{table*}[h]
\centering
\surveytablecaption{Representative chemistry and materials AutoResearch systems.}{Representative domain-native systems spanning robotic chemistry, autonomous materials synthesis, computational materials discovery, chemistry tool use, and laboratory orchestration. Stage notation follows Table \ref{tab:computational_systems}.}
\label{tab:chemistry_systems}
\renewcommand{\arraystretch}{1.10}
\setlength{\tabcolsep}{2.4pt}
\rowcolors{2}{white}{black!2}
\begin{adjustbox}{max width=\textwidth}
\small
\begin{tabularx}{\textwidth}{@{}p{3.65cm} p{0.5cm} p{0.5cm} p{0.5cm} p{0.5cm} p{0.5cm} p{3.65cm} X@{}}
\toprule[1pt]
\textbf{System} & \textbf{I} & \textbf{II} & \textbf{III} & \textbf{IV} & \textbf{V} & \textbf{Subdomain} & \textbf{Workflow Emphasis} \\
\midrule

AI-Chemist~\citep{Zhu2022AIChemist}
 & \fullmark & \fullmark & \fullmark & \fullmark & --
 & \makecell[l]{General chemistry}
 & Robotic chemistry loop \\

A-Lab~\citep{Szymanski2023AutonomousLab}
 & \halfmark & \fullmark & \fullmark & \fullmark & --
 & \makecell[l]{Inorganic materials}
 & Closed-loop solid-state synthesis \\

GNoME~\citep{Merchant2023Scaling}
 & -- & \fullmark & \fullmark & \halfmark & --
 & \makecell[l]{Materials discovery}
 & Computational materials screening \\

LLM-RDF~\citep{Ruan2024LLMRDF}
 & \fullmark & \fullmark & \fullmark & \fullmark & --
 & \makecell[l]{Reaction development}
 & End-to-end reaction development \\

ChemAgents~\citep{Song2025ChemAgents}
 & \fullmark & \fullmark & \fullmark & \halfmark & --
 & \makecell[l]{Robotic chemistry}
 & Multi-agent robotic chemist \\

Coscientist~\citep{Boiko2023Autonomous}
 & \halfmark & \fullmark & \fullmark & \halfmark & --
 & \makecell[l]{Organic chemistry}
 & Autonomous chemical experimentation \\

Robot Chemist~\citep{Burger2020MobileRobotChemist}
 & -- & \halfmark & \fullmark & \fullmark & --
 & \makecell[l]{Photocatalyst search}
 & Mobile robotic optimization \\

ChemCrow~\citep{Bran2024ChemCrow}
 & \fullmark & \halfmark & \halfmark & --
 & --
 & \makecell[l]{Chemistry tool use}
 & Chemistry tool orchestration \\

ChemOS~\citep{Roch2020ChemOS}
 & -- & \halfmark & \fullmark & \halfmark & --
 & \makecell[l]{Lab orchestration}
 & Self-driving lab infrastructure \\

\bottomrule
\end{tabularx}
\end{adjustbox}
\end{table*}

\begin{itemize}[nolistsep, leftmargin=*]

    \item[] \textcolor[HTML]{4C78A8}{\largedot} \textbf{Structured experimental substrate.}
    Chemistry and materials occupy a higher empirical-autonomy position because many research operations are already representable as structured scientific objects. Reaction conditions, molecular transformations, retrosynthetic routes, materials compositions, synthesis procedures, and assay or characterization readouts form a machine-actionable interface between hypothesis formation and feedback. Robotic platforms and computational screening further shorten the loop between candidate generation and validation. The autonomy advantage is therefore domain-specific: chemical and materials workflows can be partially closed when design spaces are bounded, actions are protocolized, and feedback is measurable.

    \item[] \textcolor[HTML]{F58518}{\largedot} \textbf{Domain-native systems.}
    Current systems span robotic chemistry, autonomous materials synthesis, computational discovery, and laboratory orchestration. \textit{AI-Chemist} integrates machine reading, robotic experimentation, and a computational brain for iterative chemical research~\citep{Zhu2022AIChemist}. \textit{A-Lab} targets autonomous solid-state synthesis by combining computational guidance, literature-derived knowledge, active learning, and robotic execution~\citep{Szymanski2023AutonomousLab}. \textit{GNoME} expands the computational side of materials discovery through large-scale candidate generation and stability filtering~\citep{Merchant2023Scaling}. \textit{LLM-RDF} addresses end-to-end reaction development across knowledge extraction, synthesis design, and optimization~\citep{Ruan2024LLMRDF}. \textit{ChemAgents}, \textit{Coscientist}, and \textit{Robot Chemist} strengthen the experimental side through multi-agent robotic chemistry, autonomous chemical experimentation, and closed-loop photocatalyst optimization~\citep{Song2025ChemAgents, Boiko2023Autonomous, Burger2020MobileRobotChemist}. \textit{ChemCrow} and \textit{ChemOS} support the tool and infrastructure layer through chemistry-specific tool orchestration and self-driving laboratory operation~\citep{Bran2024ChemCrow, Roch2020ChemOS}.

    \item[] \textcolor[HTML]{54A24B}{\largedot} \textbf{Autonomy boundary.}
    Chemistry and materials support narrow higher-autonomy islands, but not broad autonomous scientific closure. Current systems are strongest when objectives are explicit, search spaces are bounded, and execution is computationally screened or robotically controlled. Their limits appear in protocol portability, cross-laboratory reproducibility, evidence synthesis across heterogeneous studies, negative-result reuse, and continuous updating from new literature and failed experiments. The domain therefore advances beyond bounded assistance through robotic and computational closure, while remaining below robust \textcolor{RefOrange}{L4} autonomy because chemical validity still depends on physical feasibility, experimental reproducibility, laboratory context, and evidence integration.

\end{itemize}

\subsection{Biology and Biomedicine}
\label{sec:domain_biology}

Biology and biomedicine occupy a middle-to-advanced \textcolor{RefOrange}{L2} position in the AutoResearch landscape. The domain is increasingly machine-actionable, but not organized around a single substrate. Biological workflows span single-cell and omics analysis, systems-biology model refinement, DBTL-style bioengineering, automated wet-lab execution, regenerative cell-culture optimization, and biomolecular workflow design. These settings support bounded evidence loops rather than broad biomedical closure: biological systems remain heterogeneous, context-dependent, nonstationary, and difficult to stabilize across cells, tissues, organisms, protocols, laboratories, and translational settings.

\begin{table*}[h]
\centering
\surveytablecaption{Representative biology and biomedicine AutoResearch systems.}{Representative domain-native systems spanning computational biology, DBTL optimization, systems-biology automation, automated wet-lab experimentation, regenerative medicine, and biomolecular engineering. Stage notation follows Table \ref{tab:computational_systems}.}
\label{tab:biology_systems}
\renewcommand{\arraystretch}{1.10}
\setlength{\tabcolsep}{2.4pt}
\rowcolors{2}{white}{black!2}
\begin{adjustbox}{max width=\textwidth}
\small
\begin{tabularx}{\textwidth}{@{}p{3.95cm} p{0.5cm} p{0.5cm} p{0.5cm} p{0.5cm} p{0.5cm} p{3.65cm} X@{}}
\toprule[1pt]
\textbf{System} & \textbf{I} & \textbf{II} & \textbf{III} & \textbf{IV} & \textbf{V} & \textbf{Subdomain} & \textbf{Workflow Emphasis} \\
\midrule

CellVoyager~\citep{Alber2025CellVoyager}
 & \fullmark & \fullmark & \fullmark & \halfmark & --
 & \makecell[l]{Computational biology}
 & Autonomous scRNA-seq analysis \\

BioAutomata~\citep{HamediRad2019BioAutomata}
 & -- & \fullmark & \fullmark & \fullmark & --
 & \makecell[l]{Metabolic engineering}
 & Closed-loop DBTL optimization \\

Genesis~\citep{Tiukova2024Genesis}
 & \fullmark & \fullmark & \fullmark & \fullmark & --
 & \makecell[l]{Systems biology}
 & Model-refinement automation \\

BioMARS~\citep{qiu2025biomars}
 & \fullmark & \fullmark & \fullmark & \halfmark & --
 & \makecell[l]{Automated wet lab}
 & Multi-agent biological experiments \\

AutoDNA~\citep{Wu2025Native}
 & \halfmark & \fullmark & \fullmark & \halfmark & --
 & \makecell[l]{Biomolecular engineering}
 & Autonomous nucleic-acid workflow \\

\bottomrule
\end{tabularx}
\end{adjustbox}
\end{table*}

\begin{itemize}[nolistsep, leftmargin=*]

    \item[] \textcolor[HTML]{4C78A8}{\largedot} \textbf{Heterogeneous biological substrate.}
    Biology and biomedicine support partial autonomy through several distinct evidence loops rather than one unified automation pathway. Single-cell and omics datasets enable computational analysis loops; metabolic-engineering and biofoundry settings support DBTL-style optimization; systems biology provides model-revision loops; automated wet-lab platforms stabilize selected experimental procedures; biomolecular engineering links protocol design to instrument execution. The domain's current stage is therefore shaped by bounded biological scope, measurable feedback, and protocolized execution, rather than by general biomedical automation.

    \item[] \textcolor[HTML]{F58518}{\largedot} \textbf{Domain-native systems.}
    Current systems map onto different parts of the biological evidence loop. \textit{CellVoyager} autonomously explores scRNA-seq datasets and implements computational biology analyses~\citep{Alber2025CellVoyager}. \textit{BioAutomata} couples machine-learning-guided design with automated DBTL execution for metabolic engineering~\citep{HamediRad2019BioAutomata}. \textit{Genesis} targets systems-biology model improvement through structured knowledge, model revision, and closed-loop experimental planning~\citep{Tiukova2024Genesis}. \textit{BioMARS} integrates language-based reasoning, multimodal inspection, and modular laboratory automation for biological experiment design and execution~\citep{qiu2025biomars}. \textit{AutoDNA} extends autonomous experimentation toward nucleic-acid and biomolecular-engineering workflows~\citep{Wu2025Native}.

    \item[] \textcolor[HTML]{54A24B}{\largedot} \textbf{Autonomy boundary.}
    Biology and biomedicine support selective higher-autonomy islands, but not broad AI-led biomedical closure. Current systems are strongest when the biological object is narrowed, the protocol is stabilized, the assay is repeatable, and feedback can be quantified within a bounded experimental or computational setting. Their limits appear in biological heterogeneity, protocol drift, multimodal integration, nonstationary living systems, cross-laboratory reproducibility, translational uncertainty, and weak standardization across autonomous biological laboratories. Future progress depends on reproducible biological evidence loops that connect computational analysis, automated experimentation, model revision, provenance tracking, and wet-lab validation. The domain remains below robust \textcolor{RefOrange}{L4} autonomy because biological validity depends on context, embodiment, and reproducibility across living systems.

\end{itemize}

\subsection{Medicine and Clinical Research}
\label{sec:domain_medicine}

Medicine and clinical research remain in early-to-middle \textcolor{RefOrange}{L2}. The domain already supports substantial AI participation in evidence search, trial screening, data extraction, systematic review, network meta-analysis, living evidence synthesis, and supervised clinical research automation. Its autonomy ceiling, however, is shaped less by technical execution than by \emph{normative exposure}: medical claims are filtered through patient safety, clinical validity, regulatory standards, ethical oversight, liability, and retained human responsibility. The most credible medical AutoResearch workflows are therefore evidence-centric rather than patient-autonomous. They accelerate clinical research synthesis and supervised research production, but do not externalize responsibility for patient-facing decisions.

\begin{table*}[h]
\centering
\surveytablecaption{Representative medicine and clinical-research AutoResearch systems.}{Representative domain-native systems spanning clinical evidence synthesis, systematic review, network meta-analysis, living evidence summaries, medical literature mining, and supervised medical research automation. Stage notation follows Table \ref{tab:computational_systems}.}
\label{tab:medicine_systems}
\renewcommand{\arraystretch}{1.10}
\setlength{\tabcolsep}{2.4pt}
\rowcolors{2}{white}{black!2}
\begin{adjustbox}{max width=\textwidth}
\small
\begin{tabularx}{\textwidth}{@{}p{3.65cm} p{0.5cm} p{0.5cm} p{0.5cm} p{0.5cm} p{0.5cm} p{4.25cm} X@{}}
\toprule[1pt]
\textbf{System} & \textbf{I} & \textbf{II} & \textbf{III} & \textbf{IV} & \textbf{V} & \textbf{Subdomain} & \textbf{Workflow Emphasis} \\
\midrule

TrialMind~\citep{Ong_2025}
 & \fullmark & \halfmark & \halfmark & \fullmark & \halfmark
 & \makecell[l]{Clinical evidence synthesis}
 & Literature-to-evidence pipeline \\

MetaMind~\citep{MetaMind2025}
 & \fullmark & \halfmark & \fullmark & \fullmark & \halfmark
 & \makecell[l]{Network meta-analysis}
 & Automated NMA workflow \\

SOLES~\citep{SOLES2024}
 & \fullmark & \halfmark & \halfmark & \fullmark & \halfmark
 & \makecell[l]{Living evidence synthesis}
 & Continuous evidence summary \\

AID-SLR ~\citep{lee2024aid}
 & \fullmark & \halfmark & \halfmark & \halfmark & \halfmark
 & \makecell[l]{Systematic reviews}
 & Agentic SLR workflow \\

Medical AI Scientist~\citep{wu2026towards}
 & \fullmark & \fullmark & \fullmark & \halfmark & \fullmark
 & \makecell[l]{Clinical research}
 & Supervised medical research loop \\

\bottomrule
\end{tabularx}
\end{adjustbox}
\end{table*}

\begin{itemize}[nolistsep, leftmargin=*]

    \item[] \textcolor[HTML]{4C78A8}{\largedot} \textbf{Evidence-centric substrate.}
    Medical AutoResearch is strongest where clinical knowledge is represented as literature, trial records, eligibility criteria, extracted outcomes, evidence tables, systematic reviews, meta-analytic models, or living summaries. These artifacts provide a structured interface between retrieval, screening, synthesis, and validation. The domain's current progress therefore concentrates on epistemic workflows: assembling, checking, updating, and summarizing medical evidence under human oversight. This differs from chemistry, biology, and computational research, where higher autonomy can more often be tied to experimental execution. In medicine, workflow automation advances first where evidence can be curated, traced, and audited without direct patient intervention.

    \item[] \textcolor[HTML]{F58518}{\largedot} \textbf{Domain-native systems.}
    Current systems concentrate around clinical evidence synthesis and supervised medical research production. \textit{TrialMind} automates major stages of clinical systematic review, including literature search, eligibility screening, and data extraction~\citep{Ong_2025}. \textit{MetaMind} extends this pattern to network meta-analysis by integrating retrieval, extraction, and statistical model execution~\citep{MetaMind2025}. \textit{SOLES} supports living evidence synthesis through continuous collection, curation, visualization, and evidence updating~\citep{SOLES2024}. \textit{AID-SLR} and \textit{A4SLR} represent agentic systematic-review workflows for clinical research, health economics, and HTA-oriented evidence synthesis~\citep{lee2024aid}.  \textit{Medical AI Scientist} moves closest to workflow-level medical AutoResearch by connecting evidence-grounded ideation, medical experimentation, manuscript drafting, and review under supervised clinical research modes~\citep{wu2026towards}.

    \item[] \textcolor[HTML]{54A24B}{\largedot} \textbf{Autonomy boundary.}
    Medicine and clinical research support meaningful workflow automation, but not broad autonomous clinical closure. Current systems remain strongest in evidence synthesis, literature mining, structured extraction, statistical synthesis, and supervised research generation. Their limits appear when evidence must propagate into patient-relevant recommendation, clinical decision-making, guideline formation, or intervention. Stage errors in retrieval, screening, extraction, and synthesis can alter downstream conclusions; heterogeneous evidence from randomized trials, observational studies, diagnostic studies, and real-world data remains difficult to integrate; regulatory-grade provenance and uncertainty propagation remain uneven. Medical LLM competence on static QA or benchmark-style reasoning does not establish clinical autonomy, and simulated clinical-agent benchmarks expose the additional difficulty of interactive diagnosis, uncertainty handling, bias, and patient-context dependence~\citep{Qiu2024LLM, Schmidgall2024AgentClinic}. The domain remains below robust \textcolor{RefOrange}{L4} autonomy because medical validity is inseparable from safety, accountability, governance, and retained human responsibility.

\end{itemize}

\subsection{Economics and Social Sciences}
\label{sec:domain_social}

Economics and social sciences remain in early-to-middle \textcolor{RefOrange}{L2}. Their workflows are highly compatible with AI-assisted literature search, data processing, coding, drafting, and exploratory synthesis, yet much harder to close autonomously when research claims depend on causal identification, institutional context, field interpretation, and normative judgment. Unlike computational research, where execution and validation can often be encoded as benchmark or code feedback, social-scientific validity depends on whether data, assumptions, empirical strategies, and interpretations remain defensible under domain-specific standards. Current progress therefore concentrates in human--AI research-question generation and dataset-aware empirical pipelines, while strong autonomous closure remains rare.

\begin{table*}[h]
\centering
\surveytablecaption{Representative economics and social-science AutoResearch systems.}{Representative domain-native systems spanning human--AI research-question generation and dataset-aware empirical research automation. Stage notation follows Table \ref{tab:computational_systems}.}
\label{tab:social_science_systems}
\renewcommand{\arraystretch}{1.10}
\setlength{\tabcolsep}{2.4pt}
\rowcolors{2}{white}{black!2}
\begin{adjustbox}{max width=\textwidth}
\small
\begin{tabularx}{\textwidth}{@{}p{3.65cm} p{0.5cm} p{0.5cm} p{0.5cm} p{0.5cm} p{0.5cm} p{3.65cm} X@{}}
\toprule[1pt]
\textbf{System} & \textbf{I} & \textbf{II} & \textbf{III} & \textbf{IV} & \textbf{V} & \textbf{Subdomain} & \textbf{Workflow Emphasis} \\
\midrule

HybridQuestion~\citep{Zhao2025HybridQuestion}
& \fullmark & \fullmark & -- & \halfmark & --
& \makecell[l]{Research-question\\ generation}
& Human--AI question selection \\

HLER~\citep{Undermind2026HLERHumanintheLoopEconomicRe}
& \halfmark & \fullmark & \fullmark & \halfmark & \fullmark
& \makecell[l]{Empirical economics}
& Dataset-aware research pipeline \\

\bottomrule
\end{tabularx}
\end{adjustbox}
\end{table*}

\begin{itemize}[nolistsep, leftmargin=*]

    \item[] \textcolor[HTML]{4C78A8}{\largedot} \textbf{Interpretive empirical substrate.}
    Economics and social sciences are rich in text, observational data, survey instruments, administrative records, institutional documentation, and statistical code, but their research objects are not easily closed through execution alone. Empirical claims depend on identification assumptions, measurement choices, sample construction, institutional setting, historical context, and theory-informed interpretation. This substrate supports strong bounded assistance and partial workflow execution, especially in search, synthesis, data preparation, econometric coding, and manuscript drafting. The autonomy ceiling remains lower than in computational or laboratory-optimized domains because validity is tied to causal and interpretive defensibility rather than output production alone.

    \item[] \textcolor[HTML]{F58518}{\largedot} \textbf{Domain-native systems.}
    Current workflow-level systems remain sparse. \textit{HybridQuestion} targets the upstream stage of social-scientific research by combining AI-scale information gathering and candidate question generation with progressively stronger human screening for forward-looking research judgment~\citep{Zhao2025HybridQuestion}. \textit{HLER} moves further into empirical execution by organizing data auditing, dataset profiling, hypothesis generation, econometric analysis, manuscript drafting, and review into a human-in-the-loop multi-agent pipeline~\citep{Undermind2026HLERHumanintheLoopEconomicRe}. These systems are representative because they preserve human authority at precisely the points where social-scientific autonomy becomes fragile: question selection, empirical identification, economic significance, and publication-level acceptance.

    \item[] \textcolor[HTML]{54A24B}{\largedot} \textbf{Autonomy boundary.}
    Economics and social sciences support selective workflow participation, but not strong autonomous research closure. Current systems can expand candidate questions, structure datasets, generate feasible hypotheses, run bounded empirical analyses, and draft manuscripts. Their limits appear in causal identification, construct validity, external validity, institutional interpretation, ethical sensitivity, and the evaluation of social or economic significance. Future progress depends on agentic workflows that treat identification strategy, data provenance, robustness checks, theoretical framing, and human review as first-class components rather than downstream edits. The domain remains below robust \textcolor{RefOrange}{L4} autonomy because social-scientific validity cannot be reduced to data availability, model fluency, or executable analysis.

\end{itemize}

\subsection{Earth and Environmental Sciences}
\label{sec:domain_earth}

Earth and environmental sciences occupy a selective-to-advanced \textcolor{RefOrange}{L2} position in the AutoResearch landscape. The domain combines highly structured data infrastructures with limited experimental manipulability: reanalysis products, satellite observations, remote-sensing archives, climate datasets, geophysical records, and numerical simulators are increasingly machine-actionable, yet the Earth system itself cannot be freely intervened on, replayed, or branched like a software environment or laboratory protocol. Current autonomy is strongest in climate-data workflows, atmospheric diagnosis, numerical-model-supported hypothesis testing, and research-copilot settings. Broad scientific closure remains constrained by non-manipulable real-world dynamics, long validation horizons, heterogeneous standards, and the gap between predictive performance and causal or mechanistic understanding.

\begin{table*}[h]
\centering
\surveytablecaption{Representative Earth and environmental AutoResearch systems.}{Representative domain-native systems spanning climate research copilot workflows, atmospheric mechanism verification, and knowledge-graph-based climate data science. Stage notation follows Table \ref{tab:computational_systems}.}
\label{tab:earth_systems}
\renewcommand{\arraystretch}{1.10}
\setlength{\tabcolsep}{2.4pt}
\rowcolors{2}{white}{black!2}
\begin{adjustbox}{max width=\textwidth}
\small
\begin{tabularx}{\textwidth}{@{}p{3.65cm} p{0.5cm} p{0.5cm} p{0.5cm} p{0.5cm} p{0.5cm} p{3.65cm} X@{}}
\toprule[1pt]
\textbf{System} & \textbf{I} & \textbf{II} & \textbf{III} & \textbf{IV} & \textbf{V} & \textbf{Subdomain} & \textbf{Workflow Emphasis} \\
\midrule

EarthLink~\citep{EarthLink2025}
& \fullmark & \fullmark & \fullmark & \halfmark & \fullmark
& \makecell[l]{Climate / Earth system}
& Climate research copilot \\

TianJi~\citep{TianJi2026}
& \fullmark & \fullmark & \fullmark & \fullmark & \halfmark
& \makecell[l]{Atmospheric science}
& WRF-based mechanism testing \\

AutoClimDS~\citep{AutoClimDS2025}
& \halfmark & \fullmark & \fullmark & \fullmark & \halfmark
& \makecell[l]{Climate data science}
& KG-based data workflow \\

\bottomrule
\end{tabularx}
\end{adjustbox}
\end{table*}

\begin{itemize}[nolistsep, leftmargin=*]

    \item[] \textcolor[HTML]{4C78A8}{\largedot} \textbf{Non-manipulable data-rich substrate.}
    Earth and environmental sciences provide rich machine-actionable substrates through reanalysis products, Earth-observation archives, climate model outputs, geophysical datasets, meteorological indices, and numerical simulators. These substrates support strong prediction, diagnostics, data retrieval, and computational experimentation. The autonomy ceiling remains lower than in code-native or tightly instrumented domains because validation depends on delayed real-world evolution, imperfect observations, physical interpretability, and cross-scale consistency.

    \item[] \textcolor[HTML]{F58518}{\largedot} \textbf{Domain-native systems.}
    Current AutoResearch systems concentrate in climate research assistance, atmospheric mechanism verification, and climate-data workflow orchestration. \textit{EarthLink} connects natural-language interaction, planning, executable code, data analysis, physical reasoning, and interpretive reporting for climate-science workflows~\citep{EarthLink2025}. \textit{TianJi} targets atmospheric mechanism discovery by coupling literature-informed hypothesis generation with WRF-centered numerical verification and explanation~\citep{TianJi2026}. \textit{AutoClimDS} organizes climate datasets, metadata, tools, and workflows through a knowledge graph, enabling natural-language-driven data discovery, acquisition, analysis, and figure reproduction~\citep{AutoClimDS2025}. Model-centric systems such as \textit{GraphCast}, \textit{GenCast}, \textit{Pangu-Weather}, and \textit{ClimaX} provide essential forecasting and foundation-model capability, but they remain narrower than workflow-level AutoResearch because they primarily automate prediction rather than research-loop orchestration~\citep{Lam2023GraphCast, Price2024GenCast, Bi2023PanguWeather, Nguyen2023ClimaX}.

    \item[] \textcolor[HTML]{54A24B}{\largedot} \textbf{Autonomy boundary.}
    Earth and environmental sciences support strong data-centric and model-centric automation, but not broad autonomous scientific closure. Current systems can connect datasets, tools, computation, diagnostics, and reporting in bounded workflows, while forecasting models provide high-skill short- to medium-range predictive capability. Their limits appear in causal mechanism validation, multimodal Earth-system integration, semantic consistency across variables and coordinate systems, provenance preservation, and long-horizon evaluation under nonstationary climate conditions. Future progress depends on workflows that integrate literature grounding, dataset semantics, numerical simulation, physical diagnostics, uncertainty propagation, and reproducible provenance. The domain remains below robust \textcolor{RefOrange}{L4} autonomy because Earth-system validity depends on non-manipulable real-world dynamics, delayed verification, and physically interpretable evidence rather than model output alone.

\end{itemize}
\section{Discussion}
\label{sec:challenges_governance}

\subsection{Rethinking the Capabilities of AutoResearch}
\textbf{Current vibe research systems achieve procedural autonomy but fall short of true scientific agency.}
The rapid development of vibe research—ranging from The AI Scientist~\cite{Lu2024AIScientist} and SciAgents~\cite{Ghafarollahi2024SciAgents} to EvoScientist~\cite{Lyu2026EvoScientist}, as well as the broader systems catalogued in Table~\ref{tab:autoresearch_landscape_overview}—has begun to accelerate the scientific discovery process and enhance research efficiency. These systems can cover the full pipeline of a research project, including literature review, hypothesis generation, experimental design, result analysis, and manuscript drafting, all without sustained human intervention. Yet this operational fluency should not be mistaken for genuine scientific creativity. The capabilities of current systems are fundamentally determined by their architecture. Specifically, they rely on an LLM-driven agentic workflow in which each stage of the research process—such as literature review, experimental design, result analysis, and scientific writing—is decomposed into structured, sequentially executable steps. Within this paradigm, tasks that are amenable to high-quality template matching are handled with impressive competence. Stages such as literature review, experimental pipeline implementation, and manuscript drafting are well represented in the training data of foundation models, enabling capable agents to execute them reliably. However, hypothesis generation remains distinct. While other stages can be reduced to retrieving and recombining patterns from prior work, hypothesis generation requires abductive reasoning from observed anomalies to explanatory conjectures not present in the existing works. In this stage, current vibe research systems reveal a structural limitation: the ideas they generate are, in most cases, combinatorial propositions of the form $A + B \rightarrow C$, where $A$ and $B$ are concepts or methods already present in the training data or retrieved from existing work, and $C$ is a recombination rather than a genuine discovery. Scientific agency requires more than recombination. It demands the capacity to identify worthwhile research questions and to develop novel methods. The current paradigm, which relies on inference grounded in training data and retrieved prior work, falls short in this regard. In effect, these systems function as search algorithms rather than architects of the search space. Human-in-the-loop approaches offer a partial solution by preserving human judgment at the hypothesis generation stage, where autonomous systems are most limited, but they do not resolve the underlying constraint. To overcome this limitation, a fundamental rethinking of the operating paradigm of AutoResearch systems is required.

\textbf{Reflexive iteration remains an open and largely unaddressed challenge for AutoResearch.}
Reflexive iteration denotes the fundamental capacity to continuously revise hypotheses, methodologies, and experimental designs in response to experimental results. Such reflexive revision is constitutive of scientific discovery, operating from local experimental adjustments to the reconfiguration of broader research directions. In contrast, current vibe research systems are operated as end-to-end pipelines, where stages are executed sequentially with limited mechanisms for revision or reconfiguration. In these systems, experimental results inform the manuscript but do not propagate back to revise the hypothesis or the problem formulation from which it is derived. AI Scientist-v2~\cite{Yamada2025AIScientistV2} proposes a progressive agentic tree-search architecture that enables the exploration of multiple experimental branches and selection among them. 
However, this search remains confined to a pre-specified solution space, optimizing over candidate implementations within a fixed research direction rather than revising the underlying hypothesis or methodology.
Human researchers engage in reflexive iteration, which drives the accumulation of knowledge and deepens scientific understanding. By contrast, pipeline-structured vibe research systems treat the research topic as a fixed input and optimize the downstream workflow accordingly, making them resistant to iterative revision. As a result, they are effective at completing a paper but not at conducting high-quality research, often producing well-executed studies around weak hypotheses without recognizing or correcting their deficiencies. Overcoming this limitation requires a paradigm shift from pipeline-structured optimization to iterative, reflexive research systems.

\subsection{Evaluation for AutoResearch}

\textbf{Workflow success vs. scientific quality.}
A central limitation of the current AutoResearch evaluation is that it more readily measures whether a system can successfully execute a research workflow than whether it produces scientifically valuable outcomes. As summarized in Section~\ref{sec:evaluation_framework}, the scientific quality of AutoResearch encompasses five essential dimensions: novelty, validity, impact, reliability, and provenance. Among these, current benchmarks~\cite{huang2023mlagentbench, Kon2025EXPBench, SciRepBench2025} are better aligned with validity and reliability, since these dimensions are more amenable to executable checks, reruns, and performance-grounded evaluation. However, benchmarks remain partial and insufficient. Existing evaluations~\cite{Patel2025DeepScholar, Kon2025EXPBench} often focus on literature review, hypothesis generation, experiment implementation, result analysis, and manuscript drafting, all of which are important components of the research process. However, the successful completion of a research workflow does not equate to scientific value. A system may generate a polished paper, runnable code, or experimental results while still pursuing an unimportant question, advancing a derivative idea, or drawing conclusions of limited scientific significance. Therefore, future evaluations should move from checking whether a workflow is finished to assessing novelty, impact, and provenance. This ensures that AutoResearch systems are judged by their real scientific value rather than just their ability to follow a set of research steps.

\textbf{Novelty lacks an effective operational definition.}
Among all dimensions of scientific quality, novelty is the most essential and the hardest to evaluate. The fundamental difficulty is not merely technical but conceptual. There is no consensus on what novelty means as a measurable property of scientific output. Novelty refers to a genuine advance—a new way of framing a problem, an unexpected connection between domains, or a question that meaningfully extends the frontier of knowledge~\cite{boudreau2016looking}. A research idea may appear different from prior work in presentation while remaining conceptually derivative. Conversely, an idea that seems modest or incremental may still embody an important insight. This asymmetry makes novelty resistant to any definition grounded in observable surface features. Therefore, existing evaluations rely on weak proxies, such as dissimilarity from prior papers~\cite{Chen2025AIRSBench}
, LLM-as-a-judge~\cite{Majumder2024DiscoveryBench}, or human expert judgment~\cite{Liu2025ResearchBench}. Human expert judgment is costly, slow, and often variable across reviewers. In contrast, LLM-as-a-Judge is easily fooled by shallow patterns, potentially confusing a well-presented paper with a truly significant contribution. As a result, the current evaluations still lack a principled way to distinguish genuine research contributions from sophisticated remixing. Moving forward, it is essential to develop novelty assessments that are aware of the existing literature, grounded in temporal context, and mediated by expert evaluation. This approach would be more effective than relying solely on surface differences or single-pass judgments as indicators of scientific originality.

\textbf{Scientific impact resists short-term evaluation.}
Impact is even less compatible with current evaluation paradigms, because scientific impact is fundamentally long-horizon, delayed, and cumulative~\cite{wang2013quantifying}. The impact of a research work lies in whether it redirects subsequent inquiry, improves methodological practice, enables downstream discoveries, or accumulates into durable knowledge through later uptake~\cite{park2023papers}. Such impact is typically evidenced through longitudinal signals, including citations over time, reuse of artifacts such as code, datasets, models, or benchmarks, adoption of methods or problem formulations by other researchers, and expert follow-up through replications, extensions, or downstream studies. 
However, current AutoResearch evaluations are predominantly short-horizon and immediate. They typically rely on metrics that can be measured immediately after a run, such as task success or benchmark improvements~\cite{Chen2025AIRSBench}. This leads to a structural mismatch where the easiest metrics to track often fail to capture long-term scientific value. Currently, the most common way to handle the difficulty of measuring impact is to rely on human experts or LLM-as-a-judge~\cite{Wang2025LiveResearchBench}. While such methods offer flexibility and domain sensitivity, they introduce serious challenges of their own—including inconsistency across evaluators, susceptibility to presentation effects, limited scalability, and the absence of ground truth against which judgments can be calibrated. Neither paradigm constitutes a robust solution, both function as pragmatic approximations in the absence of more principled alternatives. Therefore, impact evaluation should move beyond immediate output-level judgments toward longitudinal protocols that track the adoption, reuse, and downstream influence of AutoResearch-generated papers.

\subsection{The Generalization Gap: Beyond Computational and Formal Sciences}
As introduced in Section~\ref{sec:domain}, AutoResearch has progressed much faster in computational and formal sciences than in other scientific domains. This pattern should not be overinterpreted as evidence of domain-general scientific autonomy. Instead, it reveals a clear boundary: these systems perform best when the research process is already software-based. Computational and formal sciences provide the most favorable substrate for AI-led research because their core objects of inquiry—including code, proofs, simulators, datasets, and executable outputs-are already machine-readable, executable, replayable, and comparatively easy to validate. Under these conditions, research workflows can be more readily decomposed into explicit steps with fast feedback and relatively clear success criteria~\cite{Lu2024AIScientist, ARISGitHub, Weng2025DeepScientist}. This is why current systems achieve their strongest workflow coverage and highest degree of automation in this domain. However, such success reflects a favorable research substrate rather than a universal scientific capability. Once the research process moves beyond a unified digital loop, the basis for autonomy changes substantially.

This limitation becomes especially clear outside computational and formal sciences, where scientific validity depends not only on reasoning or executable artifacts, but also on physical experimentation, embodied interaction, hardware constraints, real-world deployment, and domain-specific validation procedures. In autonomous driving, robotics, and embodied intelligence, progress cannot be validated solely by code execution or simulation. Hypotheses must withstand contact with sensors, actuators, the environment, safety constraints, and distribution shifts in the physical world. A promising result in simulation may fail once perception, control, and environment dynamics are coupled in reality. 
Similar breaks arise in chemistry and materials, where a computationally plausible design or retrosynthetic plan may not remain feasible under laboratory conditions, resource constraints, or instrument limitations. In biology and biomedicine, computational hypotheses must survive wet-lab validation, biological variability, and multi-stage experimental uncertainty; in medicine and clinical research, benchmark performance does not automatically translate into accountable intervention under clinical, ethical, and regulatory constraints. These domain-specific validation requirements make it harder to close the AutoResearch loop outside computational and formal sciences. In computational domains, feedback can often be obtained directly through code execution, simulation, or proof checking. In physical, biological, and clinical domains, however, feedback is slower, more expensive, and often mediated by hardware, laboratories, human experts, or regulatory procedures. As a result, AutoResearch progresses more slowly in these domains, since its validation pipelines are harder to automate end-to-end. This suggests that scientific autonomy will be domain-conditioned rather than universal. Computational fields may achieve more complete automation earlier, while other domains will move toward autonomy as physical platforms, automated laboratories, embodied testing systems, and real-world deployment environments become more reliable and scalable.

\subsection{Reliability, Trustworthiness, and Auditability of AutoResearch}

\textbf{Reliability.} 
In AutoResearch, reliability refers to whether an AI-generated research process can produce stable, reproducible, and scientifically valid outputs across the full research workflow. A reliable system should not only generate plausible ideas or fluent manuscripts, but also maintain correct evidence use, sound methodological choices, executable experiments, faithful result interpretation, and reproducible claims. A major source of reliability failure in AutoResearch systems stems from their dependence on LLMs as research operators. LLMs are used to retrieve and summarize literature, synthesize evidence, propose hypotheses, design experiments, write code, interpret results, and draft scientific claims. Although retrieval augmentation, tool execution, and human feedback can mitigate some errors, they do not remove the risk that LLMs generate unsupported, distorted, or fabricated content. In a multi-stage research workflow, such hallucinations are especially consequential because they can be absorbed into intermediate outputs (e.g., related works, experimental plan) and reused by later stages. For example, a hallucinated citation may distort the literature context, a misread paper may motivate an invalid hypothesis, and an overconfident interpretation of preliminary results may lead to misleading experimental decisions or overstated conclusions. Thus, reliability in AutoResearch requires not only step-level correctness but also robust evidence grounding, cross-stage consistency, and safeguards against the accumulation and propagation of errors throughout the research workflow.


\textbf{Trustworthiness.} 
Trustworthiness in AutoResearch depends on whether its scientific claims remain grounded in valid evidence, respect relevant constraints, and resist manipulation. An AutoResearch system may produce a polished hypothesis, a persuasive analysis, or a publication-ready paper while obscuring whether its central claims are supported by genuine evidence, contaminated retrieval, unsupported inference, or selectively interpreted results. This concern is intensified by safety vulnerabilities that bear directly on whether the system should be trusted in scientific settings. For example, prompt injection can cause the system to privilege malicious or irrelevant instructions over the intended research objective, such as following attacker-controlled directives embedded in retrieved documents, external tools, or intermediate artifacts, thereby distorting literature review, steering hypothesis generation toward injected goals, biasing experimental design or evaluation, and ultimately producing conclusions unsupported by the underlying evidence. 
Recent work has further shown that modular AutoResearch infrastructures introduce new workflow-level attack surfaces. \textit{BadSkill}~\citep{tie2026badskill} demonstrated that adversaries can poison reusable agent skills through model-in-skill backdoor attacks, causing downstream research agents to execute malicious behaviors while preserving seemingly benign functionality. In addition, jailbreaks can weaken safeguards designed to enforce scientific, safety, or ethical constraints, and privacy-sensitive data may be improperly exposed, memorized, or propagated through downstream outputs. Under these safety threats, fluency, coherence, and even empirically styled outputs are weak indicators of trustworthiness. What matters instead is whether the system remains grounded in valid evidence, respects methodological and ethical constraints, handles sensitive information appropriately, and stays robust to manipulation throughout the research process.

\textbf{Auditability.} 
Auditability is crucial because failures in AutoResearch often remain hidden once they are integrated into a seemingly coherent research workflow. Current systems may log prompts, retrieved documents, tool calls, intermediate outputs, and execution traces. However, simply collecting these traces does not ensure auditability. An auditable AutoResearch system should enable the reconstruction of how a claim was formed, including which evidence and tools influenced it, whether intermediate decisions were altered by prompt injection or other forms of adversarial interference, and whether sensitive data were improperly accessed, retained, or exposed. 
Additionally, the system should support error localization, helping to determine whether a failure originated from model hallucinations, compromised retrieval, unsafe tool behavior, orchestration errors, privacy breaches, or insufficient human oversight. In the context of scientific research, auditability also necessitates evidence provenance, recoverable intermediate states, and clear responsibility boundaries between model inference, tool outputs, and human intervention. 
Without such mechanisms, attributing scientific failures and safety violations becomes challenging, making it difficult to implement corrections. Therefore, auditability is not just a technical convenience but a fundamental requirement for reproducibility, accountability, and responsible AutoResearch systems.

\subsection{Ethical and Societal Implications}

The societal implications of AutoResearch extend beyond efficiency and automation. As these systems become increasingly involved in the production of scientific knowledge, they begin to reshape how resources, credit, responsibility, and epistemic authority are distributed across the scientific ecosystem.

\textbf{Asymmetric access to research resources.}
AutoResearch may not democratize science as much as it redistributes advantages toward already resource-rich institutions and organizations. While these systems can lower some local barriers to research, such as drafting, coding, and literature navigation, the decisive bottlenecks of frontier science remain expensive and unevenly distributed. Access to stronger foundation models, large-scale compute, proprietary data, specialized tools, high-quality experimental infrastructure, and domain-specific validation environments remains concentrated in well-funded institutions and large technology organizations. As a result, AutoResearch may reduce the cost of participation at the margins while simultaneously amplifying the advantages of actors who already control the resources needed to operationalize and validate AI-driven research at scale. The societal risk, therefore, is not merely unequal adoption, but a further concentration of epistemic power in a small number of organizations that are able to couple autonomous research agents with superior infrastructure.

\textbf{Scalable paper production.}
A major risk of AutoResearch is that it makes scientific papers easier to optimize as outputs than scientific discovery itself. Once literature synthesis, experiment scripting, result visualization, and manuscript drafting can be partially automated, the paper increasingly becomes an optimizable artifact that can be strategically produced at scale. Under current academic incentives, this creates pressure to use AutoResearch not only to accelerate discovery but also to accelerate the production of publishable-looking manuscripts. Systems may be used to rapidly package marginal ideas, overstate novelty, tailor narratives to venue expectations, or generate multiple superficially distinct manuscripts from limited underlying contributions. The concern extends beyond plagiarism or ghostwriting to include fundamental distortions in scientific communication, in which papers may not reflect genuine intellectual and empirical contributions.

\textbf{Knowledge ecosystem pollution.}
The resulting danger is not merely an increase in weak papers, but the large-scale pollution of the scientific knowledge ecosystem. If AutoResearch enables the mass production of low-value, weakly validated, or strategically optimized research artifacts, these papers do not remain isolated. They enter search results, citation networks, literature review pipelines, benchmark construction, and future training corpora. In this way, low-quality outputs can recursively shape both human and machine knowledge production. Researchers must navigate a noisier literature environment, while future AutoResearch systems may retrieve, summarize, and build upon artifacts whose scientific value is limited or uncertain.

\textbf{Authorship, ownership, and accountability.}
AutoResearch challenges existing norms for assigning authorship, intellectual ownership, and responsibility in scientific work. 
As AI systems become involved in brainstorming, literature synthesis, hypothesis generation, experimental design, analysis, and manuscript drafting, it becomes harder to distinguish human intellectual contribution from automated generation or optimization. 
This ambiguity complicates both credit and accountability: researchers may receive authorship for outputs substantially shaped by AI systems, while responsibility for invalid, harmful, plagiaristic, or ethically problematic results may be distributed across users, model developers, platform providers, data curators, and institutions. 
The issue is further complicated by the use of copyrighted or restricted training data, which leaves the ownership status of AI-generated research outputs legally and ethically unsettled. 
Because existing academic norms were designed around human research teams, they are poorly equipped to govern workflows in which agency, contribution, and responsibility are distributed across humans, models, tools, and infrastructures. 
A central challenge is to develop governance frameworks that can assign credit, ownership, and accountability across human--AI research workflows.


\section{Conclusion}
\label{sec:future_conclusion}

This survey argues that AutoResearch is no longer a loose collection of isolated AI tools for scientific assistance, but an emerging workflow-level paradigm for reorganizing how scientific work is grounded, planned, executed, validated, and communicated. Recent systems have made this transition increasingly visible: The AI Scientist connects idea generation, code writing, experiment execution, manuscript drafting, and review-style feedback within an end-to-end AI research pipeline; Co-Scientist emphasizes multi-agent hypothesis generation and collaborative scientific reasoning; Robin extends the discussion toward iterative lab-in-the-loop discovery; and ERA highlights scientific software generation as a central implementation bottleneck in computational research. Together, these developments show that the frontier of AutoResearch is shifting from local assistance toward broader workflow automation, but they also reinforce the need for a conservative distinction between pipeline breadth and scientific autonomy. Current systems can increasingly generate, execute, analyze, and report, yet robust scientific closure still depends on evidence preservation, provenance, rejection of weak directions, failure recovery, domain-grounded validation, and accountable acceptance. By organizing the field through a workflow-centered autonomy spectrum, this survey shows that today’s practical frontier remains concentrated in human-steered assistance, human-verified execution, interactive co-research, and increasingly integrated pipeline automation, while fully autonomous scientific discovery remains an aspirational horizon rather than an achieved regime. This frontier is also strongly domain-conditioned: computational and formal sciences provide the most favorable conditions for workflow closure because artifacts are executable, replayable, and comparatively easy to verify, whereas chemistry, materials, biology, medicine, social science, Earth science, and embodied settings impose stricter autonomy ceilings through experimental latency, physical intervention, heterogeneous evidence, causal uncertainty, safety constraints, and institutional accountability. The future of AutoResearch should therefore not be framed as an unconstrained race to remove humans from science, but as the deliberate construction of reliable, domain-aware, and auditable research infrastructures that expand the search space of inquiry, accelerate executable parts of the workflow, preserve inspectable provenance, and amplify human scientific creativity under accountable oversight. In this sense, AutoResearch should ultimately be evaluated not by whether it replaces scientific judgment, but by whether it enables more rigorous, reproducible, and trustworthy scientific discovery.

\bibliographystyle{unsrt}

\bibliography{main}

\end{document}